\documentclass{article}
\usepackage{iclr2023_conference,times}
\usepackage[utf8]{inputenc} 
\usepackage[T1]{fontenc}    
\usepackage{times}
\usepackage{graphicx}
\usepackage{amsmath,amssymb,mathbbol}
\usepackage{acronym}
\usepackage{enumitem}
\usepackage[pagebackref,breaklinks]{hyperref}\hypersetup{pdfborder={0 0 0}}
\usepackage{xspace}
\usepackage{setspace}
\usepackage[skip=3pt,font=small]{subcaption}
\usepackage[skip=3pt,font=small]{caption}
\usepackage[capitalise]{cleveref}
\usepackage{booktabs,tabularx,colortbl,multirow,array,makecell}
\usepackage{overpic,wrapfig}
\usepackage[misc]{ifsym}

\makeatletter
\DeclareRobustCommand\onedot{\futurelet\@let@token\@onedot}
\def\@onedot{\ifx\@let@token.\else.\null\fi\xspace}
\def\eg{\emph{e.g}\onedot} 

\def\ie{\emph{i.e}\onedot}

\def\wrt{w.r.t\onedot}

\makeatother

\makeatletter
\renewcommand{\paragraph}{%
  \@startsection{paragraph}{4}%
  {\z@}{0ex \@plus 0ex \@minus 0ex}{-1em}%
  {\hskip\parindent\normalfont\normalsize\bfseries}%
}
\makeatother

\graphicspath{{figures/}}

\crefname{algorithm}{Alg.}{Algs.}
\Crefname{algocf}{Algorithm}{Algorithms}
\crefname{section}{Sec.}{Secs.}
\Crefname{section}{Section}{Sections}
\crefname{table}{Tab.}{Tabs.}
\Crefname{table}{Table}{Tables}
\crefname{figure}{Fig.}{Fig.}
\Crefname{figure}{Figure}{Figure}

\definecolor{gblue}{HTML}{4285F4}
\definecolor{gred}{HTML}{DB4437}
\definecolor{ggreen}{HTML}{0F9D58}

\definecolor{mygray}{gray}{.92}
\definecolor{emphypurple}{rgb}{0.302, 0.055, 0.659}

\definecolor{highlightgreen}{HTML}{009901}
\definecolor{highlightred}{HTML}{FD6864}

\acrodef{eru}[ERU]{Embodied Reference Understanding}
\acrodef{rec}[REC]{Referring Expression Comprehension}
\acrodef{ewl}[EWL]{Elbow-Wrist Line}
\acrodef{vtl}[VTL]{Virtual Touch Line}
\newcommand{\dataset}{\textit{YouRefIt}\xspace}

\title{Understanding Embodied Reference\\with Touch-Line Transformer}

\author{%
    \bf Yang Li$^{1\,\textrm{\Letter}}$, Xiaoxue Chen$^1$, Hao Zhao$^{2\,\textrm{\Letter}}$,\\
    \bf Jiangtao Gong$^1$, Guyue Zhou$^1$, Federico Rossano$^3$, Yixin Zhu$^{4\,\textrm{\Letter}}$\\
    $^1$ Institute for AI Industry Research, Tsinghua University
    $^2$ Intel Labs China \& Peking University\\
    $^3$ Department of Cognitive Science, UCSD
    $^4$ Institute for Artificial Intelligence, Peking University\\
    $\textrm{\Letter}$\phantom\,\,\texttt{liyang@air.tsinghua.edu.cn, hao.zhao@intel.com,}\\
    \quad{}\phantom\,\,\texttt{zhao-hao@pku.edu.cn, yixin.zhu@pku.edu.cn}\\
}

\iclrfinalcopy
\begin{document}
\maketitle

\begin{abstract}
We study embodied reference understanding, the task of locating referents using embodied gestural signals and language references. Human studies have revealed that objects referred to or pointed to do not lie on the \textit{elbow-wrist line}, a common misconception; instead, they lie on the so-called \textit{virtual touch line}. However, existing human pose representations fail to incorporate the virtual touch line. To tackle this problem, we devise the touch-line transformer: It takes as input tokenized visual and textual features and simultaneously predicts the referent's bounding box and a touch-line vector. Leveraging this touch-line prior, we further devise a geometric consistency loss that encourages the co-linearity between referents and touch lines. Using the touch-line as gestural information improves model performances significantly. Experiments on the \dataset dataset show our method achieves a +25.0\% accuracy improvement under the 0.75 IoU criterion, closing 63.6\% of the gap between model and human performances. Furthermore, we computationally verify prior human studies by showing that computational models more accurately locate referents when using the \textit{virtual touch line} than when using the \textit{elbow-wrist line}.
\end{abstract}

\section{Introduction}

\begin{wrapfigure}{r}{.5\linewidth}
    \centering
    \vspace{-6pt}
    \includegraphics[width=\linewidth]{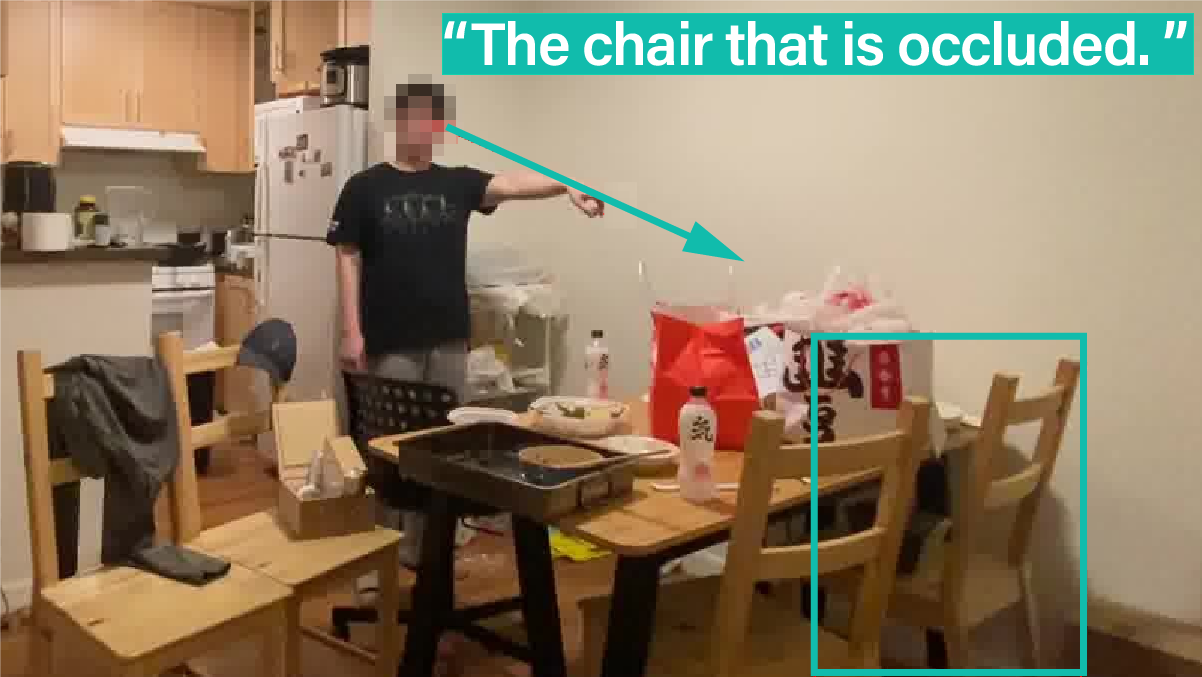}
    \caption{\textbf{Both nonverbal and verbal expressions are necessary to properly locate the referent in complex scenes.} Without nonverbal expression (the pointing gesture in this case), the verbal expression (``the chair'') cannot uniquely refer to \textit{the} chair because multiple chairs are present in this scenario. If only nonverbal expression were used, one cannot distinguish the intended referent ``the chair'' from other objects in that region.}
    \label{fig:chairs}
\end{wrapfigure}

Understanding human intents is critical to intelligent robots when interacting with humans. However, most prior work in the modern learning community neglects the multimodal aspect in human-robot communication scenarios. Let us take an example shown in \cref{fig:chairs}, wherein a person commands the robot to interact with a chair behind the table. In response, the robot ought to understand what is being referred to before taking any actions (\eg, moving towards it and cleaning it). Notably, both \textbf{embodied} gesture signals and language \textbf{reference} play significant roles. Without the pointing gesture, the robot would not be able to distinguish the two chairs using the utterance \textit{the chair that is occluded}. Without the language expression, the robot would not be able to distinguish the chair from other objects in that region (\eg, bags on the table). To fill this gap, we study the \ac{eru} task introduced by \citet{chen2021yourefit}. This task requires an algorithm to detect the referent (referred object) using (i) an image/video with nonverbal communication signals and (ii) a sentence as a verbal communication signal.

\begin{wrapfigure}{r}{.5\linewidth}
    \vspace{-6pt}
    \centering
    \includegraphics[width=\linewidth]{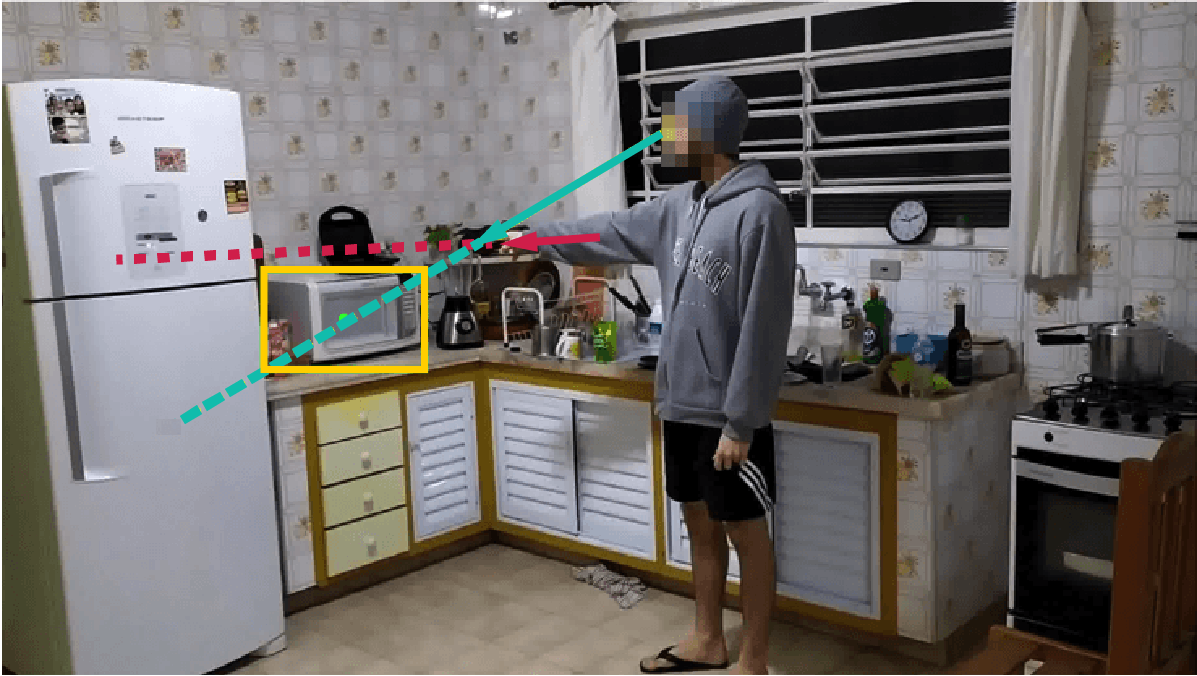}
    \caption{\textbf{\acf{vtl} (in green) vs. \acf{ewl} (in red).} \acp{vtl} affords a more accurate location of referents than \acp{ewl}.     }
    \label{fig:gesture}
\end{wrapfigure}

The \textbf{first} fundamental issue in addressing \ac{eru} is the representation of the human pose. The \textit{de facto} human pose representation in modern computer vision is defined by COCO \citep{lin2014microsoft}---a graph consisting of 17 nodes (keypoints) and 14 edges (keypoint connectivity). This representation is prevalent; The existing model for \ac{eru} \citep{chen2021yourefit} take for granted to use pre-extracted COCO-style pose features as the algorithm input. However, we rethink the limitation of the COCO-style pose graph in the context of \ac{eru} and identify a counter-intuitive fact: The referent does not lie on the elbow-wrist line (\ie, the line that links the human elbow and wrist). As shown in \cref{fig:gesture}, this line (in red) does not cross the referred microwave, illustrating a common misinterpretation of human pointing \citep{herbort2018point}.

Interestingly, a recent psychology study \citep{o2019origin} provides strong evidence that supports the above hypothesis from the behavior perspective. It investigates how human beings mentally develop pointing gestures and argues that it is a virtual form of \textit{reaching out to touch}. This new finding challenges conventional psychological theories \citep{mcginn1981mechanism,kita2003pointing} that the pointing gesture is mentally a behavior of using the limb as an arrow. Inherited from the notations in \citet{o2019origin}, we coin the red line in \cref{fig:gesture} as an \ac{ewl} and the yellow line in \cref{fig:gesture} (\ie, a line that connects the eye and the fingertip) as a \ac{vtl}. Inspired by this essential observation that \acp{vtl} are more accurate than \acp{ewl} in embodied reference, we augment the existing COCO-style pose graph with an edge that connects the eye and the fingertip. As validated in a series of experiments in \cref{sec:exp}, this augmentation brings significant performance boosts on \dataset \citep{chen2021yourefit}.

The \textbf{second} fundamental issue in addressing \ac{eru} is how to model gestural signals and language references jointly. Inspired by the success of multimodal transformers \citep{chen2020uniter,li2020unicoder,tan2019lxmert,lu2019vilbert,kamath2021mdetr} in multimodal tasks \citep{hudson2019gqa,antol2015vqa,zellers2019recognition}, we devise the Touch-Line Transformer, a transformer-based model that takes inputs from both the visual and natural-language modalities. Our Touch-Line Transformer jointly models gestural signals and language references by simultaneously predicting the touch-line vector and the referent's bounding box. To further help our model utilize gestural signals (\ie, the touch-line vector), we integrate a geometric consistency loss to encourage co-linearity between the touch-line and the predicted location of the referent, resulting in significant performance improvements.

Leveraging the above two insights, our proposed method achieves a +25.0\% accuracy improvement under the 0.75 IoU criterion compared with state-of-the-art methods on the \dataset dataset, closing 63.6\% of the gap between model performance and human performance. 

This paper makes four contributions. (i) Inspired by recent behavior studies, we devise a novel human pose representation--\acf{vtl}--in computer vision. (ii) We introduce the Touch-Line Transformer that jointly models nonverbal gestural signals and verbal references. (iii) We encourage our model to utilize the touch line by a novel geometric consistency loss that improves the co-linearity between the touch line and the predicted object. (iv) Our model achieves a new state-of-the-art performance on \ac{eru}, attaining a +25.0\% performance gain under the 0.75 IoU threshold.

\section{Related Work}

\paragraph{Misinterpretation of pointing gestures}

Pointing allows observers and pointers to direct visual attention to establish references in communication. Recent research reveals that observers surprisingly make systematic errors \citep{herbort2016spatial}; they found that, while the pointer produces gesture using \acp{vtl}, observers interpret pointing gestures using the ``arm–finger'' line. The \ac{vtl} mechanism is founded by \citet{o2019origin}: Pointing gestures orient toward their targets as if the pointers were to touch these targets. In neuroscience, gaze effects occur for tasks that require gaze alignment with finger pointing \citep{bedard2008gaze}. Together, the above evidence shows that eye position and gaze direction are significant for understanding pointing. Critically, \citet{herbort2018point} verify that when human observers were instructed to extrapolate the touch-line vector, this systematic misinterpretation during human-human communication is reduced. Inspired by these discoveries, we integrate the touch-line vector to improve the pointing gesture interpretation performance.

\paragraph{Gaze estimation}

The direction of gazing can sometimes help locate referents. Gaze estimation is finding the direction of gaze. \citet{kellnhofer2019gaze360} introduced an efficient way to collect gaze direction data from panoramic cameras in both indoor and outdoor scenes. In addition, they proposed models to perform gaze estimation in the wild. For face-to-face interaction scenarios, \citet{chong2017detecting} and \citet{chong2020detection} proposed approaches to detect whether children are gazing at the adult's eyes when interacting with that adult. \citet{funes2014eyediap} and \citet{zhang2015appearance} created large-scale image datasets for the training and evaluation of gaze estimation models. \citet{krafka2016eye} introduced a dataset and a model for estimating the gaze direction to a screen in real-time on mobile devices with limited computation resources.

\paragraph{Saliency estimation}

People are more likely to look at objects that are visually more salient to them. Saliency estimation predicts how salient each portion of an image is to humans. \citet{itti1998model} took inspiration from resource-constrained primate visual systems and designed a bottom-up model to efficiently locate the most salient regions in an image by considering the colors, intensity, and orientations of the input image. Additionally, when designing bottom-up models, \citet{hou2008dynamic} utilized entropy from information theory, \cite{bruce2005saliency} used a closely related concept of self-information, and \cite{harel2006graph} exploited concepts from graph theory. \citet{erdem2013visual}'s approach took top-down information into account and exploited object-specific features for objects from a wide range of classes. \citet{cheng2014global} analyze regional contrast and improved enabled saliency estimation on more challenging internet images.

\paragraph{Gaze target detection}

The gazing gesture itself can refer to some objects in the absence of language expressions. Gaze target detection is the localization of the objects that people are gazing at. \citet{recasens2015they} and \citet{chong2018connecting} proposed two-state methods that generate saliency maps and estimate gaze directions first before combining them and predicting the gaze target. \citet{tu2022end} designed a one-stage end-to-end method to simultaneously locate human heads and gaze targets for all people in an image. \citet{fang2021dual} designed a three-stage method that predicts 2D and 3D gaze direction, predicts field of views and depth ranges, and then predicts the gaze target. \citet{zhao2020learning} maintains that humans find gaze targets from salient objects on the sight line. They designed a model to simulate the hypothesized human way of finding gaze targets in images and extended its ability to video frames. \citet{li2021looking} extended gaze target detection to 360-degree images. \citet{chong2020detecting} and \citet{recasens2017following} extended gaze target detection into videos, where the gazed object may not be in the same frame as the gazing gesture. 

\paragraph{Referring expression comprehension}

Language expressions can also refer to some objects. Referring expression comprehension is to locate referents in an image using only language expressions. Multiple datasets (\eg, RefCOCO \citep{yu2016modeling}, RefCOCO+ \citep{yu2016modeling}, RefCOCO-g \citep{mao2016generation}, and Guesswhat \citep{de2017guesswhat}) help benchmark models for this task. On methods and models, \citet{du2021visual} devise a transformer-based framework and use the language to guide the extraction of ``discriminative visual features'' when encoding images and sentences. \citet{rohrbach2016grounding} recognize the bi-directional nature between language expressions and objects in images, obviating the need for bounding box annotations. \citet{yu2018mattnet} devise a modular network; three modules respectively attend to subjects, their location in images, and their relationships with nearby objects. While the above works locate referents using language expressions, they rarely use nonverbal gestures produced by humans in the input images.

\paragraph{Human-robot interaction and collaboration}

Understanding referring expressions is beneficial for human-robot interaction. Robots need to understand the object referred to by a human through referring expressions during their interaction with a human to collaborate efficiently with a human in a shared environment \citep{whitney2016interpreting}. Prior robotic systems partly address these difficulties in locating the referents caused by ambiguities in referring expressions through the identification of ambiguities and the raising of accurate and relevant questions after detecting ambiguities \citep{whitney2016interpreting,zhang2021invigorate,pramanick2022talk}. While these robot systems can ask humans for more specific references to an object before doing a task for humans or collaborating with humans, they do not explicitly consider an important signal that humans naturally use when referring to objects: pointing gestures. Incorporating nonverbal pointing gestures may resolve some ambiguities in the first place, obviating the need to ask humans for additional information subsequently. The reduced need in raising some questions makes human-robot interactions and collaborations more efficient in some situations.

\paragraph{Language-conditioned imitation learning}

Imitation helps robots to learn skills \citep{schaal1999imitation}, and robots need to identify target objects when completing goal-conditioned tasks \citep{stepputtis2020language}. \citet{stepputtis2020language} argue that specifying the target object via vectors and one-hot vectors, while helpful for robots to identify the target object, is not flexible enough to support continued learning in deployed robots. As a result, \citet{stepputtis2020language} propose to use the more flexible natural language to refer target objects to robots for goal-conditioned manipulation tasks. Specifically, their architecture contains a semantic module to help robots understand referents referred to by languages. Their architecture helps robots to locate and attend to the target object when the natural language expression non-ambiguously refers to an object. However, their approach does not consider pointing gestures and has difficulties in handling ambiguities when referring involves pointing gestures. As a result, some recent works in language-conditioned imitation learning \citep{stepputtis2020language,lynch2021language,mees2022matters} will potentially benefit from our proposed \ac{vtl} by more accurately locating the target object, especially when the natural language alone is ambiguous. 

\section{Method}


\begin{figure}[b!]
    \centering
    \vspace{+12pt}
    \includegraphics[width=\linewidth]{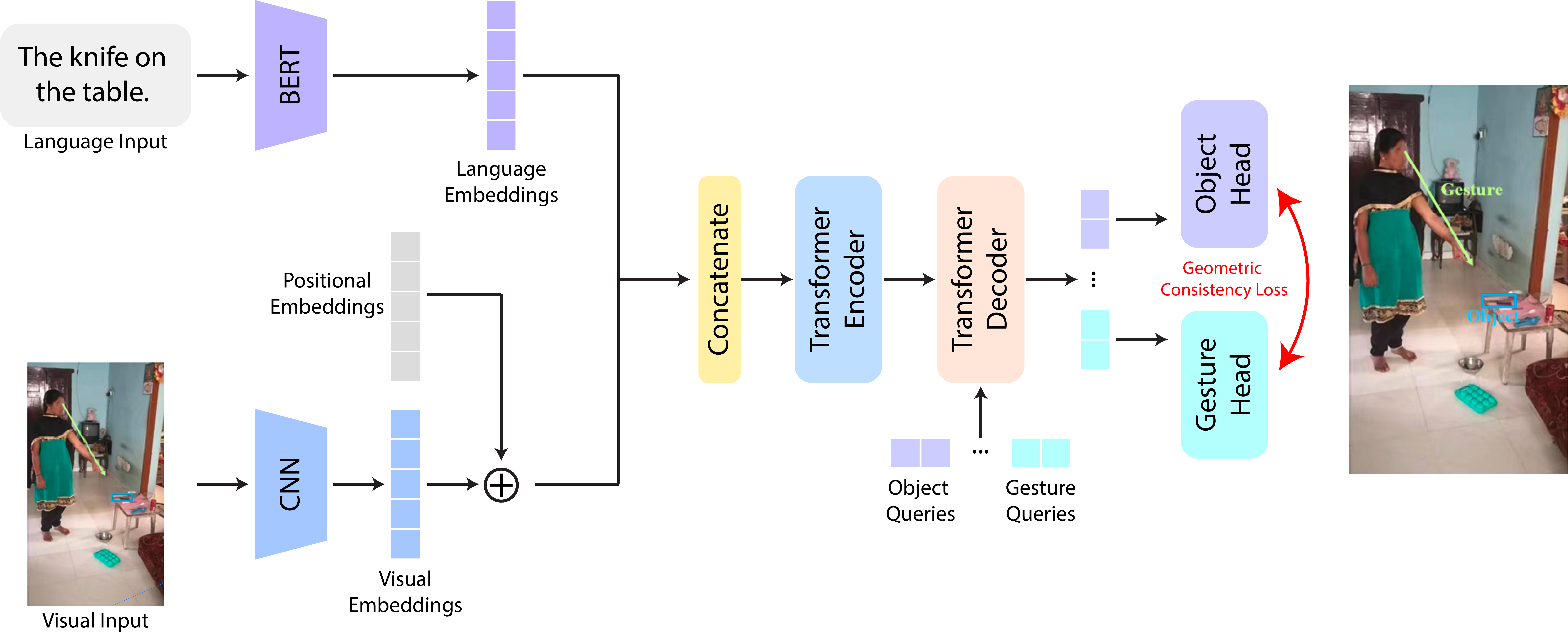}
    \caption{\textbf{Overall network architecture.} Language and visual inputs are first encoded by the text encoder and visual encoder to obtain language and visual embeddings, respectively. Next, these embeddings are concatenated and fed into the transformer encoder to learn multimodal representations. The transformer decoder and prediction heads output the predicted bounding box and \ac{vtl}/\ac{ewl}. A geometric consistency loss is integrated to encourage the use of gestural signals.}
    \label{fig:architecture}
\end{figure}

\subsection{Network Architecture}\label{sec:network}

Our framework, shown in \cref{fig:architecture}, consists of a multi-modal encoder, a transformer decoder, and prediction heads. We detail each component below.

\paragraph{Multimodal encoder}

We generate a visual embedding vector by extracting visual features from input images with a ResNet \citep{he2016deep} backbone,  flattening them, and adding them to a set of position embeddings \citep{parmar2018image, bello2019attention}. Meanwhile, we generate a textural embedding vector from input texts using a pre-trained BERT \citep{liu2019roberta}. After obtaining visual and textural embedding vectors, we concatenate and feed them into a transformer encoder to learn multi-modal representations.

\paragraph{Transformer decoder}

We feed multi-modal representations generated above to our transformer decoder. Our transformer decoder, additionally, takes a set of learnable object queries and gestural key point queries as inputs. With multi-modal representations and input queries, our transformer decoder generates object output embeddings and gestural output embeddings. 

\paragraph{Prediction heads}

Object and gestural output embeddings from the transformer decoder are inputs for our prediction heads (MLPs). Our prediction heads use them to predict bounding boxes (for referents) and gestural key points. We keep one object bounding box and one pair of gestural key points with the highest scores as the final prediction. Specifically, we define the score of a bounding box as $1$ minus the non-object class column of the softmax of the predicted object logits. For a pair of gestural key points, the score is the is-a-\ac{vtl}/\ac{ewl} column of the predicted arm logits' softmax.

\subsection{Explicit Learning of Nonverbal Gestural Signals}\label{sec:explicit}

\paragraph{Measure co-lineaerity}

Ideally, a referent should have a high co-linearity with the \ac{vtl}. We measure this co-linearity using cosine similarity: 
\begin{align}
    cos\_sim = \text{CosineSimilarity}[(x_f-x_e,y_f-y_e),\ (x_o-x_e,y_o-y_e)],
    \label{eq:cos_eye}
\end{align}
where $(x_f, y_f)$, $(x_e, y_e)$, and $(x_o, y_o)$ are the x and y coordinates of the fingertip, the eye, and the center of referent bounding box, respectively. 

\paragraph{Encourage co-lineaerity}

We encourage our model to predict referent bounding boxes that are highly co-linear with \acp{vtl} using a geometric consistency loss:
\begin{align}
    L_{aligned} = \text{ReLU}(cos\_sim_{gt} - cos\_sim_{pred}),
    \label{eq:geo_cosis_loss}
\end{align}
where $cos\_sim_{gt}$ is computed using ground truth referent box and $cos\_sim_{pred}$ is computed using predicted referent box. In both $cos\_sim_{gt}$ and $cos\_sim_{pred}$, ground truth \acp{vtl} are used. Of note, the design of the offset $cos\_sim_{gt}$ ensures the predicted object is as co-linear as the ground-truth object to the \ac{vtl}. In other words, the loss in \cref{eq:geo_cosis_loss} is minimized to zero when the predicted object box is as co-linear as the ground-truth one to the \ac{vtl}. 

\paragraph{Modify for \acp{ewl}}

In experiments using \acp{ewl} instead of \acp{vtl}, we replace the fingertip $(x_f, y_f)$ and the eye $(x_e, y_e)$ in \cref{eq:cos_eye} using the wrist $(x_w, y_w)$ and the elbow $(x_l, y_l)$, respectively. 

\subsection{Implicit Learning of Nonverbal Gestural Signals}\label{sec:implicit}

We eliminate postural signals (by removing humans) from images to investigate whether our model can implicitly learn to utilize useful nonverbal gestural signals without being explicitly asked to learn. Please refer to \cref{appx:inpaint} for details of our procedure.

\subsection{Additional Losses}\label{sec:loss}

The total loss in training is defined as:
\begin{align}
	\mathcal{L}_{\rm total}=\lambda_1 \mathcal{L}_{\rm box} + \lambda_2 \mathcal{L}_{\rm gesture} + \lambda_3 \mathcal{L}_{\rm aligned} + \lambda_4 \mathcal{L}_{\rm token} + \lambda_5 \mathcal{L}_{\rm contrastive},
	\label{eq:total_loss}
\end{align}
where $\lambda_i$s are the weights of various losses, $\mathcal{L}_{\rm box}$ is the weighted sum of the L1 and GIoU losses for predicted referent bounding boxes, and $\mathcal{L}_{\rm gesture}$ is the L1 loss for predicted \acp{vtl} or \acp{ewl}. The soft token loss $\mathcal{L}_{\rm token}$ and the contrastive loss $\mathcal{L}_{\rm contrastive}$ follow those in \citet{kamath2021mdetr}; they help our model align visual and textural signals.

\section{Experiments}\label{sec:exp}

\subsection{Dataset, Evaluation Metric, and Implementation Details}

We use the \dataset dataset \citep{chen2021yourefit}, consisting of $2,950$ training instances and $1,245$ test instances. It has two portions: videos and images. We use the image portion; the inputs to our model are images that do not have temporal information. Each instance contains an image, a sentence, and the location of the referent. We provide additional annotations of \acp{vtl} and \acp{ewl} for the \dataset dataset.

To make a fair comparison with prior models, we follow \citet{chen2021yourefit} and report precision under three different IoU thresholds: 0.25, 0.50, and 0.75. A prediction is correct if its IoU with the ground-truth box is greater than the threshold. Additionally, we choose our best models by the precision at the GIoU \citep{rezatofighi2019generalized} threshold of 0.75; compared with IoU, GIoU is an improved indicator of the model's ability to locate objects accurately. We report the results using GIoU thresholds in \cref{sec:precision_giou}.

During training, we use the Adam optimizer\textbf{}'s AMSGrad variant \citep{reddi2018convergence} and train our models for 200 epochs. We use the AMSGrad variant because we observe a slow convergence of the standard Adam optimizer in experiments. We set the learning rate to 5e-5 except for the text encoder, whose learning rate is 1e-4. We do not perform learning rate drops because we rarely observe demonstrable performance improvements after dropping them. We use A100 GPUs. The sum of the batch sizes on all graphic cards is 56. Augmentations follow those in \citet{kamath2021mdetr}. The total number of queries is 20 (15 for objects; 5 for gestural key points).

\subsection{Comparison with State-of-the-art Methods}

Our approach outperforms prior state-of-the-art methods by 16.4\%, 23.0\%, and 25.0\% under IoU threshold of 0.25, 0.50, and 0.75, respectively; see \cref{tab:comparison}. Specifically, our model performs better than visual grounding methods \citep{yang2019fast,yang2020improving} which do not explicitly utilize nonverbal gestural signals. Our method also performs better than the method proposed in \dataset \citep{chen2021yourefit}, which did not leverage the touch line or the transformer models for multimodal tasks.

\begin{wraptable}{r}{0.65\linewidth}
    \vspace{-6pt}
    \caption{\textbf{Comparison with the state-of-the-art methods.}}
    \label{tab:comparison}
    \centering
    \small
    \resizebox{\linewidth}{!}{
        \begin{tabular}{llll}
            \toprule
            & IoU=.25     & IoU=.50 & IoU=.75 \\
            \midrule
            FAOA \citep{yang2019fast} & 44.5 & 30.4 & 8.5 \\
            ReSC \citep{yang2020improving} & 49.2 & 34.9 & 10.5 \\
            YouRefIT PAF-only \citep{chen2021yourefit} & 52.6 & 37.6 & 12.7 \\
            YouRefIt Full \citep{chen2021yourefit} & 54.7 & 40.5 & 14.0 \\
            \midrule
            Ours (Inpainting) & 59.1 \textcolor[RGB]{200,0,0}{(+4.4)} & 51.3 \textcolor[RGB]{200,0,0}{(+10.8)} & 32.4 \textcolor[RGB]{200,0,0}{(+18.4)}   \\
            Ours (No Pose) & 64.9 \textcolor[RGB]{200,0,0}{(+10.2)} & 57.4 \textcolor[RGB]{200,0,0}{(+16.9)} & 37.2 \textcolor[RGB]{200,0,0}{(+23.2)}   \\
            Ours (\ac{ewl}) & 69.5 \textcolor[RGB]{200,0,0}{(+14.8)} & 60.7 \textcolor[RGB]{200,0,0}{(+20.2)} & 35.5 \textcolor[RGB]{200,0,0}{(+21.5)}   \\
            Ours (\ac{vtl}) & 71.1 \textcolor[RGB]{200,0,0}{(+16.4)} & 63.5 \textcolor[RGB]{200,0,0}{(+23.0)} & 39.0 \textcolor[RGB]{200,0,0}{(+25.0)}   \\\midrule
            Human & 94.2 & 85.8 & 53.3\\
            \bottomrule
        \end{tabular}%
    }%
\end{wraptable}

\begin{figure}[h!]
    \centering
    \vspace{+18pt}
    \begin{subfigure}[b]{0.333\linewidth}
        \centering
        \begin{overpic}
            [width=.5\linewidth]{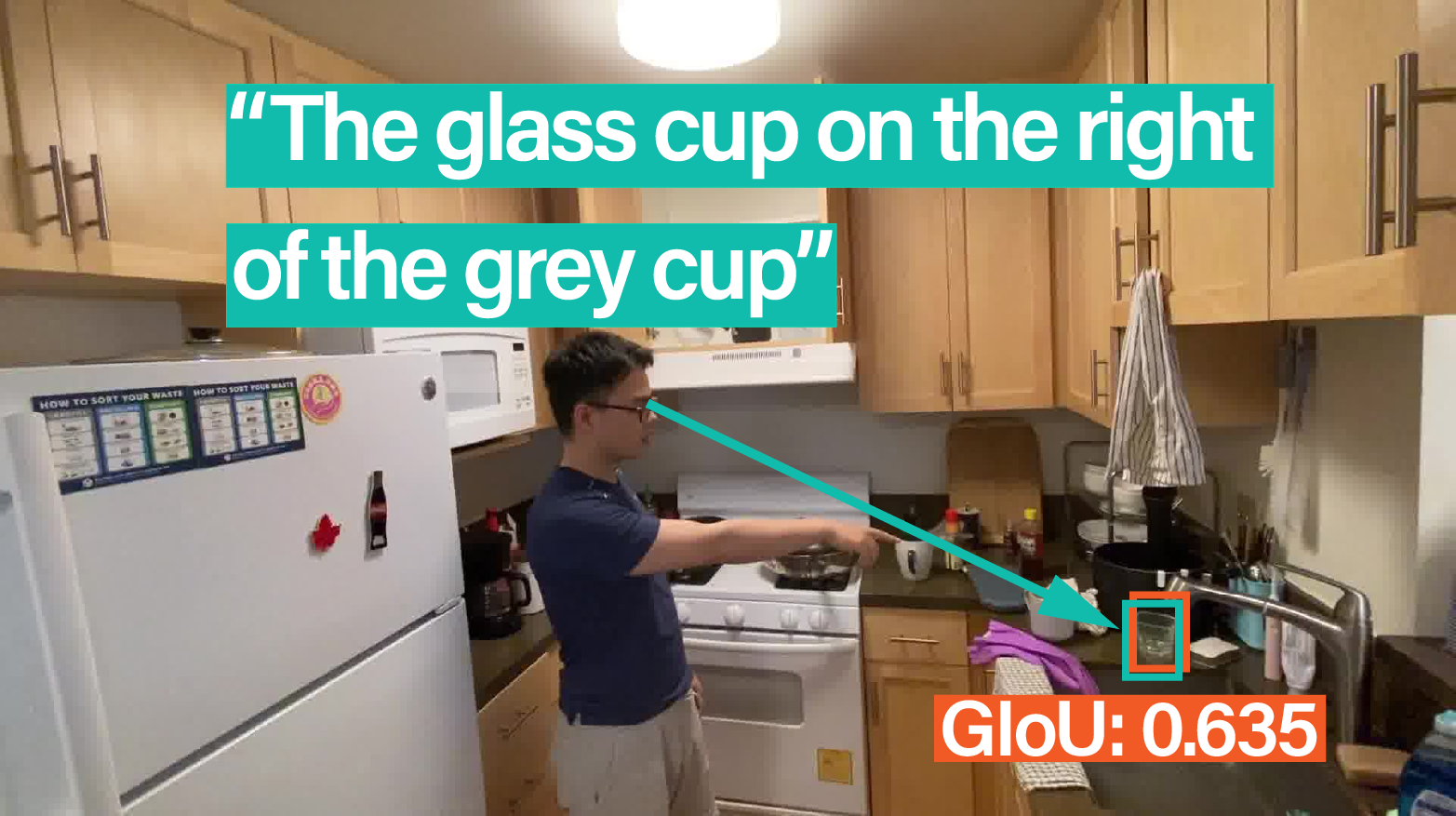}
            \put(35,60){\color{black}\acp{vtl}}
        \end{overpic}%
        \begin{overpic}
            [width=.5\linewidth]{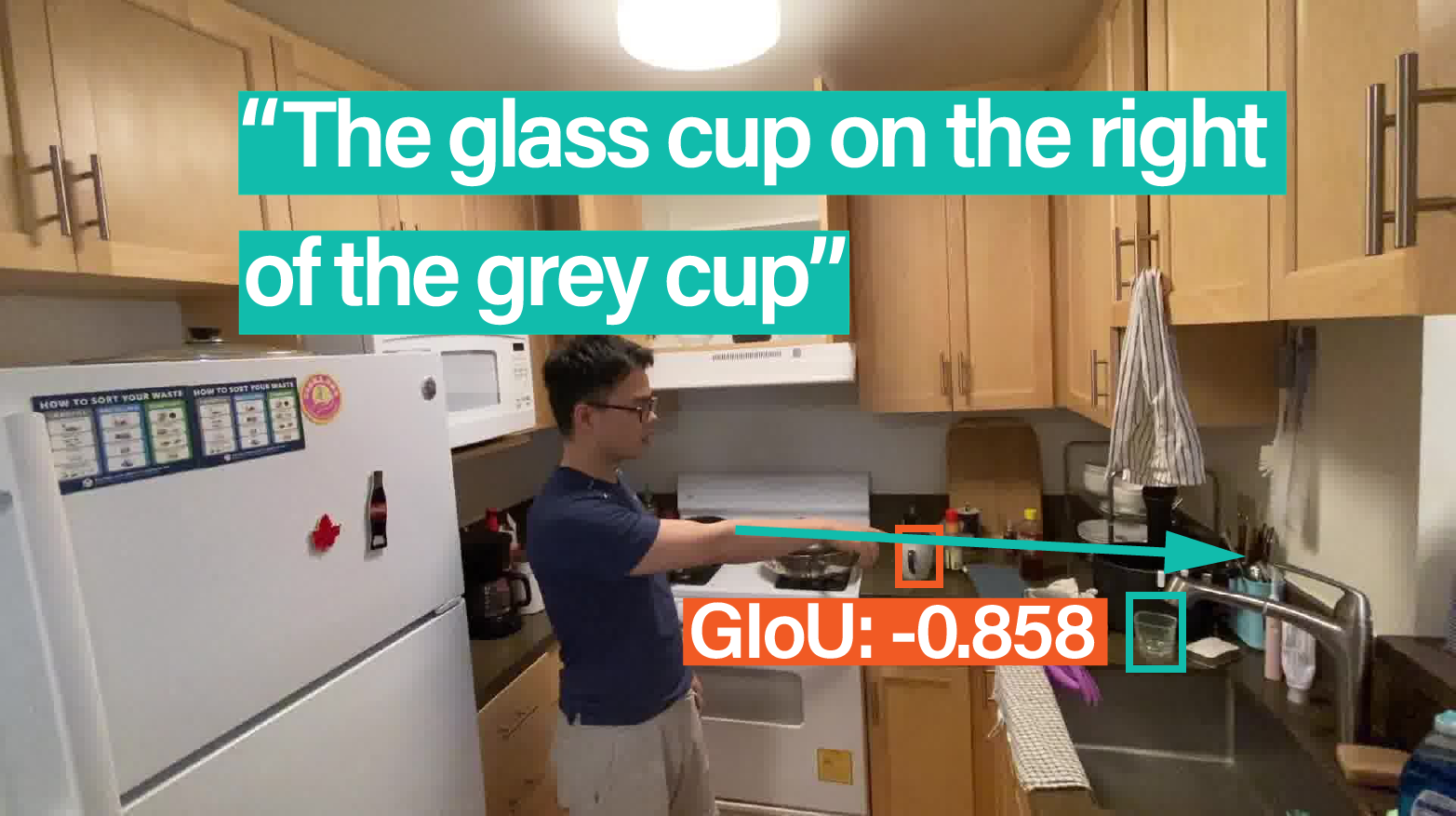}
            \put(35,60){\color{black}\acp{ewl}}
        \end{overpic}%
        \caption{}
        \label{fig:improve:a}
    \end{subfigure}%
    \begin{subfigure}[b]{0.333\linewidth}
        \centering
        \begin{overpic}
            [width=.5\linewidth]{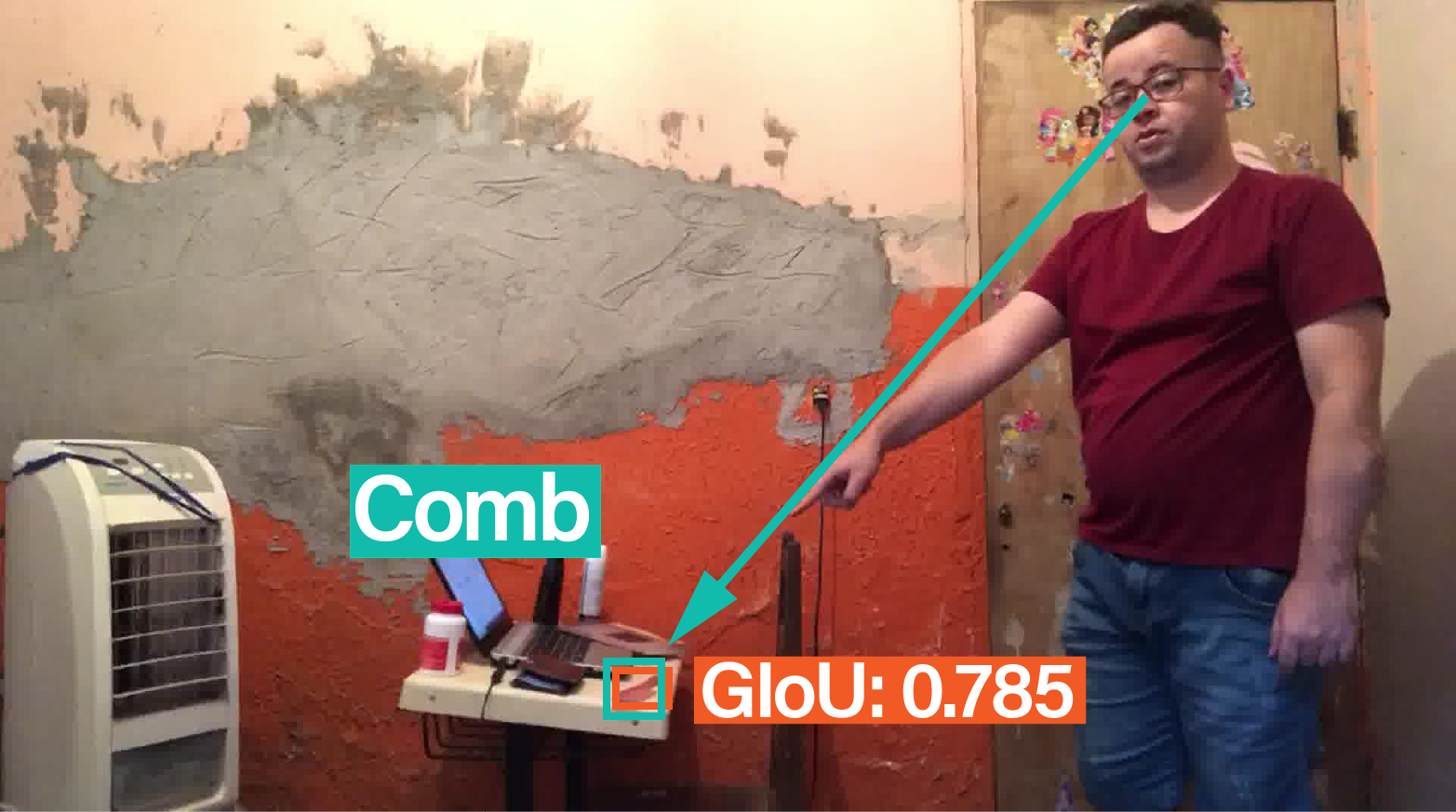}
            \put(35,60){\color{black}\acp{vtl}}
        \end{overpic}%
        \begin{overpic}
            [width=.5\linewidth]{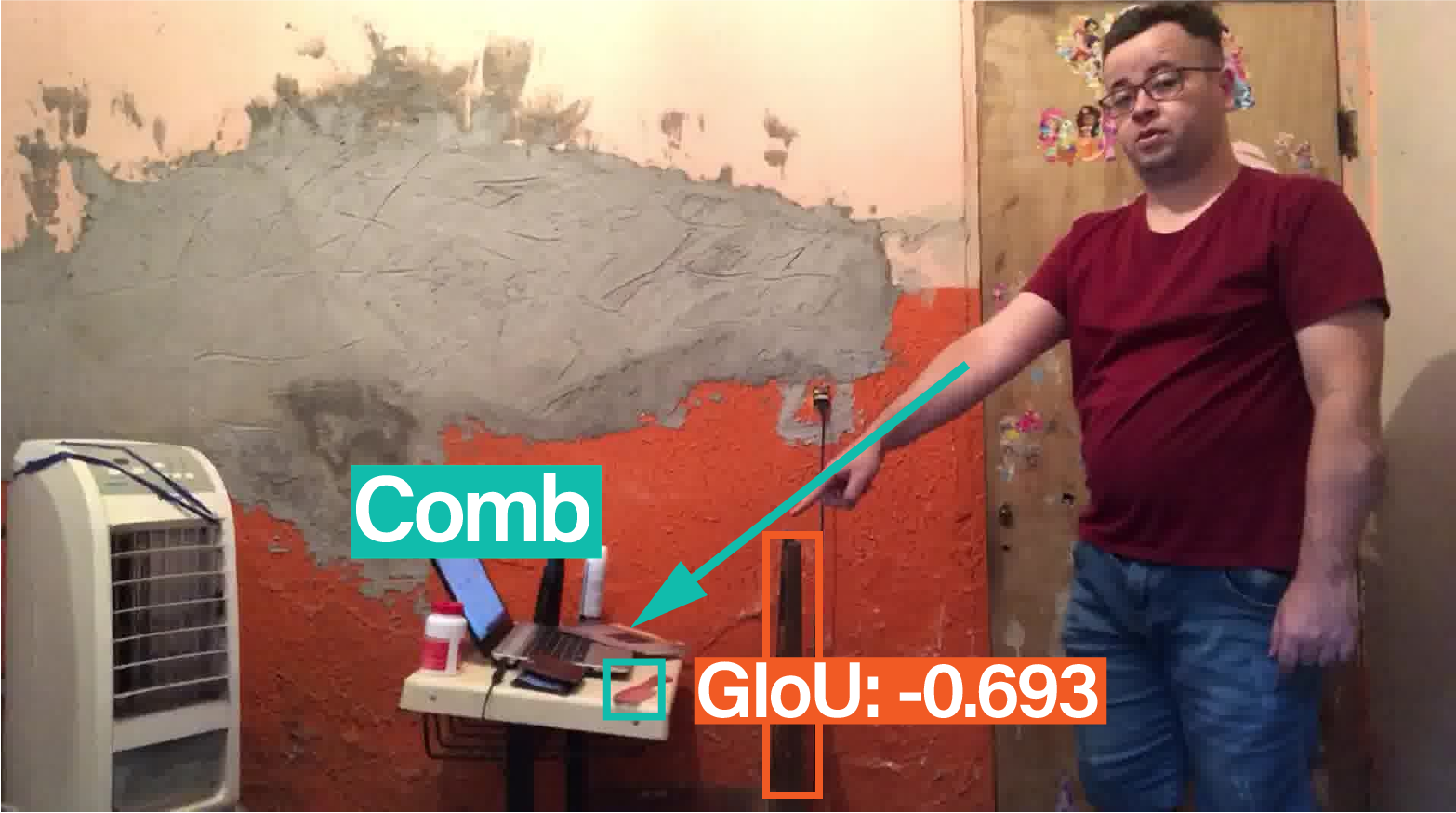}
            \put(35,60){\color{black}\acp{ewl}}
        \end{overpic}%
        \caption{}
        \label{fig:improve:b}
    \end{subfigure}%
    \begin{subfigure}[b]{0.333\linewidth}
        \centering
        \begin{overpic}
            [width=.5\linewidth]{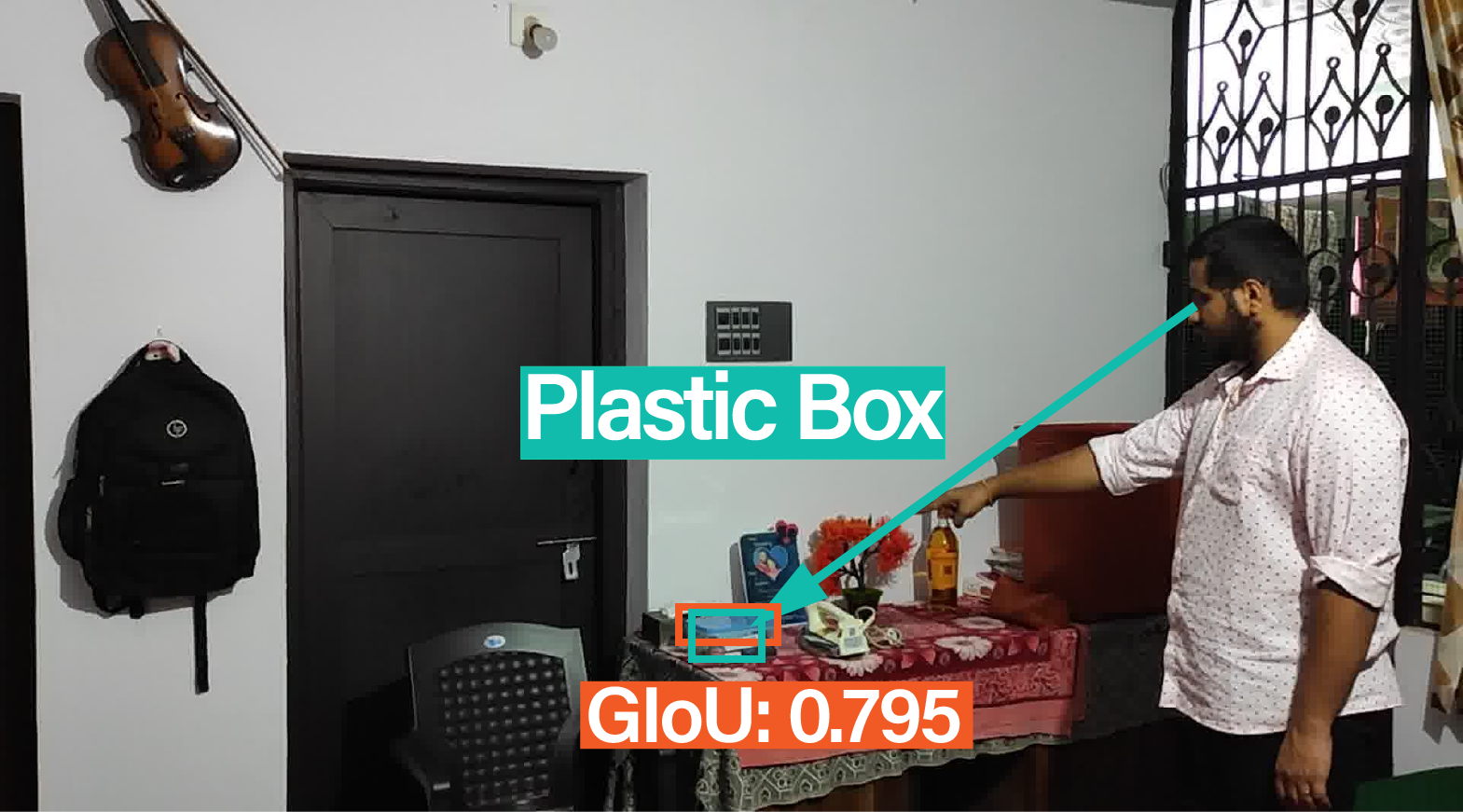}
            \put(35,60){\color{black}\acp{vtl}}
        \end{overpic}%
        \begin{overpic}
            [width=.5\linewidth]{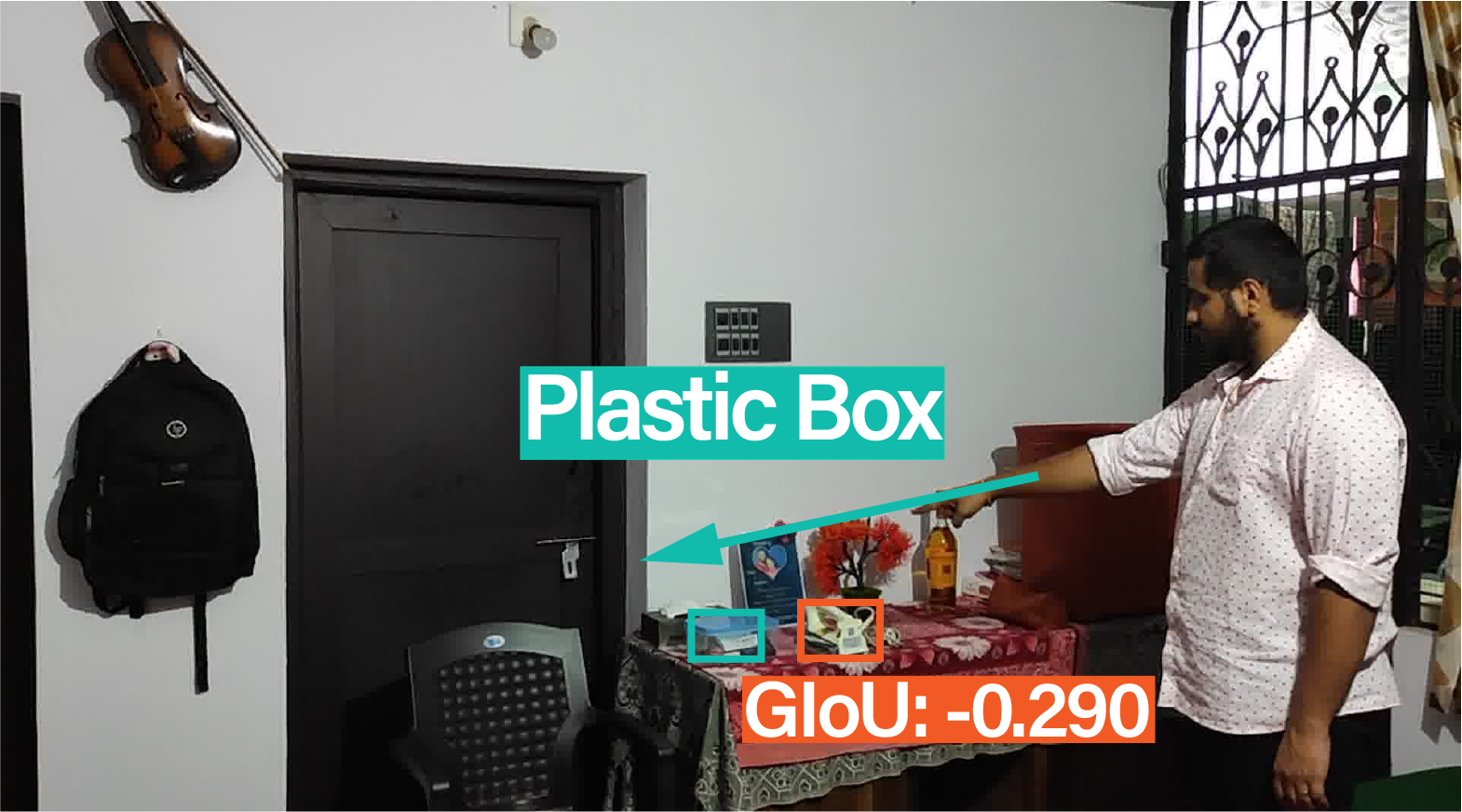}
            \put(35,60){\color{black}\acp{ewl}}
        \end{overpic}%
        \caption{}
        \label{fig:improve:c}
    \end{subfigure}%
    \\%
    \begin{subfigure}[b]{0.333\linewidth}
        \centering
        \includegraphics[width=.5\linewidth]{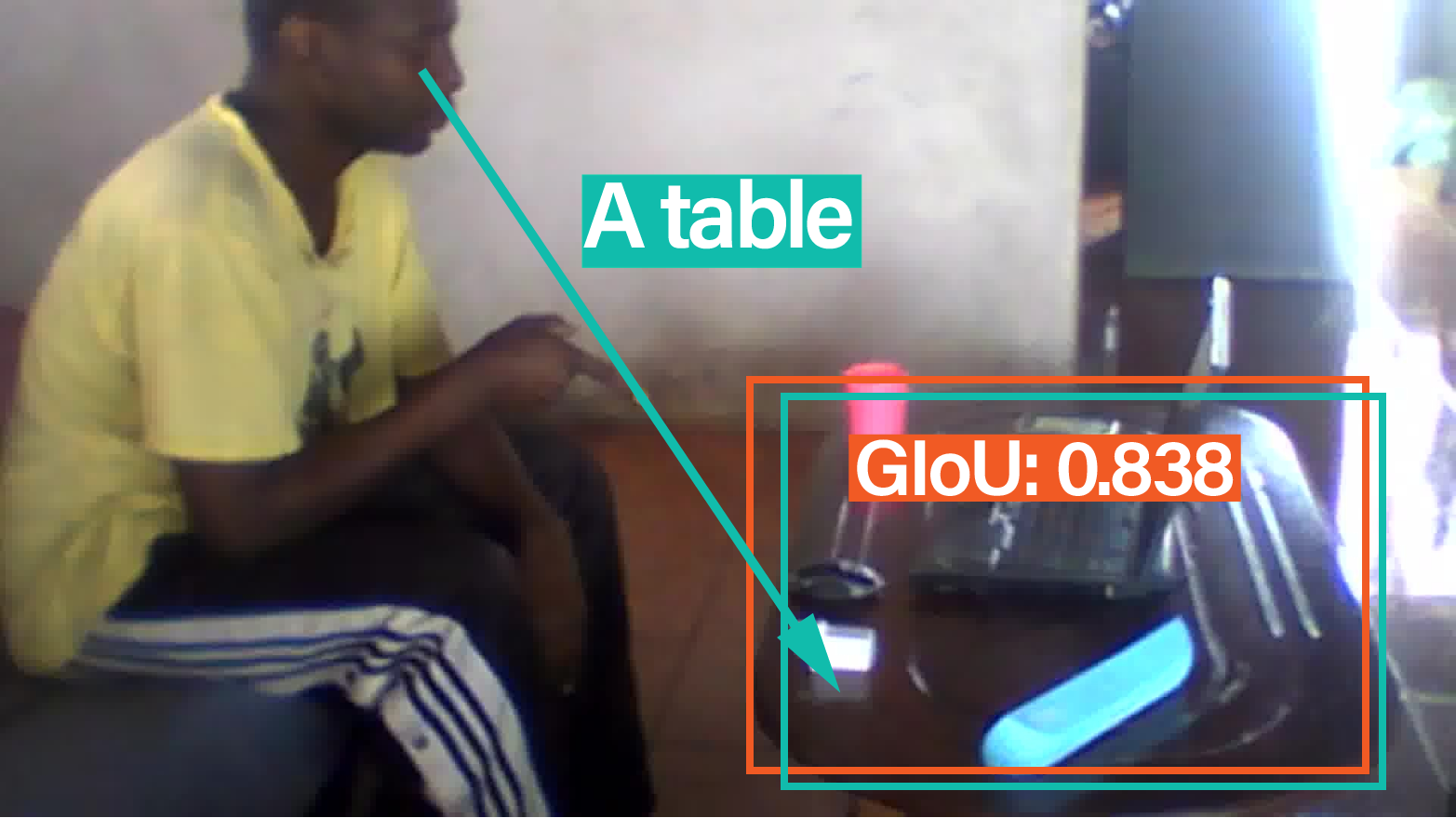}%
        \includegraphics[width=.5\linewidth]{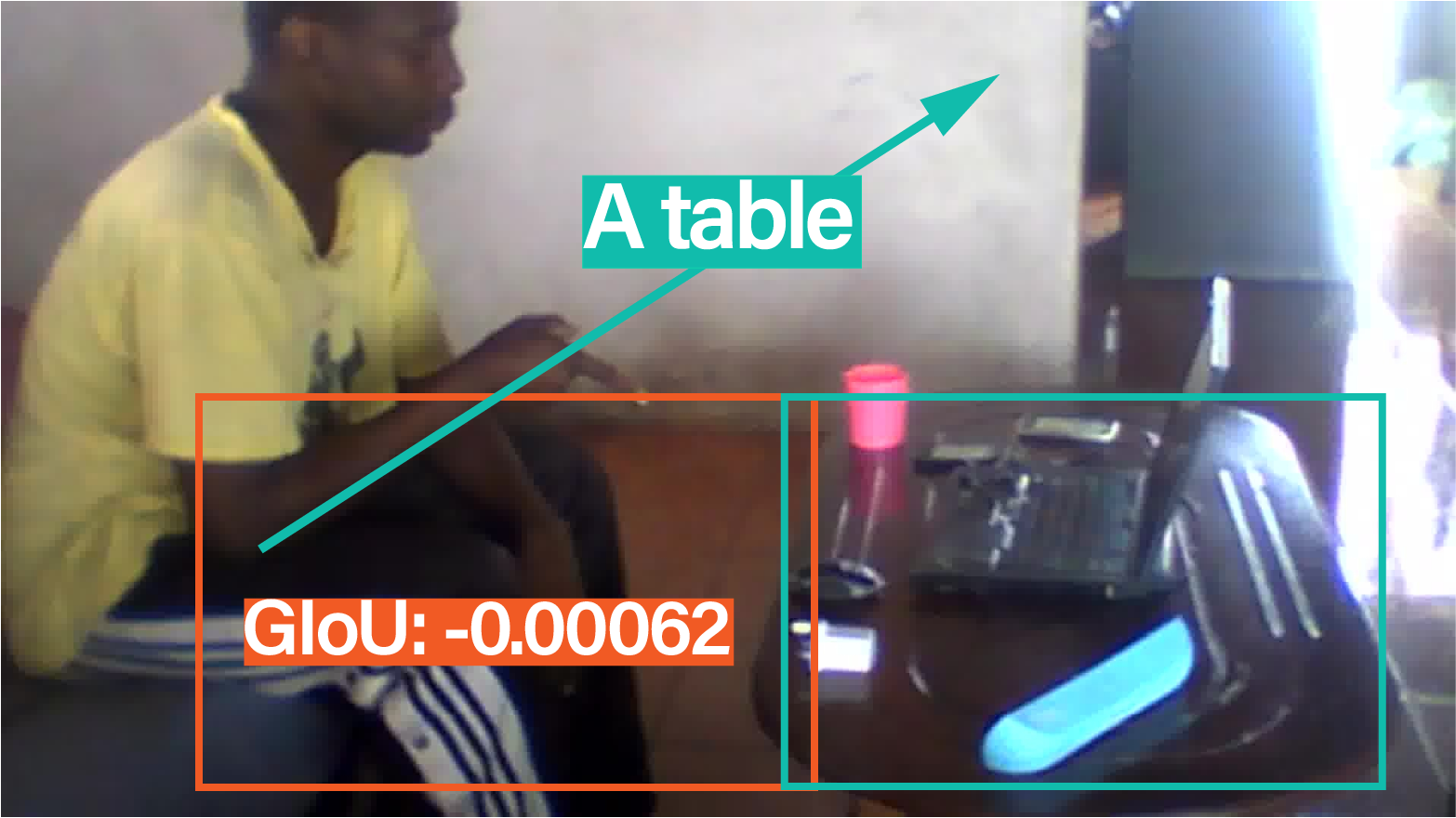}%
        \caption{}
        \label{fig:improve:d}
    \end{subfigure}%
    \begin{subfigure}[b]{0.333\linewidth}
        \centering
        \includegraphics[width=.5\linewidth]{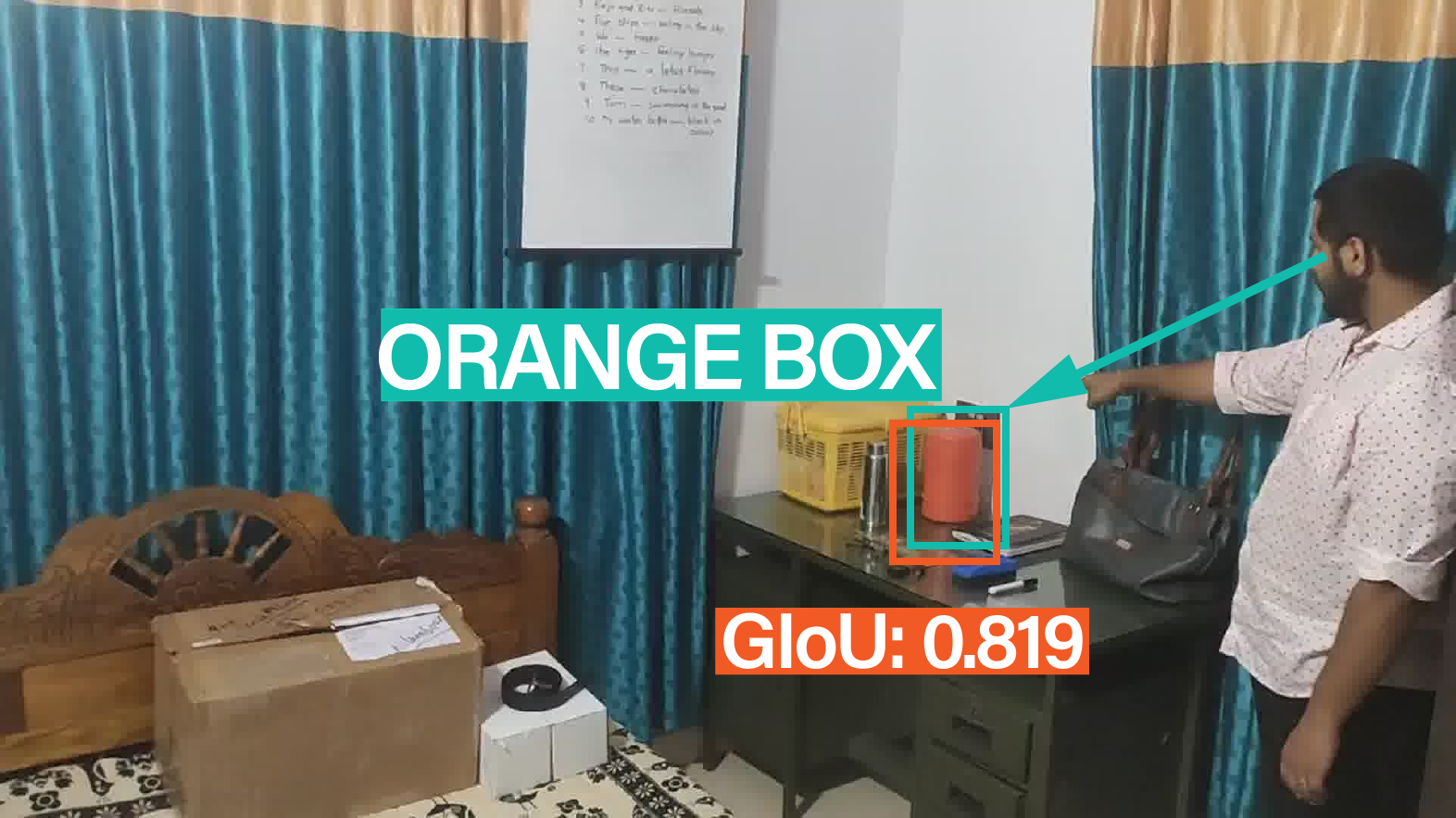}%
        \includegraphics[width=.5\linewidth]{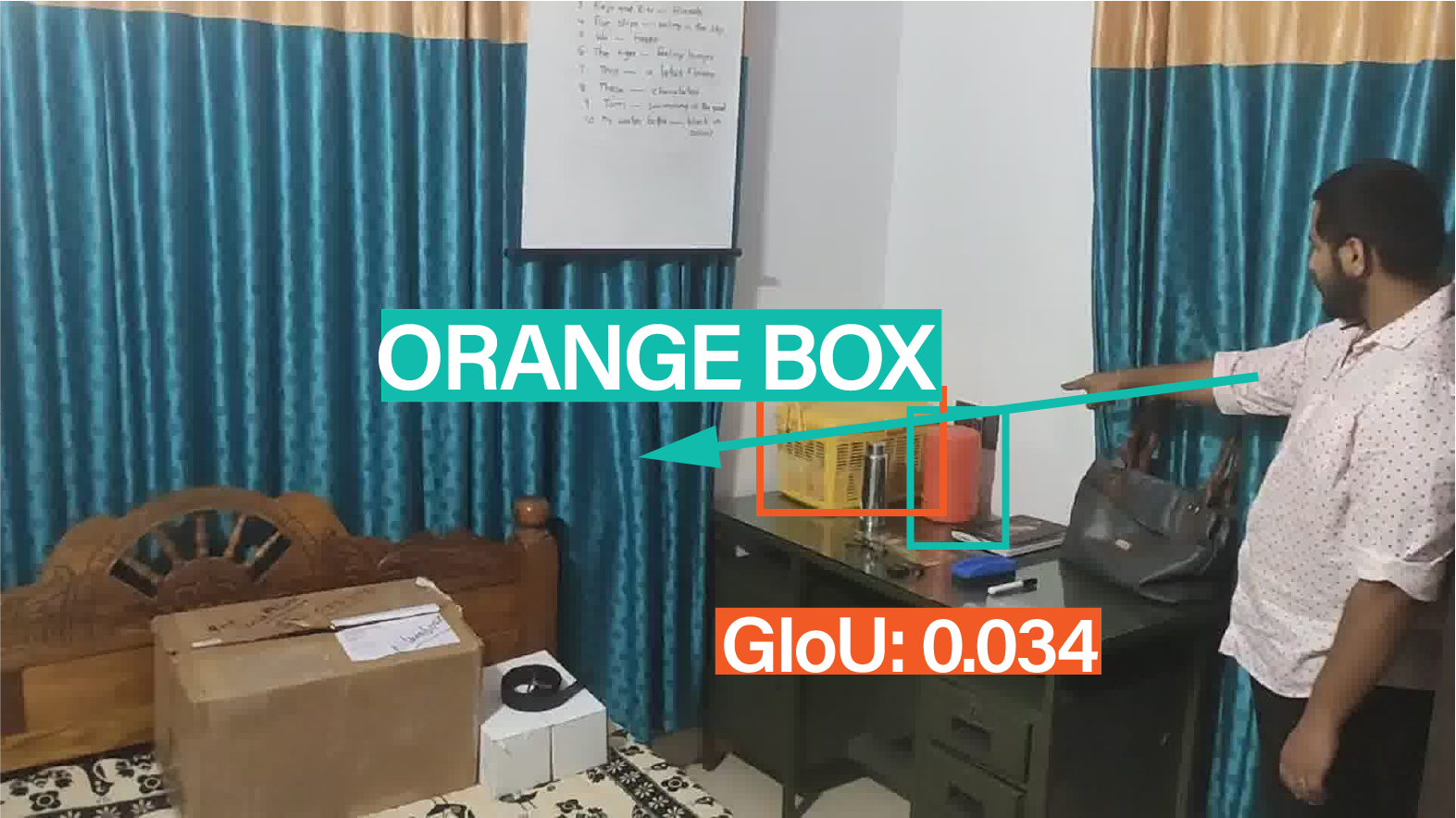}%
        \caption{}
        \label{fig:improve:e}
    \end{subfigure}%
    \begin{subfigure}[b]{0.333\linewidth}
        \centering
        \includegraphics[width=.5\linewidth]{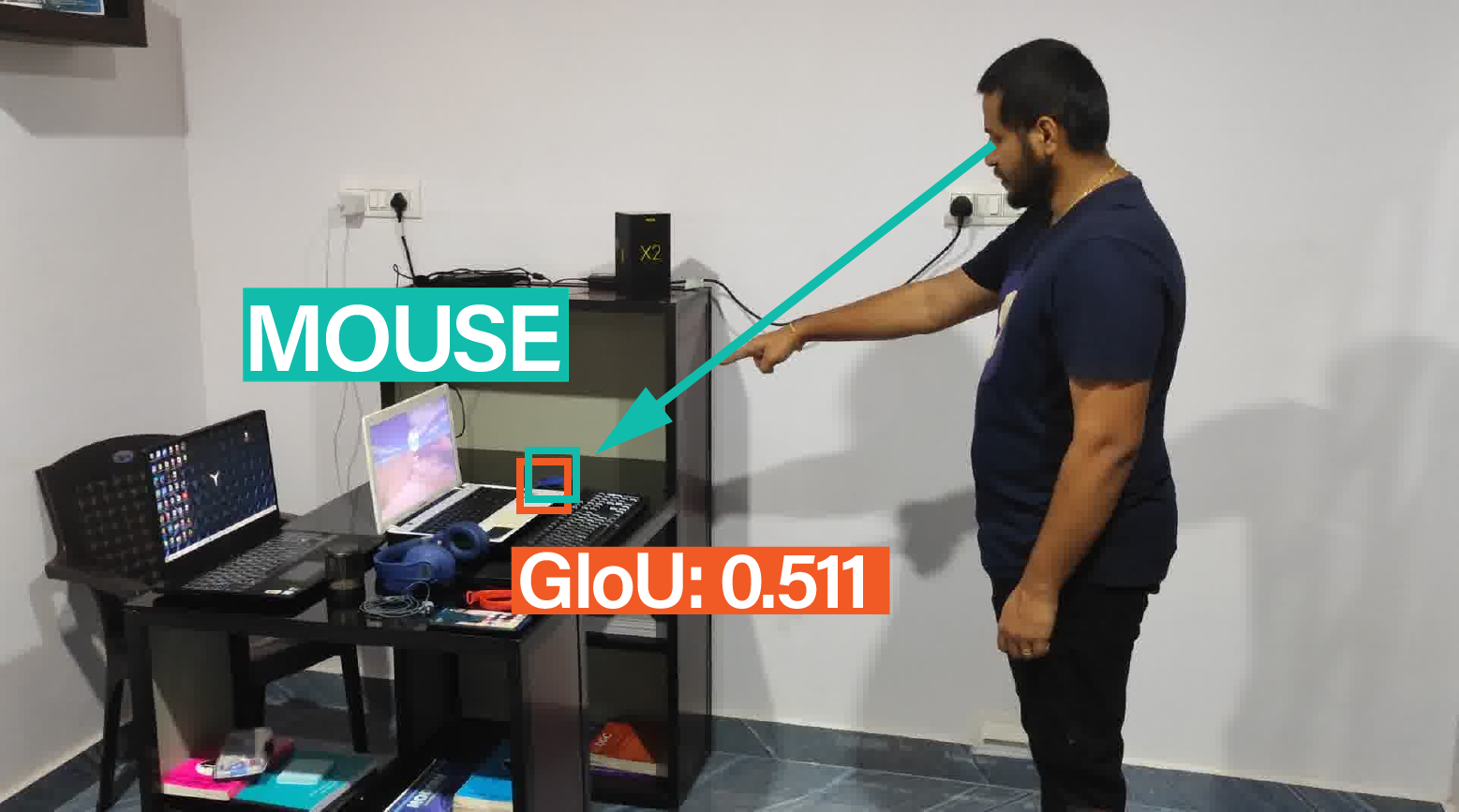}%
        \includegraphics[width=.5\linewidth]{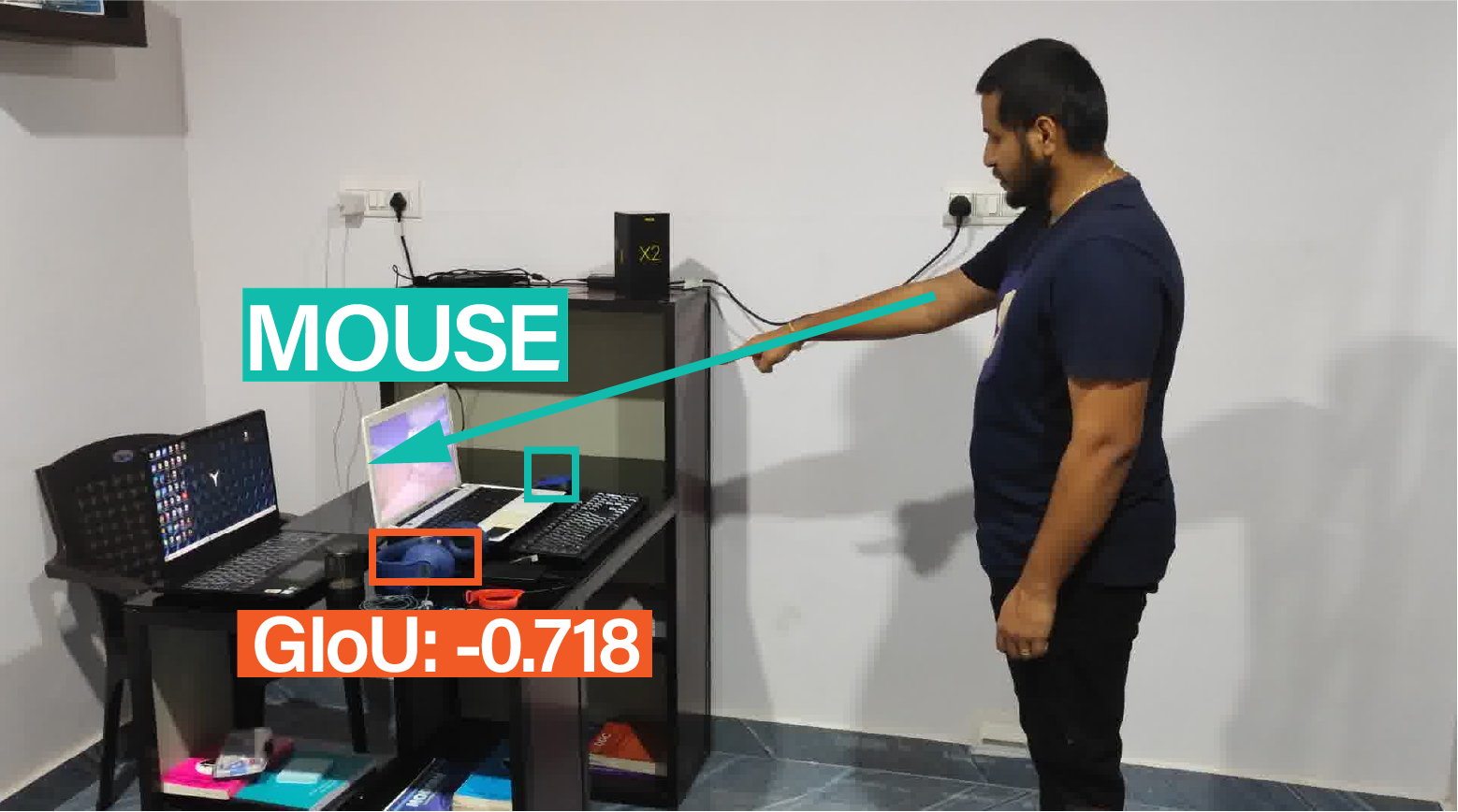}%
        \caption{}
        \label{fig:improve:f}
    \end{subfigure}%
    \\%
    \begin{subfigure}[b]{0.333\linewidth}
        \centering
        \includegraphics[width=.5\linewidth]{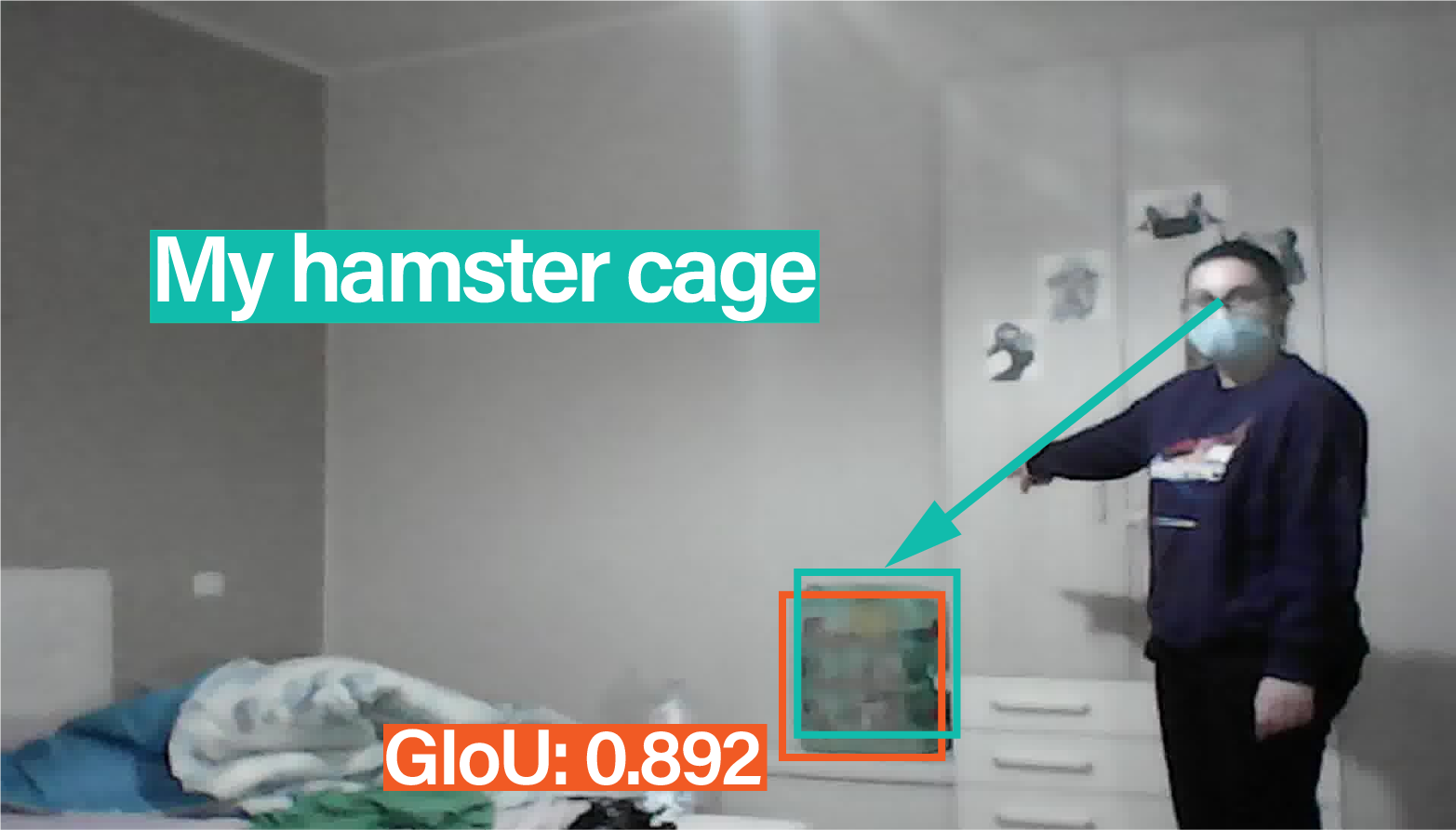}%
        \includegraphics[width=.5\linewidth]{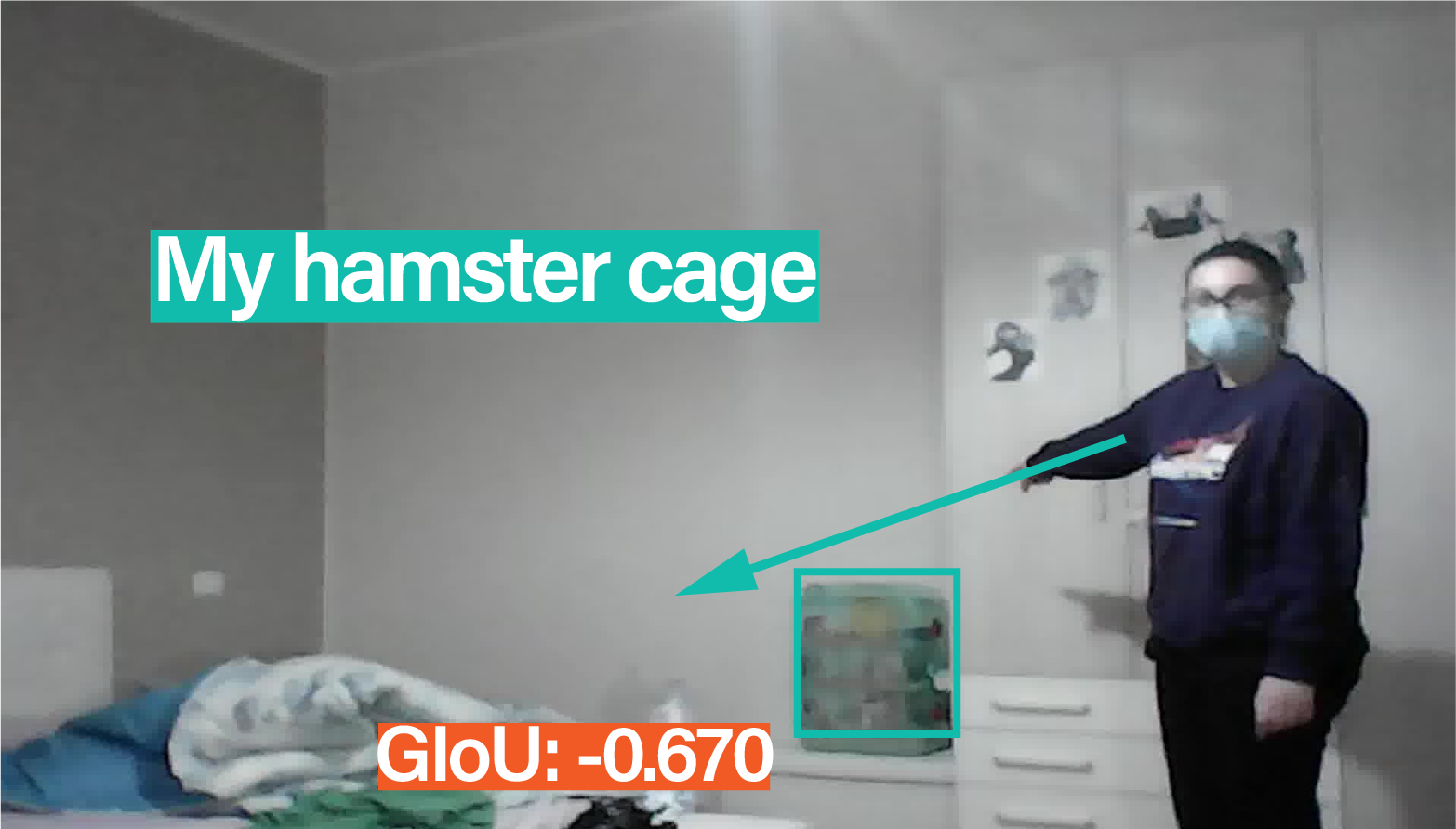}%
        \caption{}
        \label{fig:improve:g}
    \end{subfigure}%
    \begin{subfigure}[b]{0.333\linewidth}
        \centering
        \includegraphics[width=.5\linewidth]{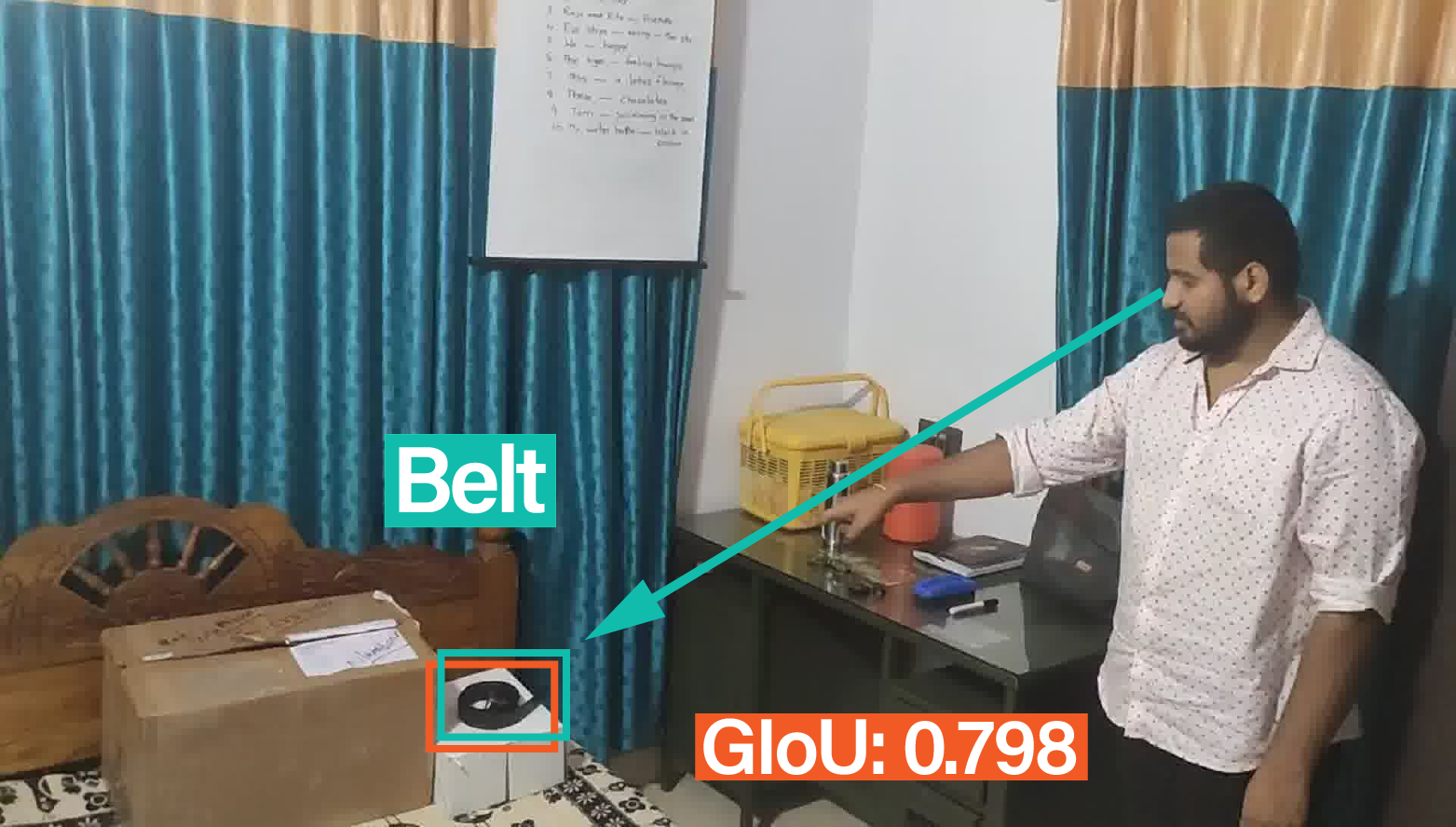}%
        \includegraphics[width=.5\linewidth]{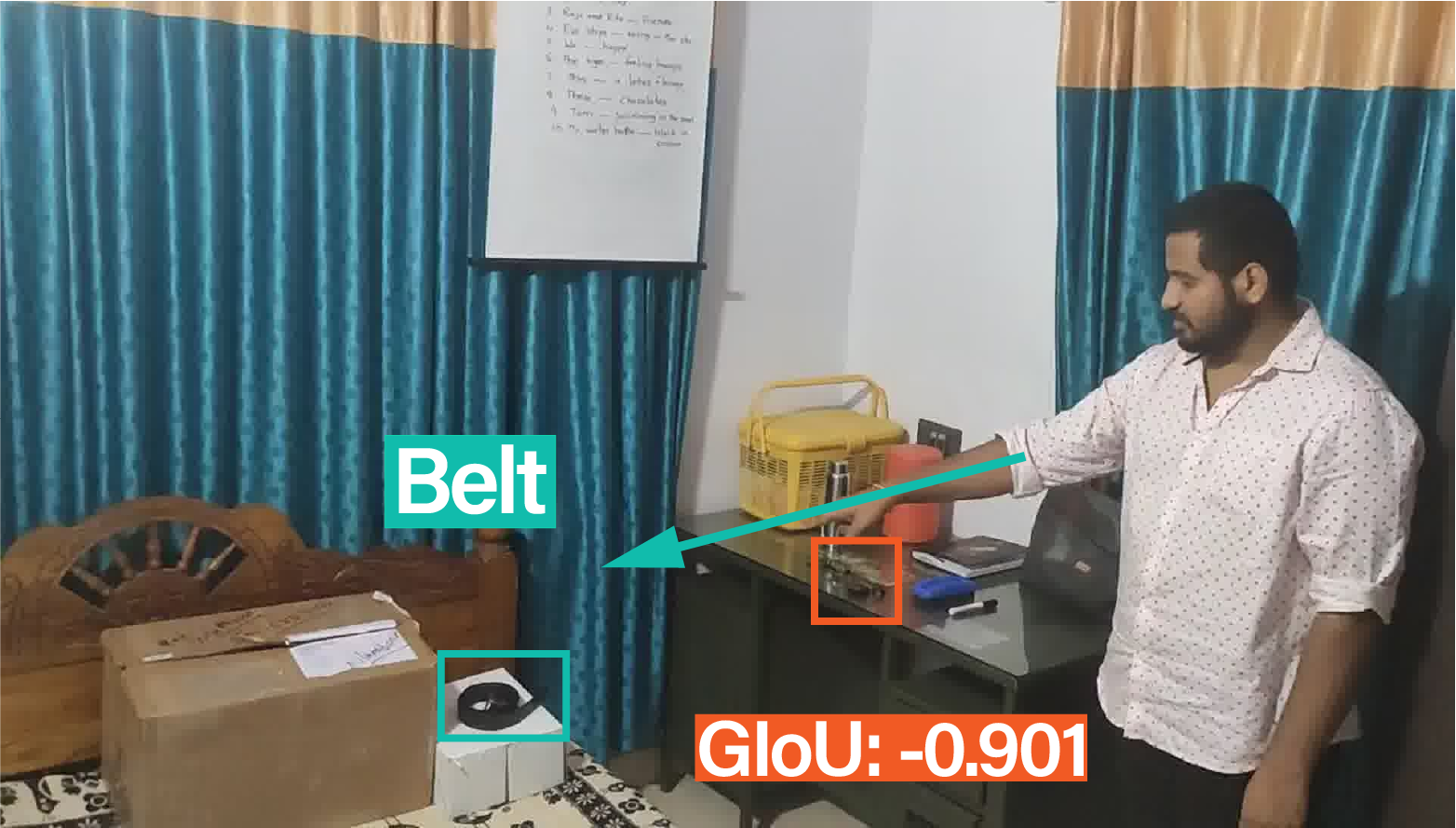}%
        \caption{}
        \label{fig:improve:h}
    \end{subfigure}%
    \begin{subfigure}[b]{0.333\linewidth}
        \centering
        \includegraphics[width=.5\linewidth]{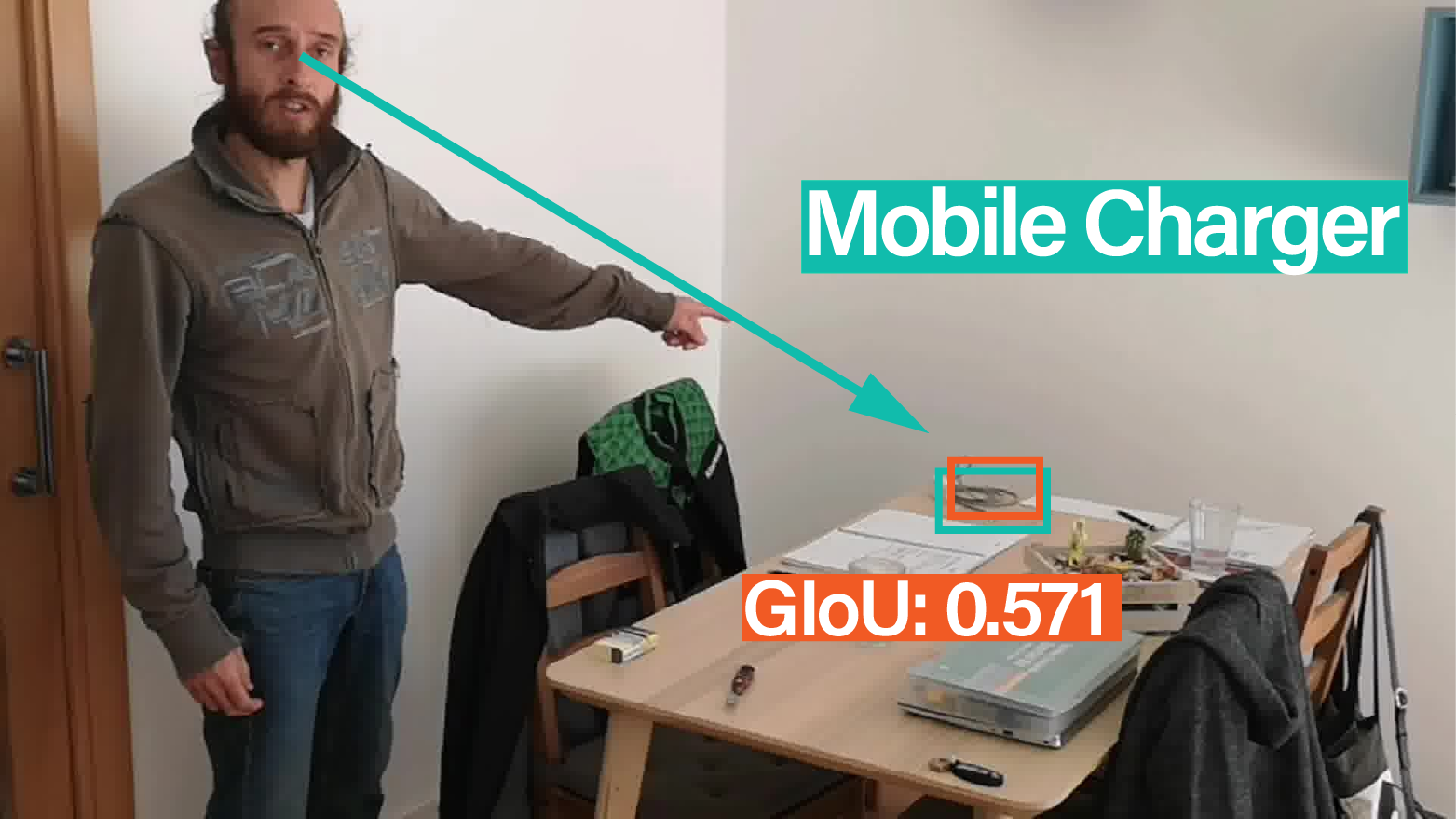}%
        \includegraphics[width=.5\linewidth]{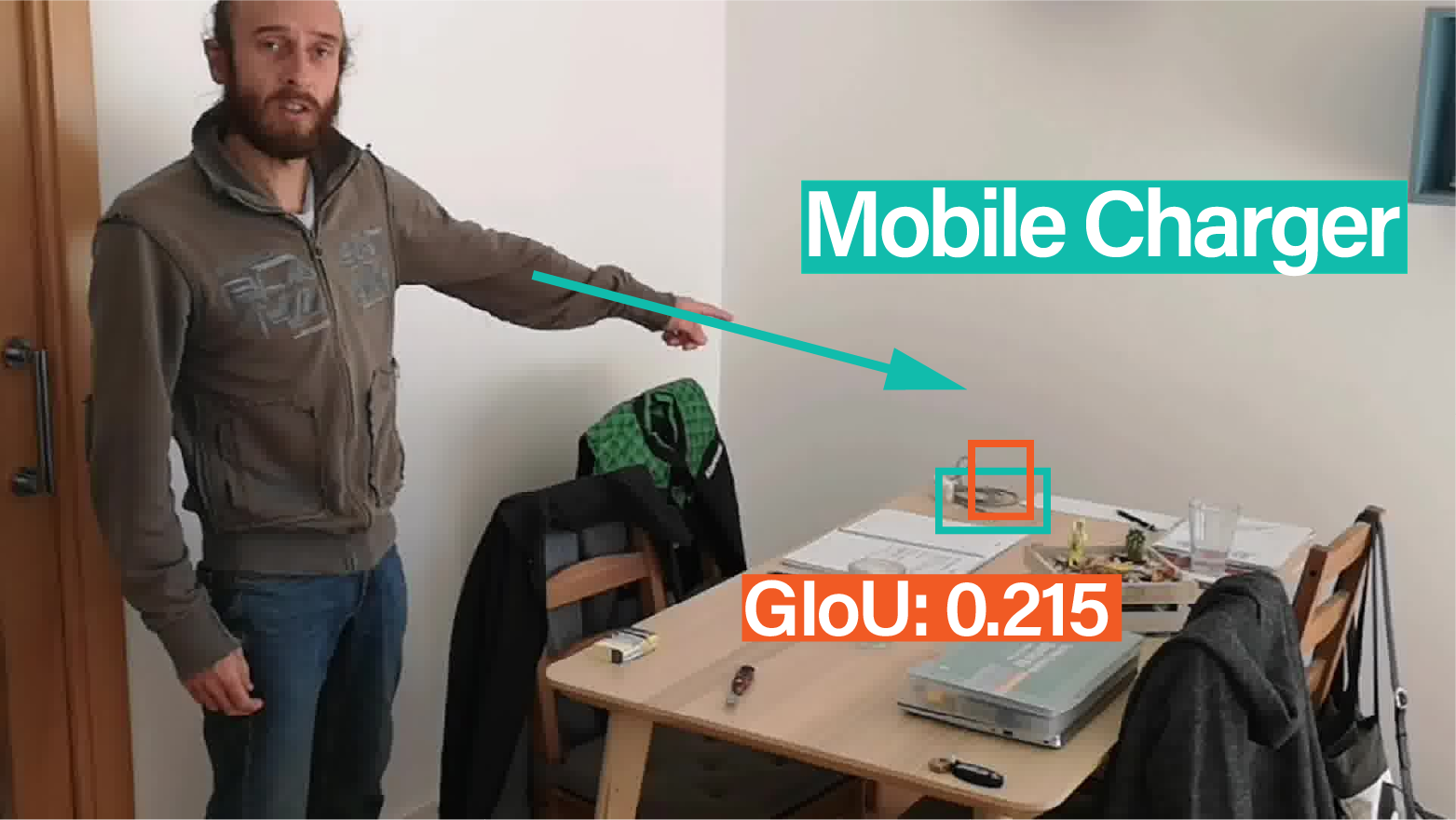}%
        \caption{}
        \label{fig:improve:i}
    \end{subfigure}%
    \\%
    \begin{subfigure}[b]{0.333\linewidth}
        \centering
        \includegraphics[width=.5\linewidth]{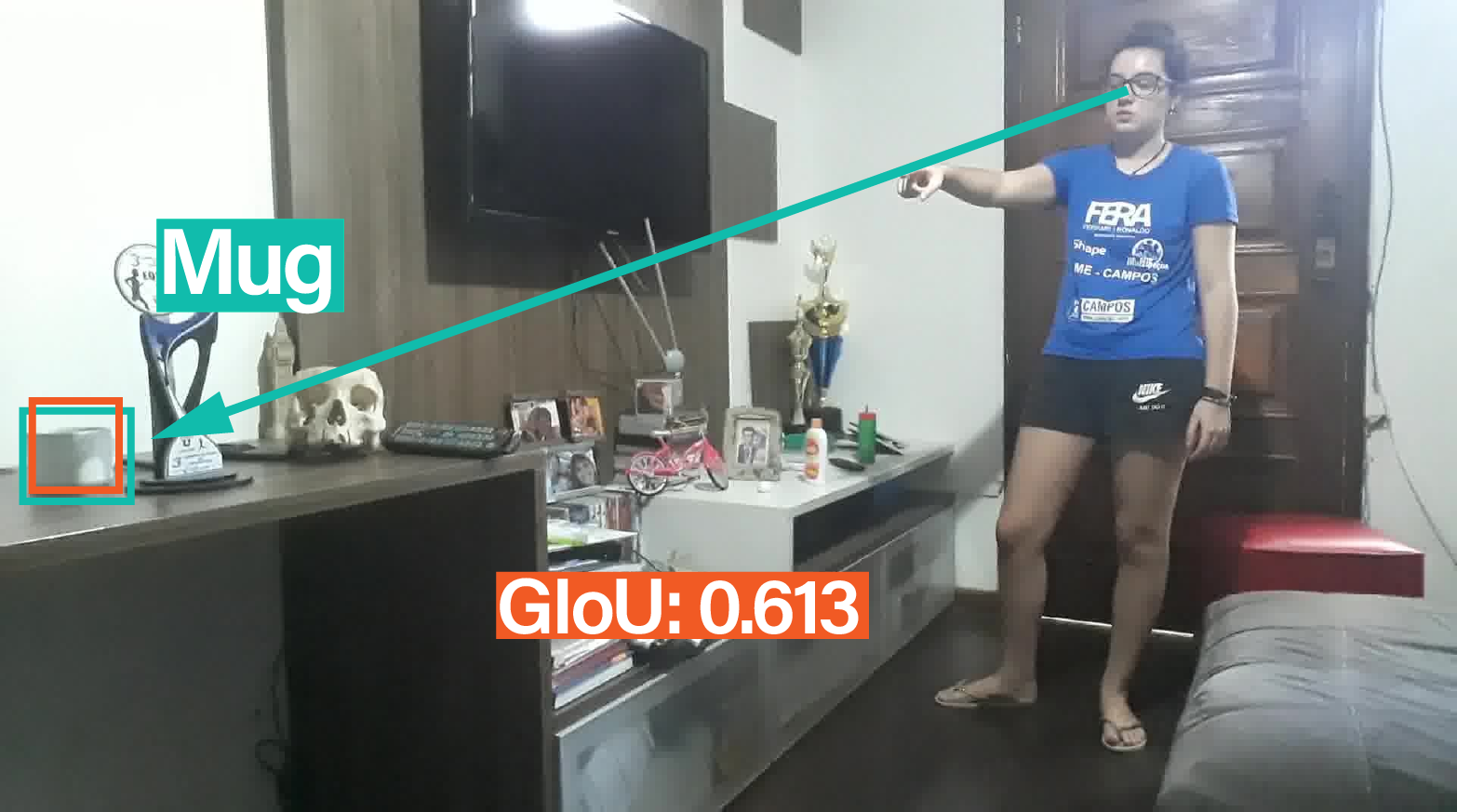}%
        \includegraphics[width=.5\linewidth]{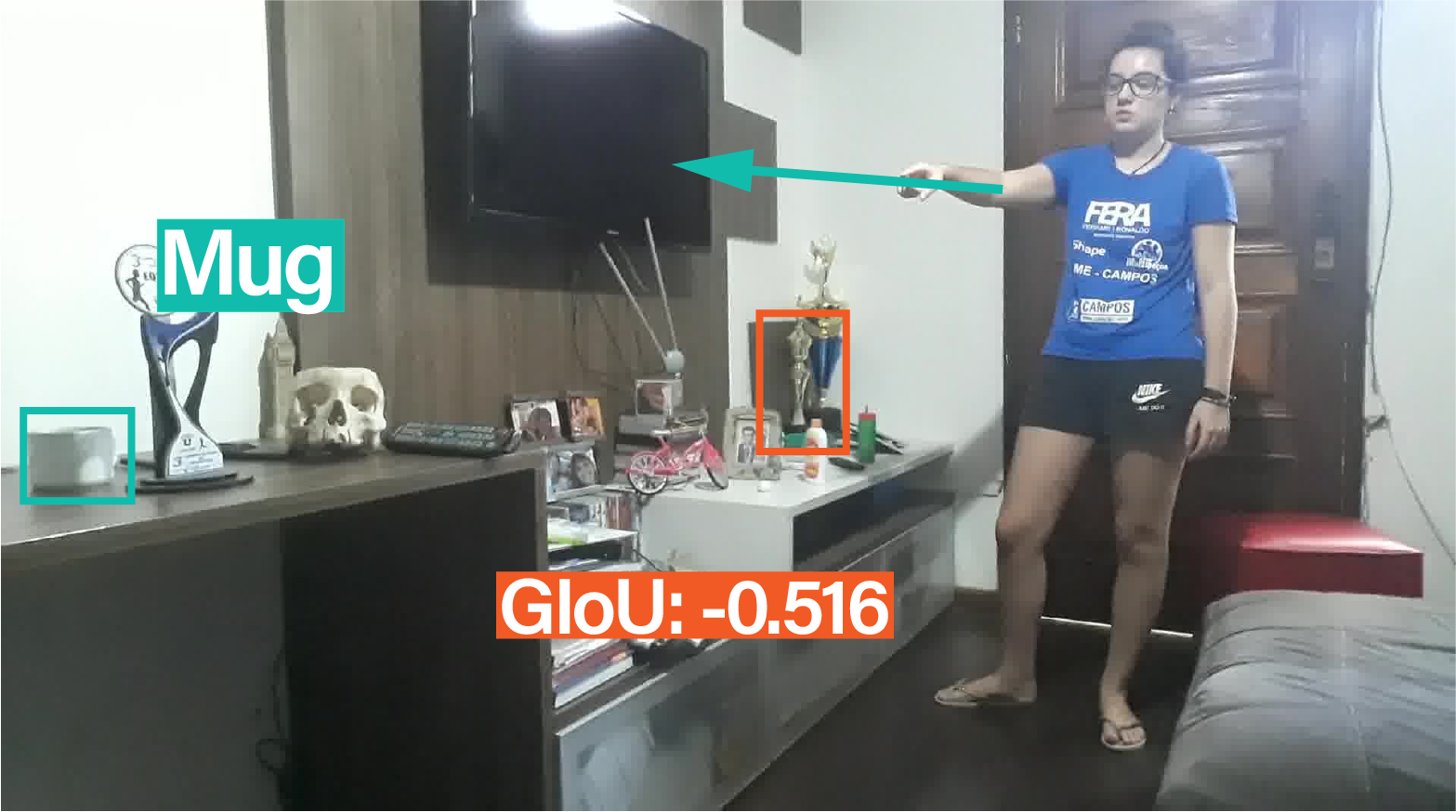}%
        \caption{}
        \label{fig:improve:j}
    \end{subfigure}%
    \begin{subfigure}[b]{0.333\linewidth}
        \centering
        \includegraphics[width=.5\linewidth]{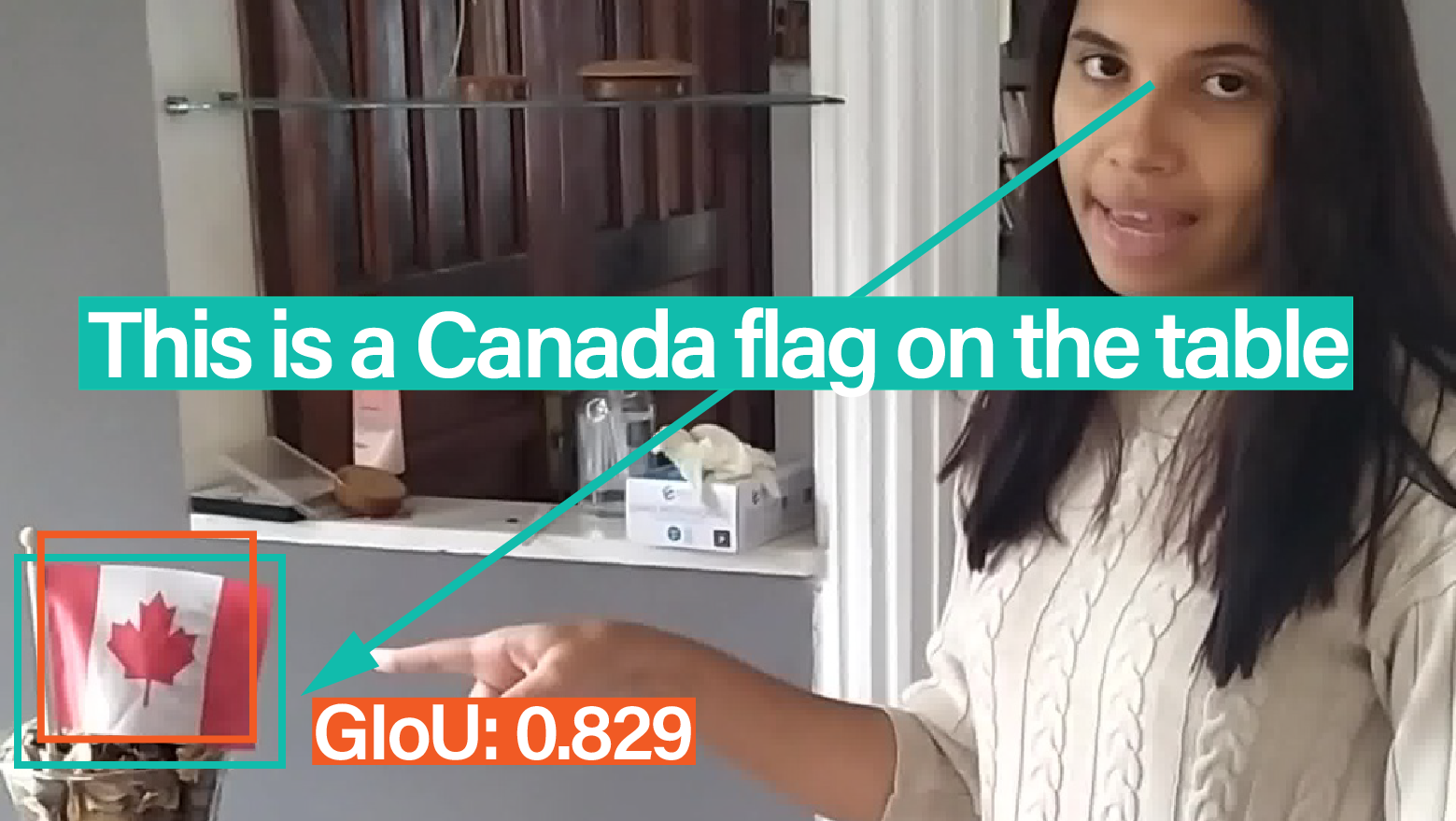}%
        \includegraphics[width=.5\linewidth]{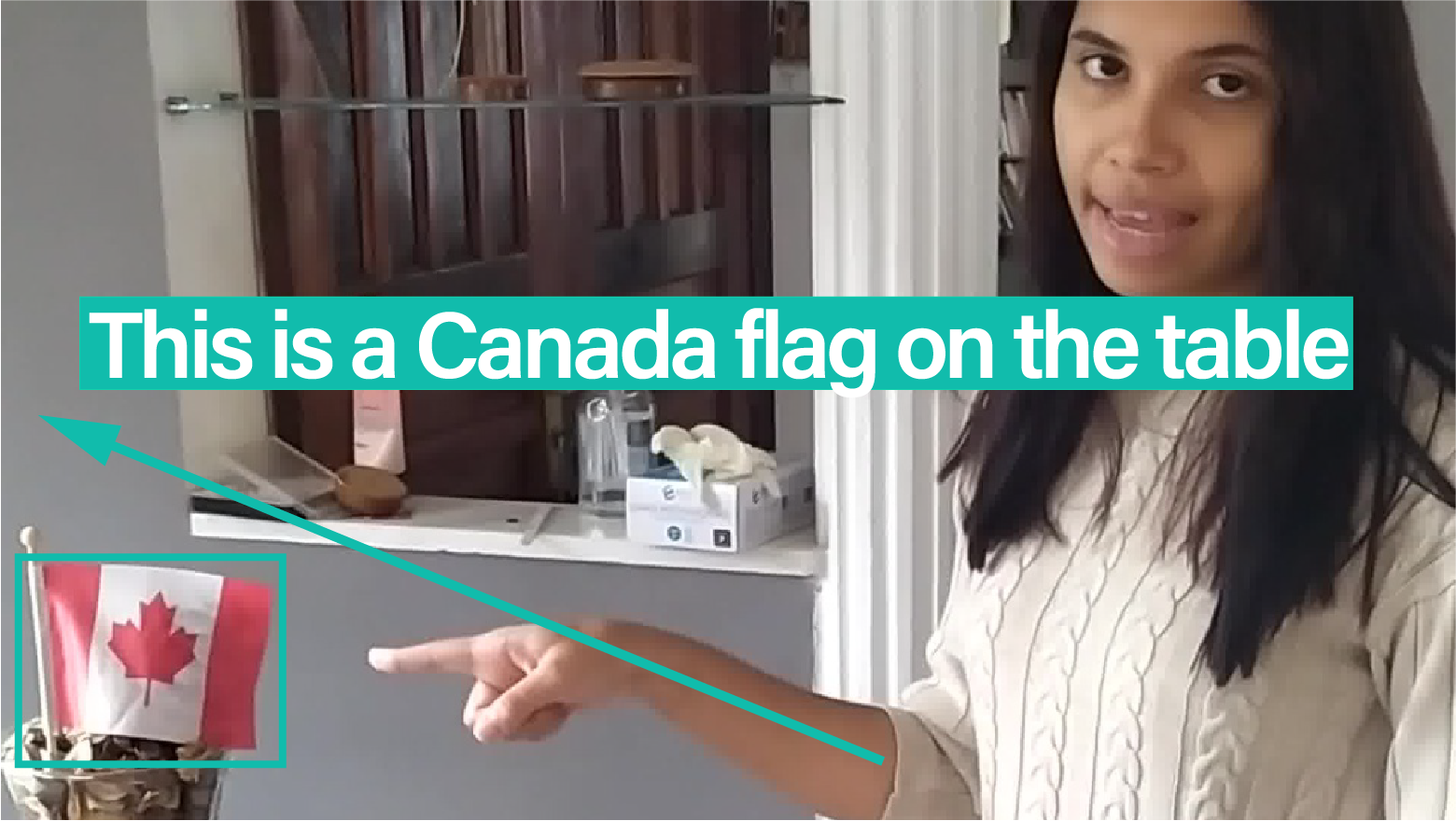}%
        \caption{}
        \label{fig:improve:k}
    \end{subfigure}%
    \begin{subfigure}[b]{0.333\linewidth}
        \centering
        \includegraphics[width=.5\linewidth]{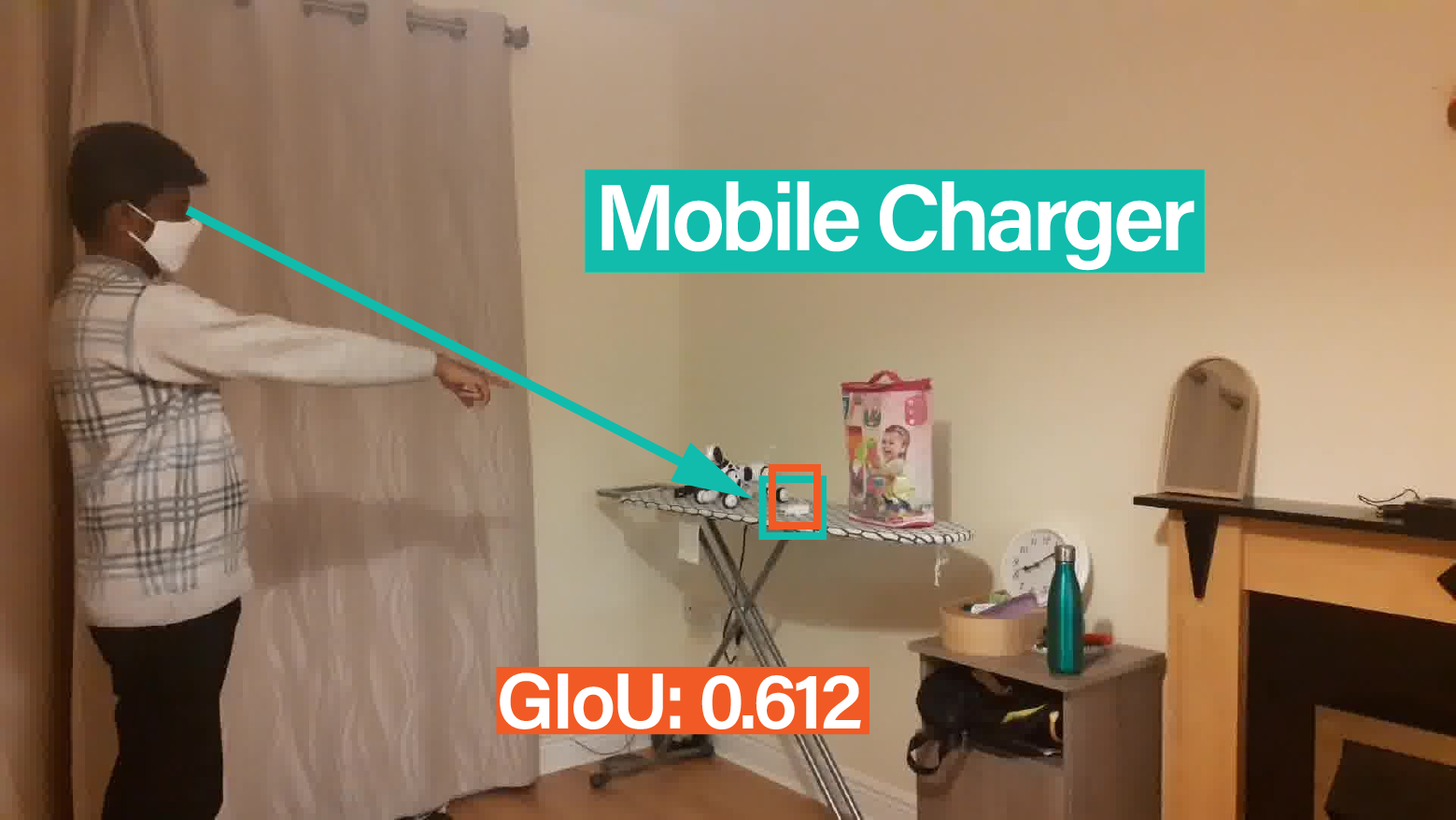}%
        \includegraphics[width=.5\linewidth]{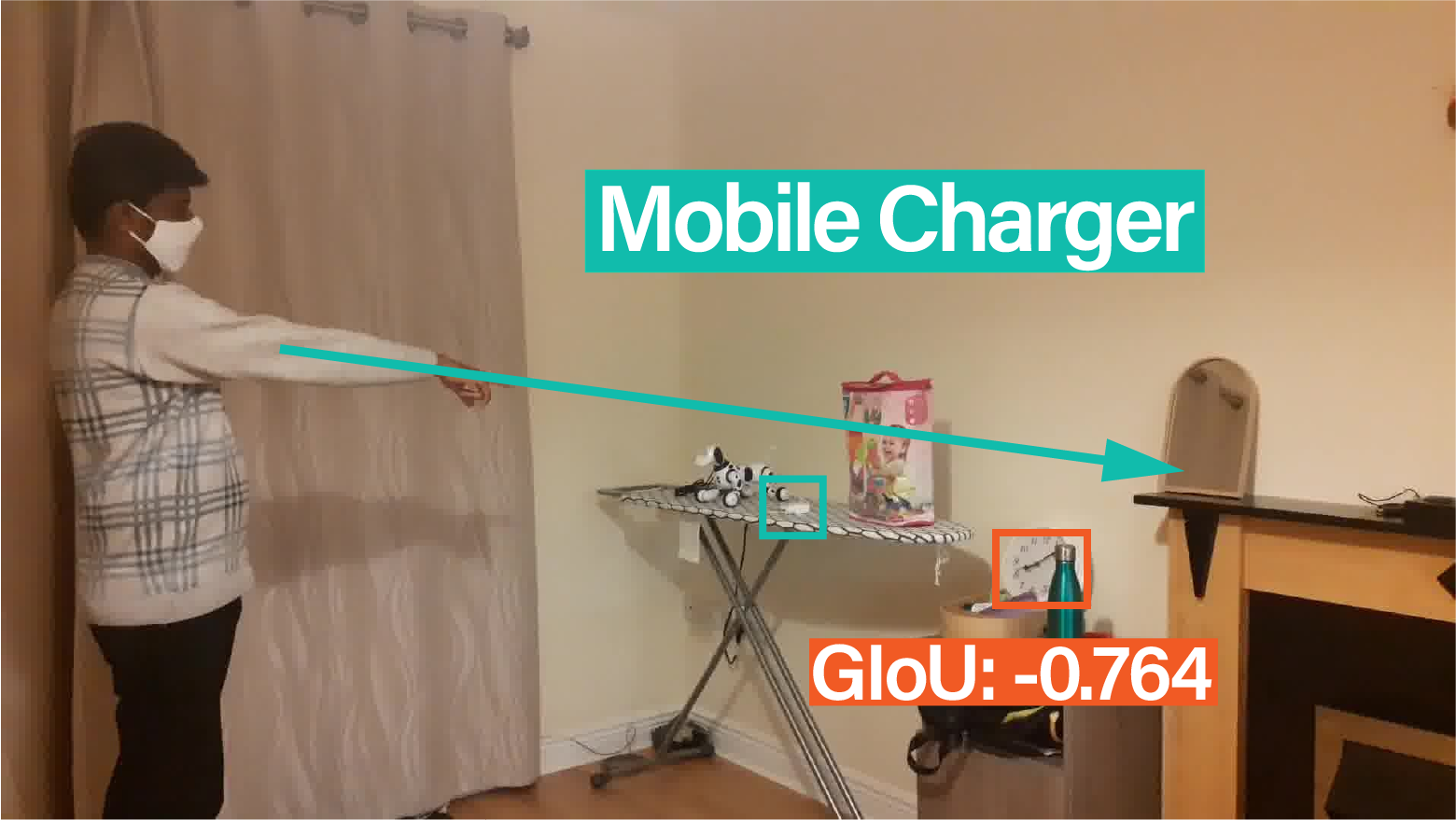}%
        \caption{}
        \label{fig:improve:l}
    \end{subfigure}%
    \caption{\textbf{The model explicitly trained to predict \acp{vtl} more accurately locates referents than the one trained to predict \acp{ewl}.} We draw green arrows (different from the predictions) to illustrate eyes and fingertips more accurately indicate object locations. (Green sentence: natural language inputs; Green box: ground-truth referent location; Red box: predicted referent location; Red numbers near predicted box: GIoU between predicted box and ground truth box.)}
    \label{fig:improve} 
    \vspace{+6pt}
\end{figure}

\subsection{Explicitly Learned Nonverbal Signals}

We report the model performance that explicitly predicts either the \acp{vtl} or the \acp{ewl}.

\paragraph{Results}

Overall, the model trained to predict the \acp{vtl} performs better than the one trained to predict the \acp{ewl} under all three IoU thresholds; see \cref{tab:diff_poses}. The model trained to explicitly predict the \acp{ewl} performs even worse than the one that was not trained to explicitly predict any gestural signals under the IoU threshold of 0.75.

\paragraph{Analysis}

Under the IoU threshold of 0.75, the worse performance of the model with explicit \ac{ewl} prediction can be partly attributed to the unreliability of the arm's orientation and the stringent precision requirement of the 0.75 IoU threshold. As we observed in \cref{fig:improve}, the \acp{ewl} are not reliable for predicting objects' locations, simply because they oftentimes do not pass through the referents. This mismatch prevents the model from correctly determining the referents' locations using the \acp{ewl}' orientations. For instance, in \cref{fig:improve:a} (right), the \ac{ewl} fails to pass the glass cup in the yellow box; it passes a ceramic cup instead. With contradictory nonverbal (the ceramic cup) and verbal (the glass cup) signals, the model struggles to identify the referent correctly.

\begin{wraptable}{r}{0.45\linewidth}
    \vspace{+6pt}
    \caption{\textbf{Effects of learning two different types of postural key points.}}
    \label{tab:diff_poses}
    \centering
    \small
    \begin{tabular}{cccc}
        \toprule
        IoU & None & EWL & VTL \\
        \midrule
        0.25 & 64.9 & 69.5 \textcolor[RGB]{200,0,0}{(+4.6)} & 71.1 \textcolor[RGB]{200,0,0}{(+6.2)} \\
        0.50 & 57.4 & 60.7 \textcolor[RGB]{200,0,0}{(+3.3)} & 63.5 \textcolor[RGB]{200,0,0}{(+6.1)} \\
        0.75 & 37.2 & 35.5 \textcolor[RGB]{0,200,0}{(-1.7)} & 39.0 \textcolor[RGB]{200,0,0}{(+1.8)} \\
        \bottomrule
    \end{tabular}
\end{wraptable}

In contrast, \acp{vtl} partly explain the model's improved performance. For example in \cref{fig:improve:b}, the \ac{vtl} passes the chair in the yellow box while the \ac{ewl} fails. Similarly, in \cref{fig:improve:c,fig:improve:e,fig:improve:f,fig:improve:h,fig:improve:l}, \acp{vtl} passes the referent while \acp{ewl} do not. Additionally in \cref{fig:improve:a,fig:improve:d,fig:improve:g,fig:improve:i,fig:improve:j,fig:improve:k}, \acp{vtl} are closer to the referents' box centers than \ac{ewl} are. The higher consistency between verbal and nonverbal signals increases the chance of successfully identifying the referents.

Under the IoU thresholds of 0.25 and 0.50, the improved model performance can be partly attributed to the rough orientations provided by the \acp{ewl} and the more lenient precision requirements of the lower thresholds. Specifically, the rough orientations provided by the \acp{ewl} might help the model eliminate objects that significantly deviate from this direction. Hence, the model can choose from a smaller number of objects, leading to a higher probability of correctly locating the referents.

\begin{figure}[h!]
    \centering
    \vspace{+18pt}
    \begin{minipage}[b]{0.729\linewidth}
        \begin{subfigure}[b]{.5\linewidth}
            \centering
            \begin{overpic}
                [width=.5\linewidth]{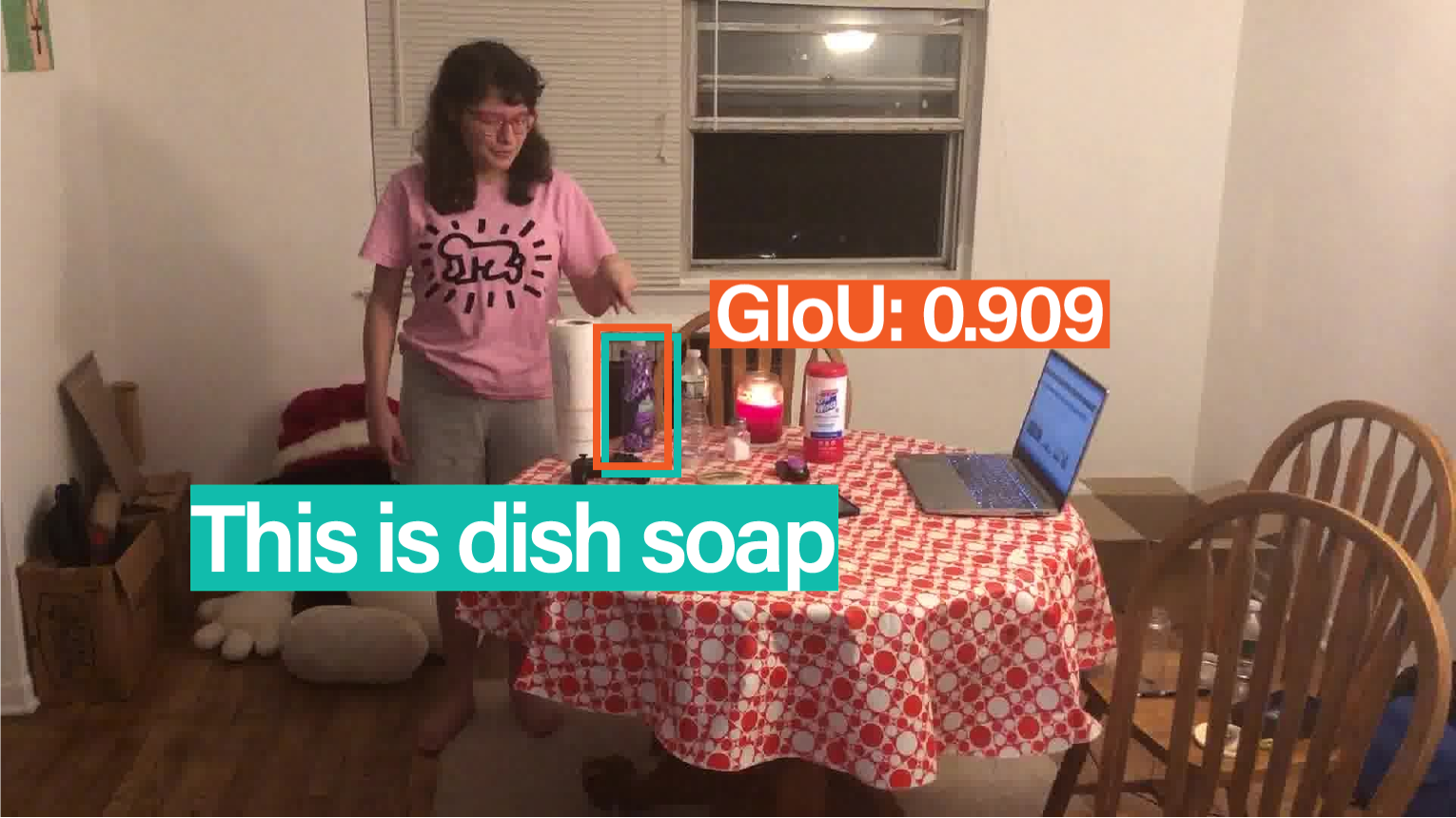}
                \put(30,60){\color{black}original}
            \end{overpic}%
            \begin{overpic}
                [width=.5\linewidth]{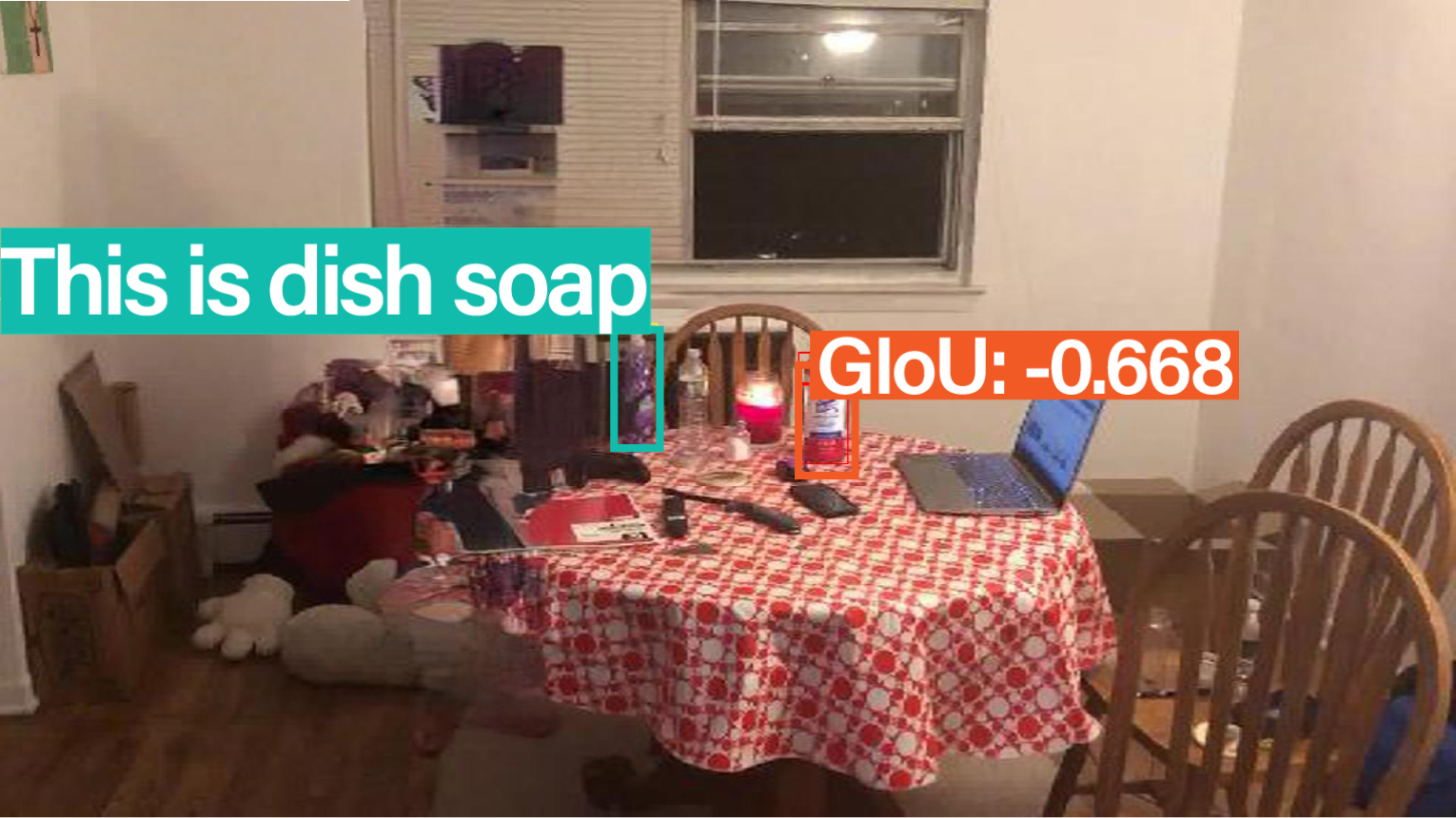}
                \put(25,60){\color{black}inpainted}
            \end{overpic}%
            \caption{}
        \end{subfigure}%
        \begin{subfigure}[b]{.5\linewidth}
            \centering
            \begin{overpic}
                [width=.5\linewidth]{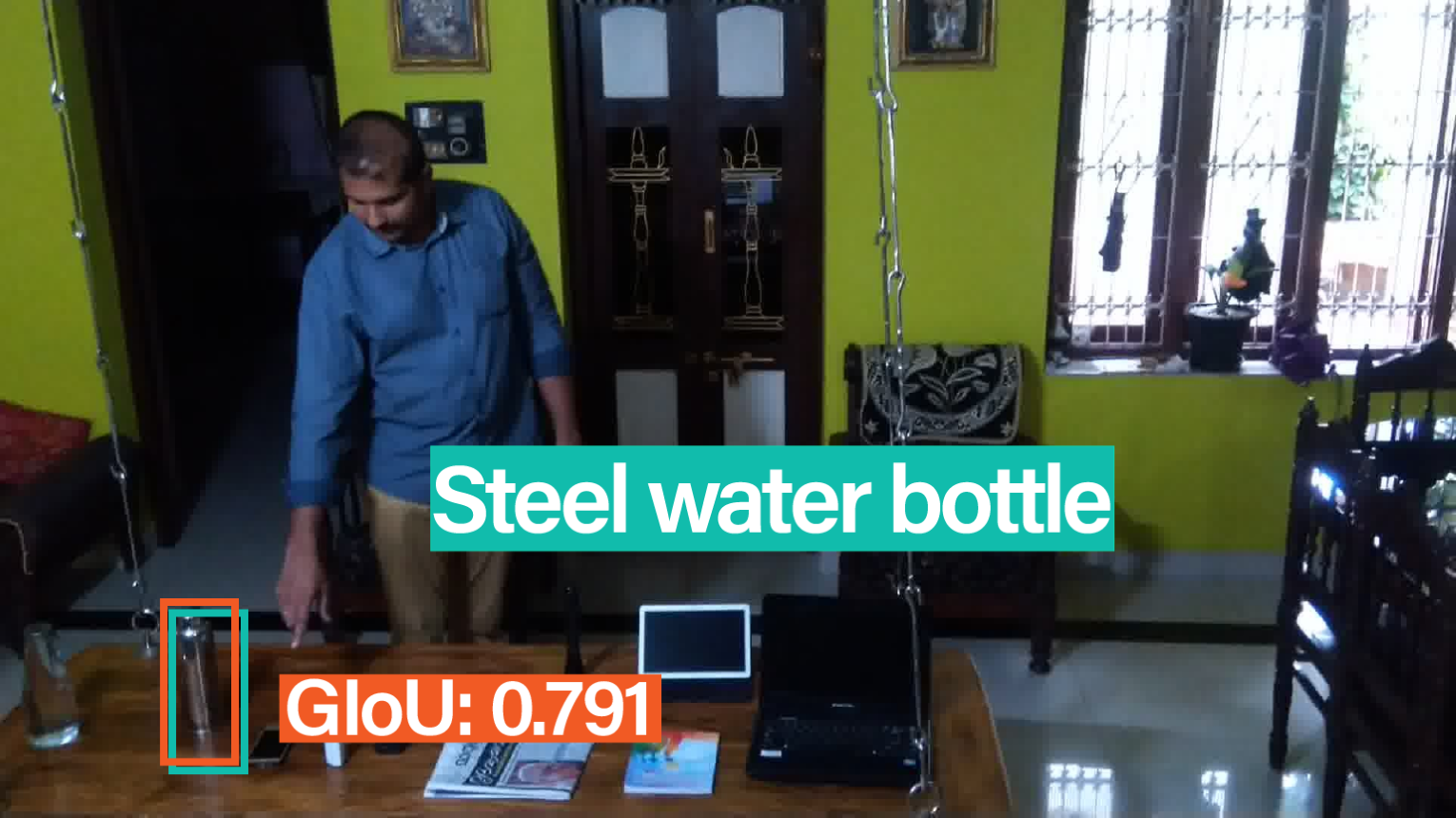}
                \put(30,60){\color{black}original}
            \end{overpic}%
            \begin{overpic}
                [width=.5\linewidth]{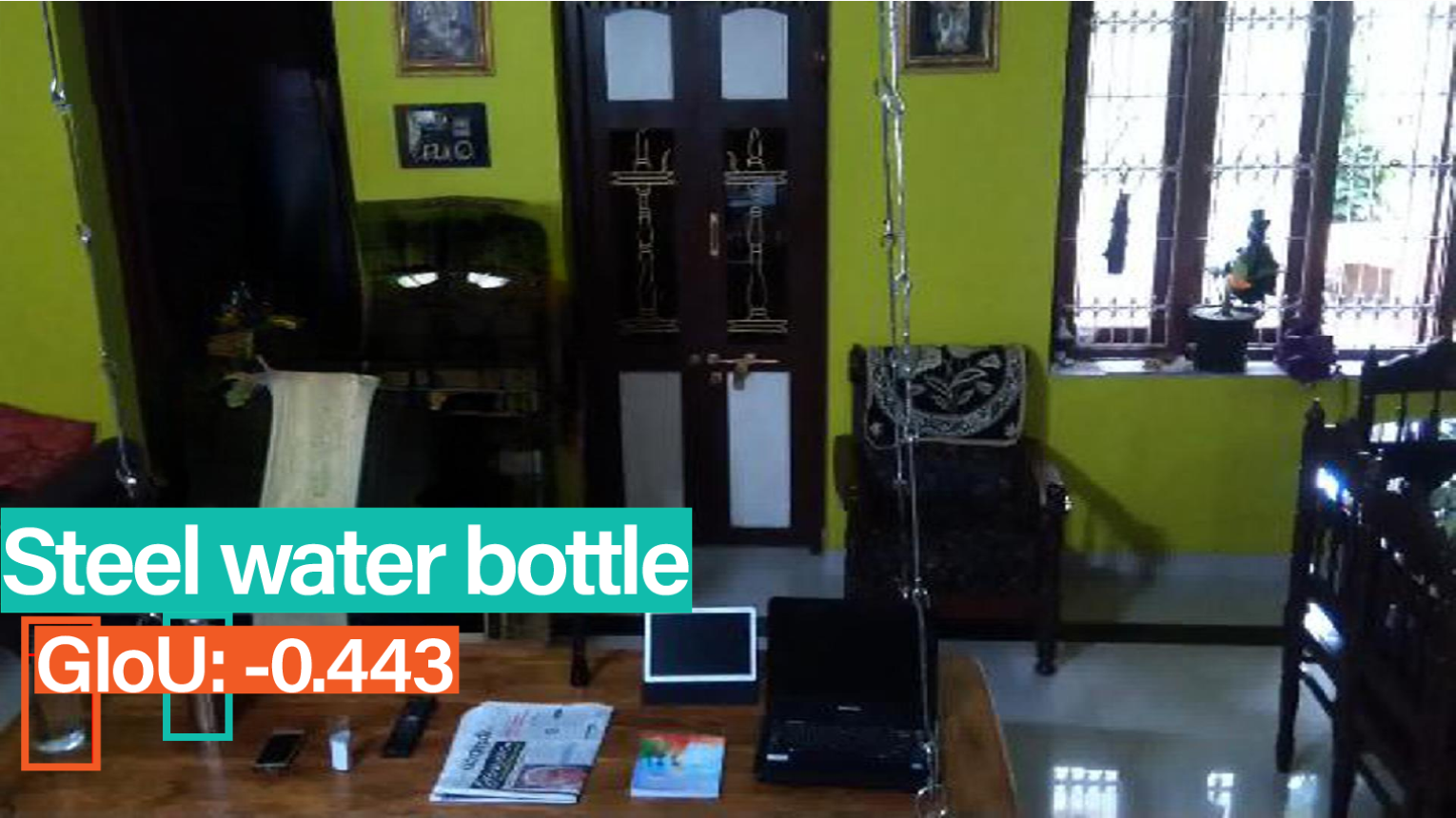}
                \put(25,60){\color{black}inpainted}
            \end{overpic}%
            \caption{}
        \end{subfigure}%
        \\%
        \begin{subfigure}[b]{.5\linewidth}
            \centering
            \includegraphics[width=.5\linewidth]{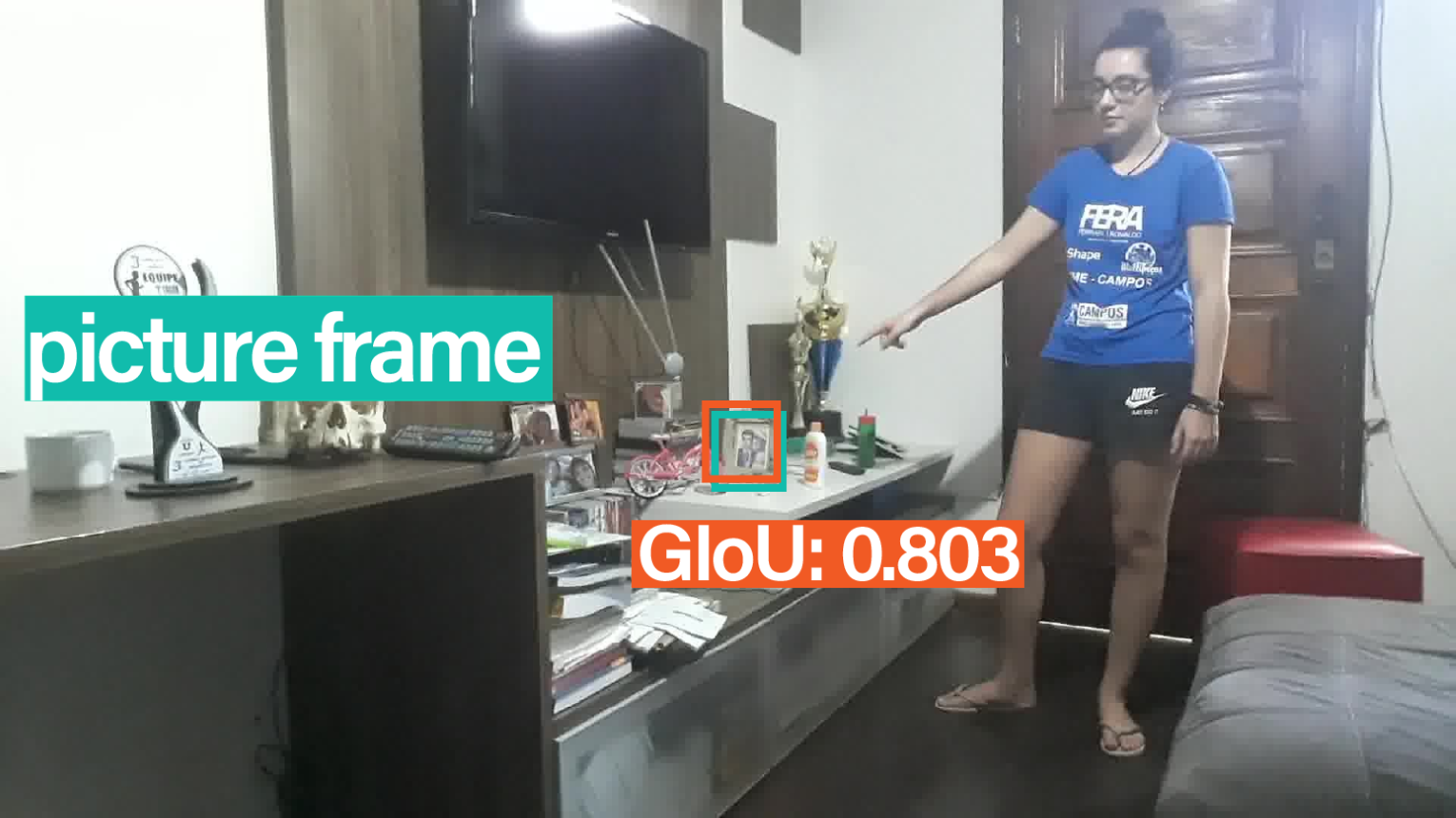}%
            \includegraphics[width=.5\linewidth]{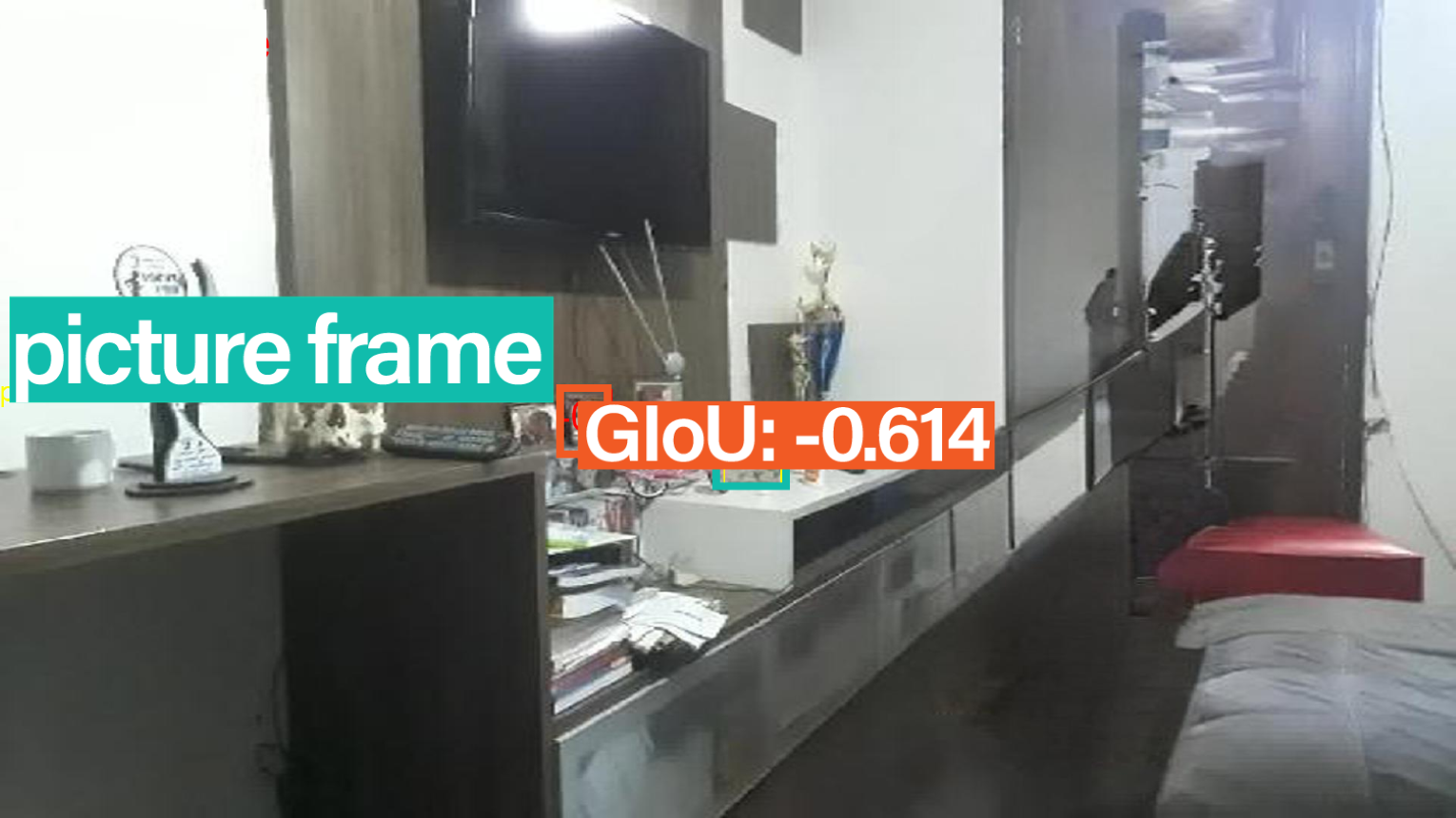}%
            \caption{}
        \end{subfigure}%
        \begin{subfigure}[b]{.5\linewidth}
            \centering
            \includegraphics[width=.5\linewidth]{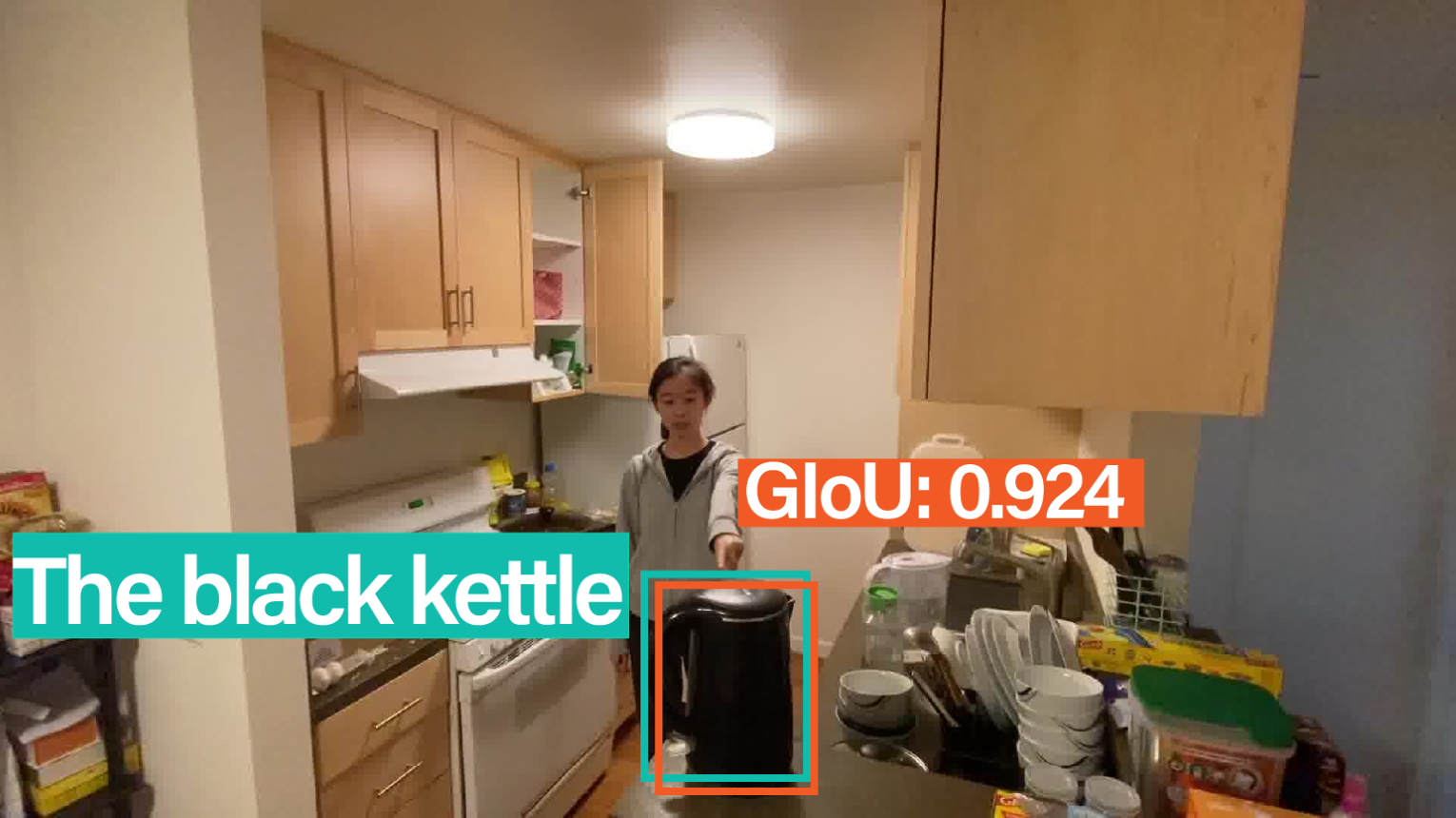}%
            \includegraphics[width=.5\linewidth]{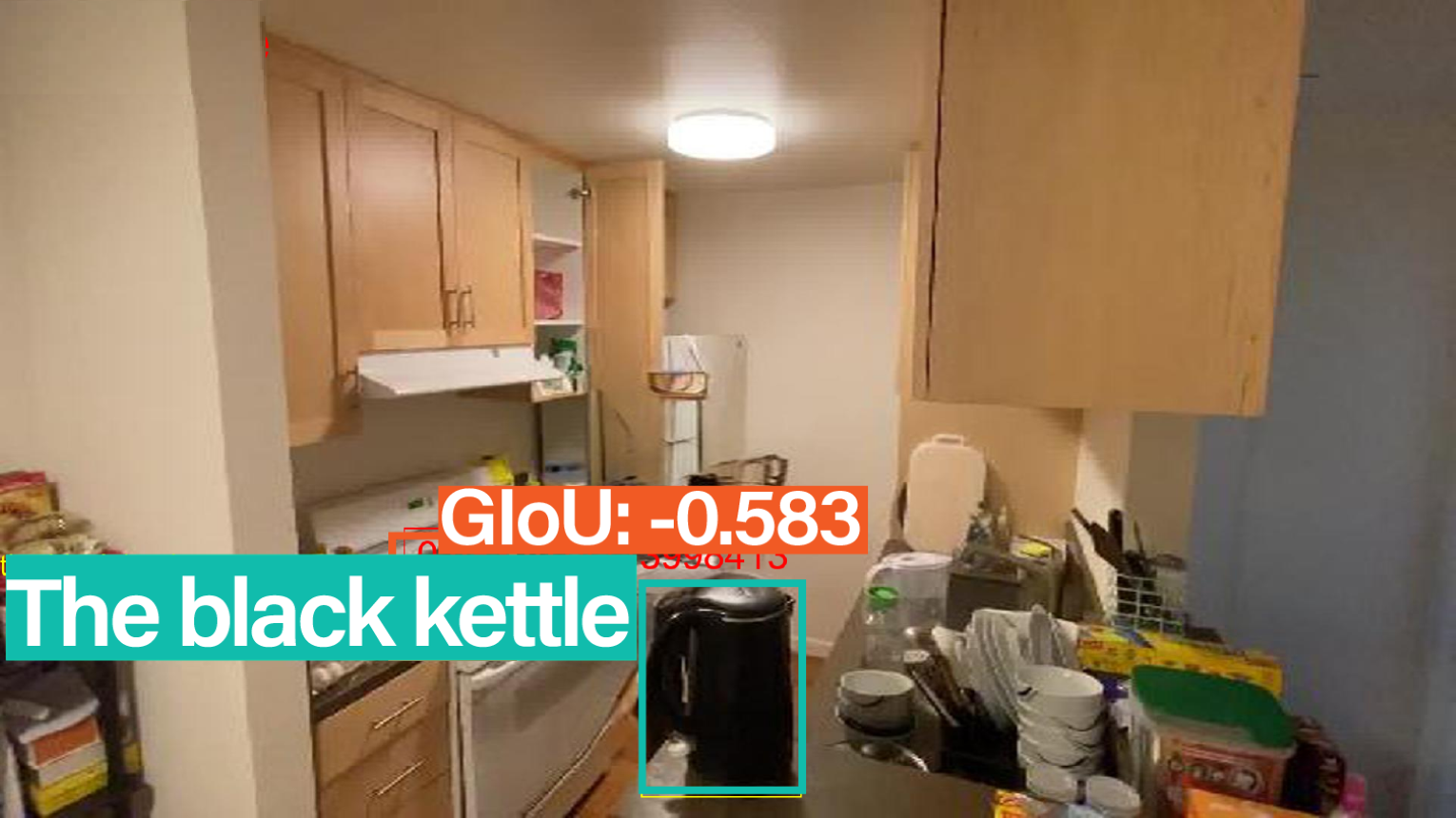}%
            \caption{}
        \end{subfigure}%
        \\%
        \begin{subfigure}[b]{.5\linewidth}
            \centering
            \includegraphics[width=.5\linewidth]{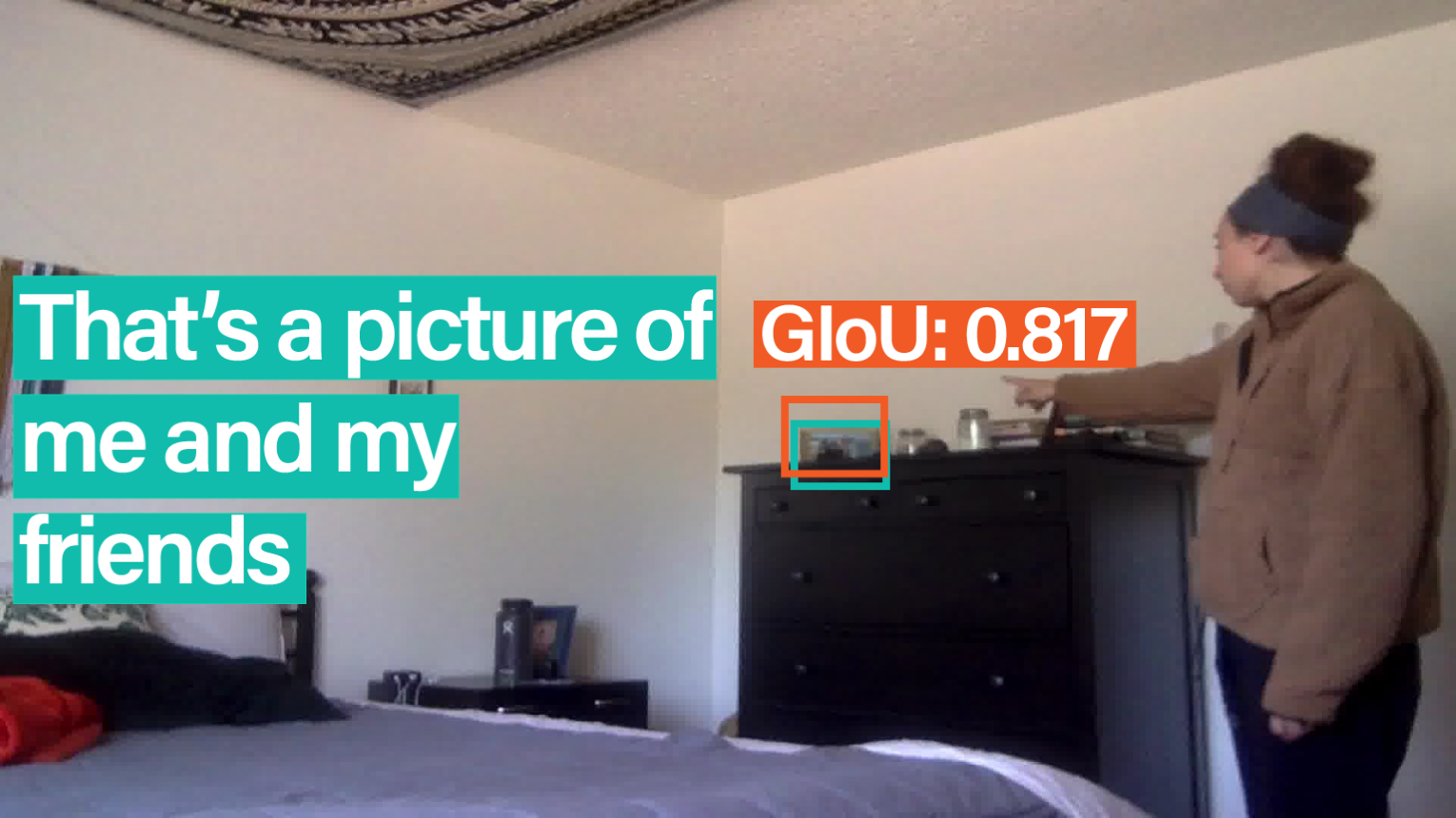}%
            \includegraphics[width=.5\linewidth]{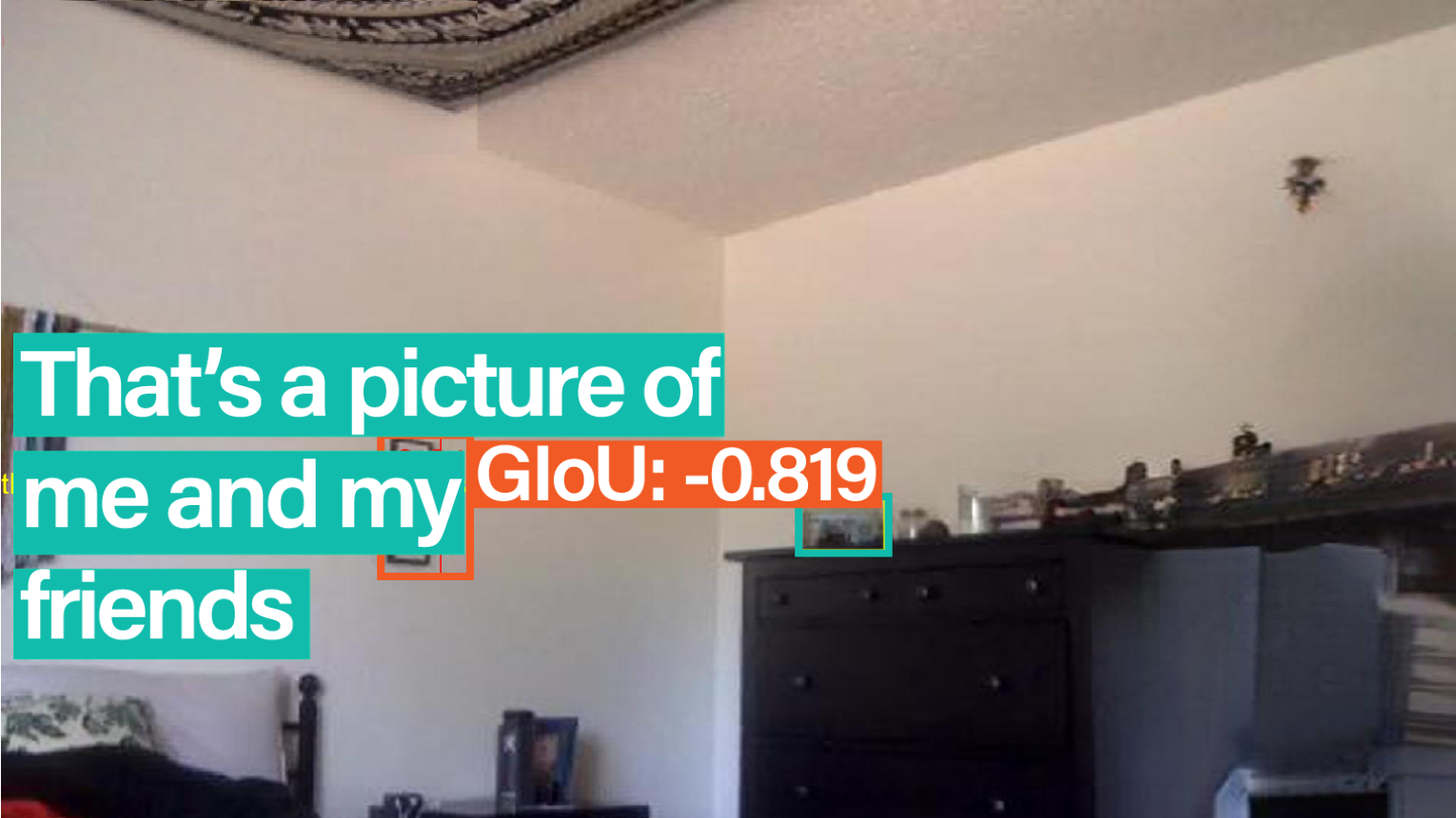}%
            \caption{}
            \label{fig:perf_drop:d}
        \end{subfigure}%
        \begin{subfigure}[b]{.5\linewidth}
            \centering
            \includegraphics[width=.5\linewidth]{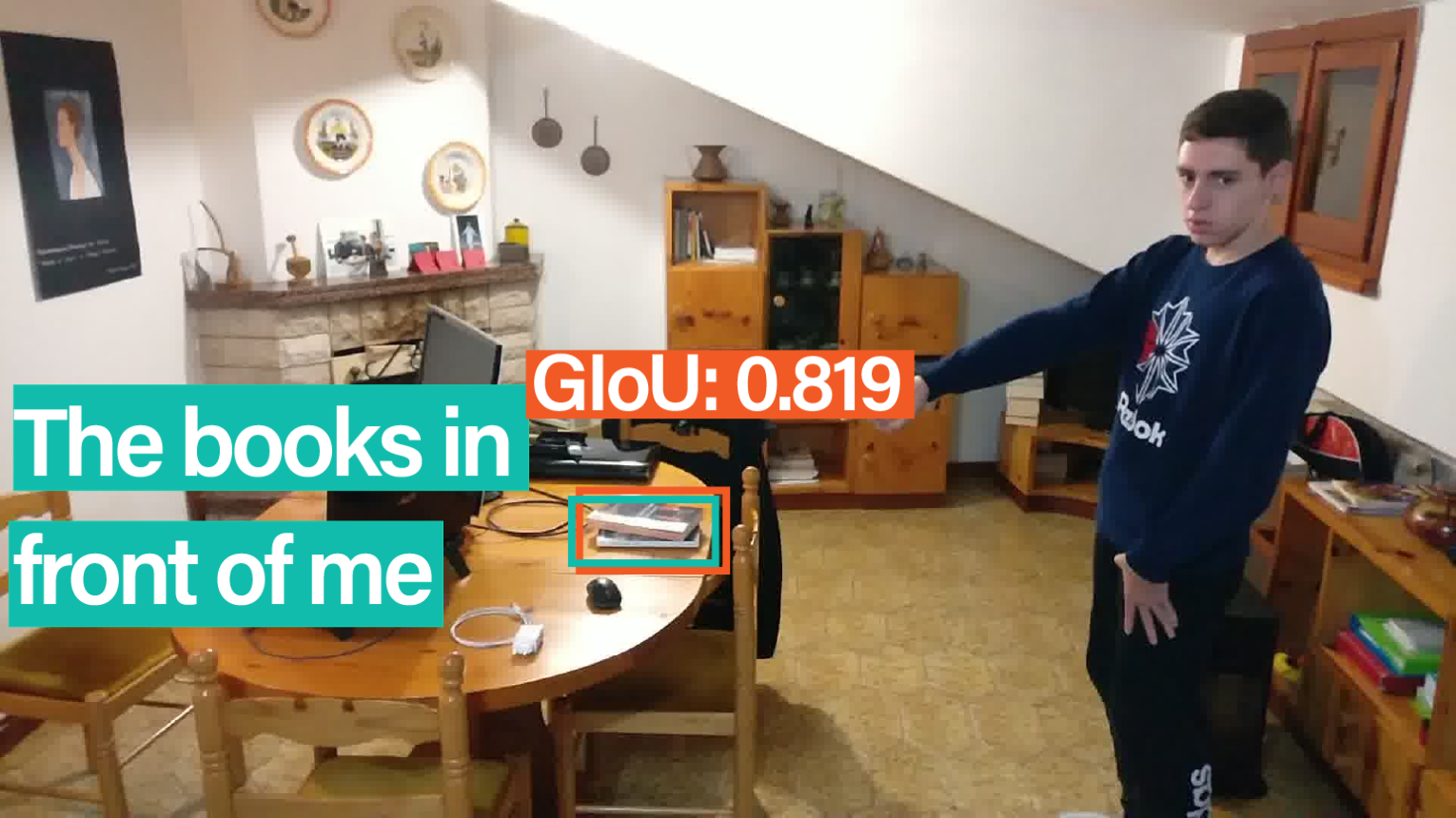}%
            \includegraphics[width=.5\linewidth]{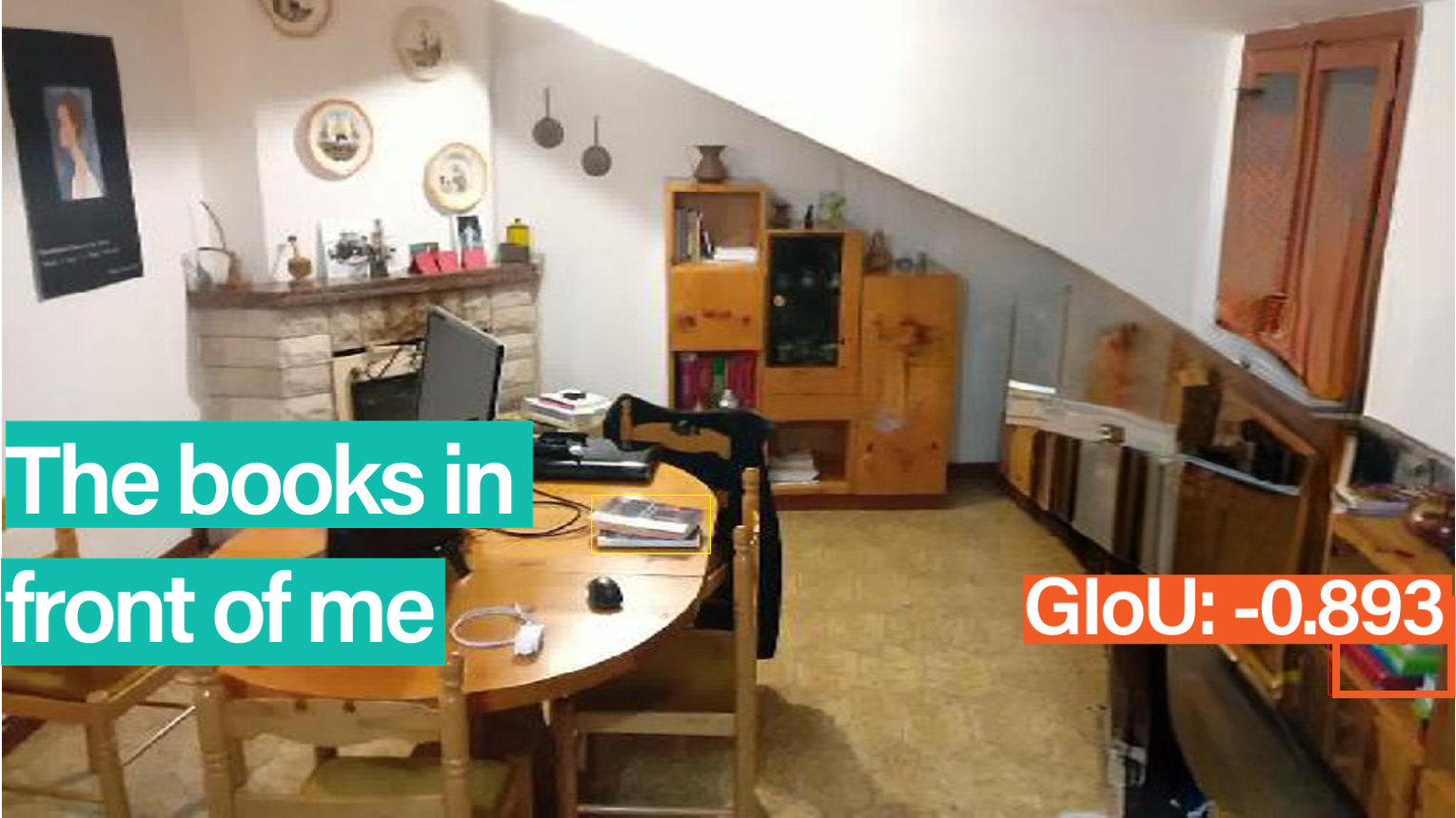}%
            \caption{}
            \label{fig:perf_drop:f}
        \end{subfigure}%
        \\%
        \begin{subfigure}[b]{.5\linewidth}
            \centering
            \includegraphics[width=.5\linewidth]{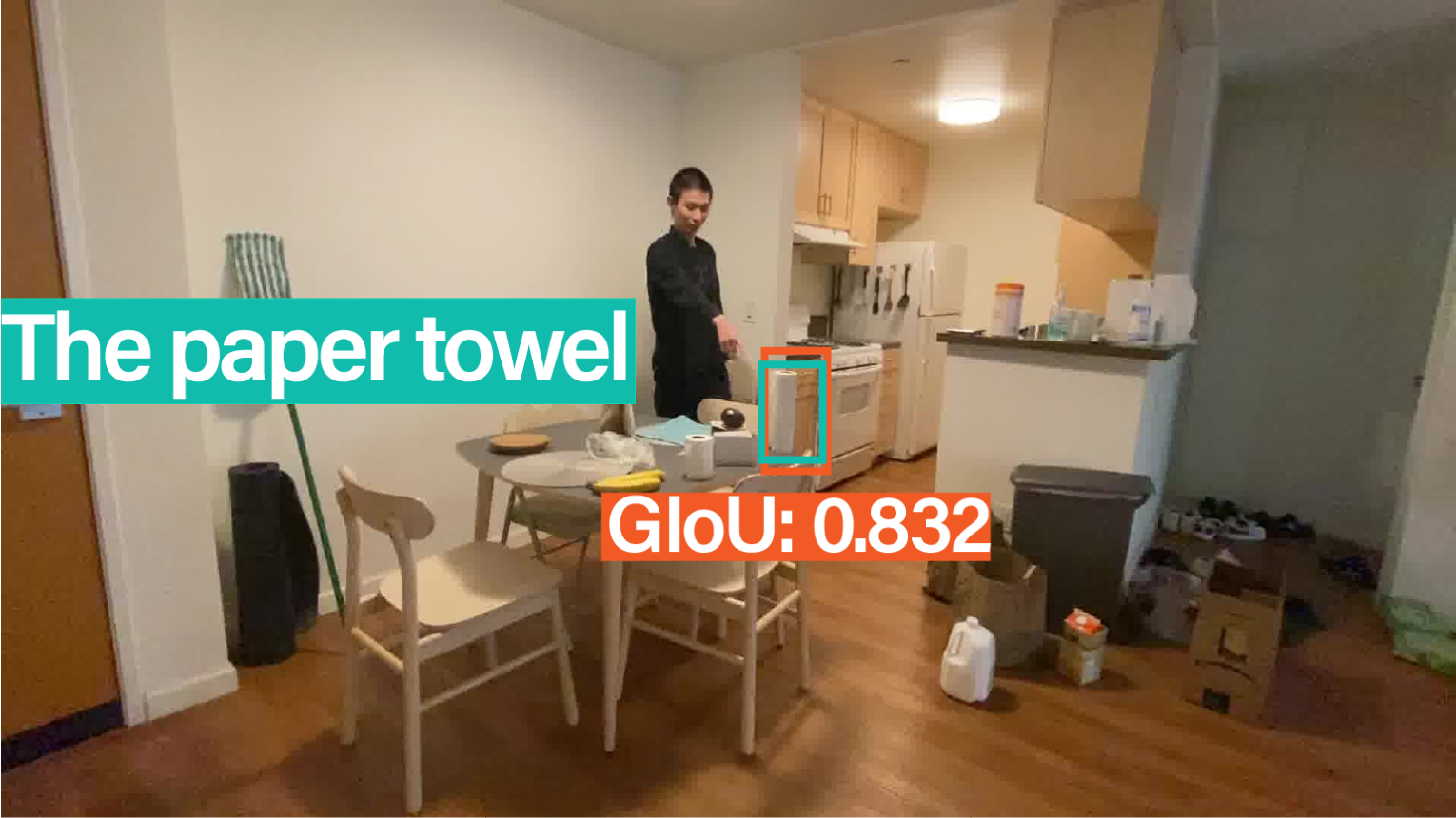}%
            \includegraphics[width=.5\linewidth]{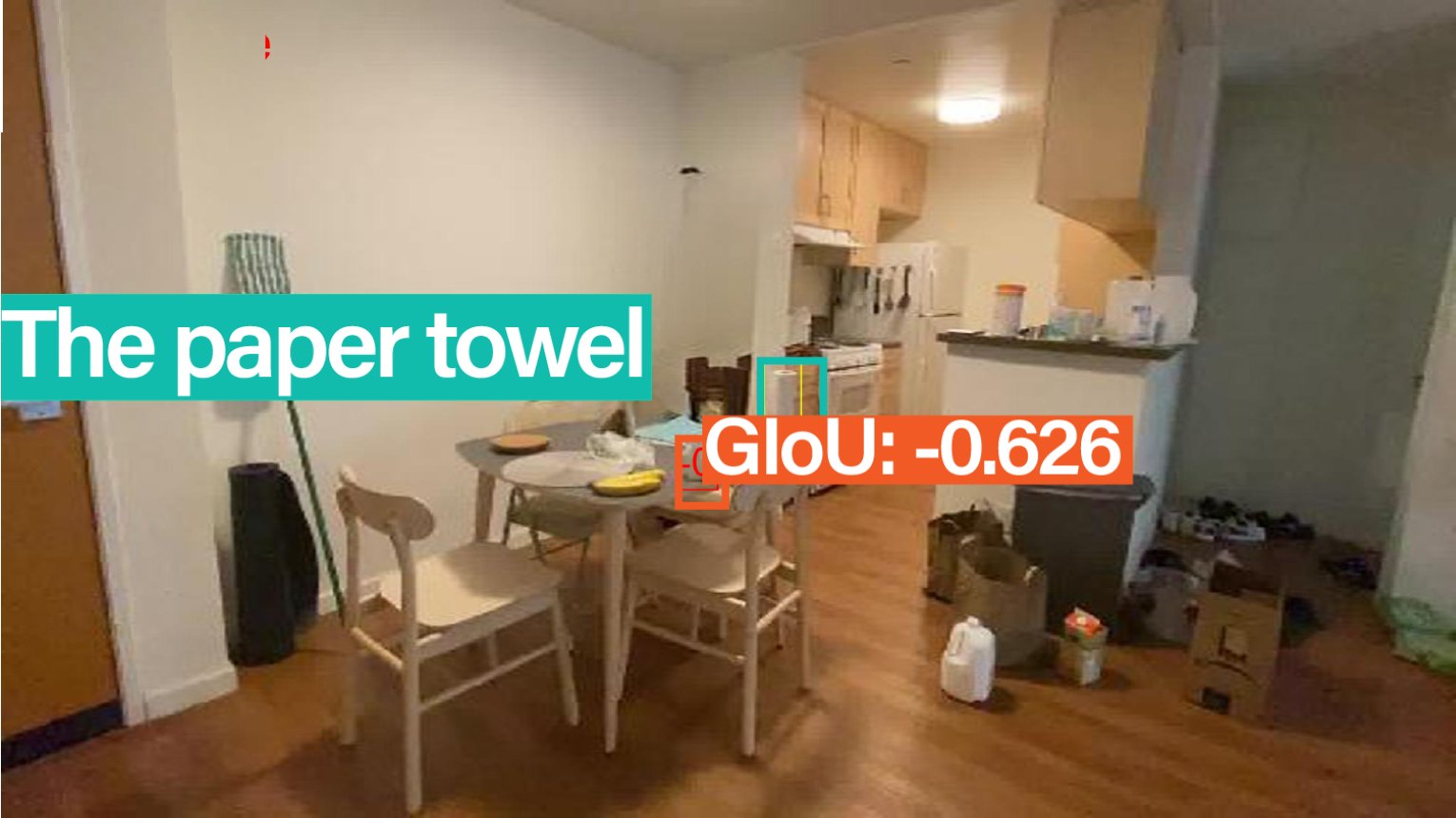}%
            \caption{}
        \end{subfigure}%
        \begin{subfigure}[b]{.5\linewidth}
            \centering
            \includegraphics[width=.5\linewidth]{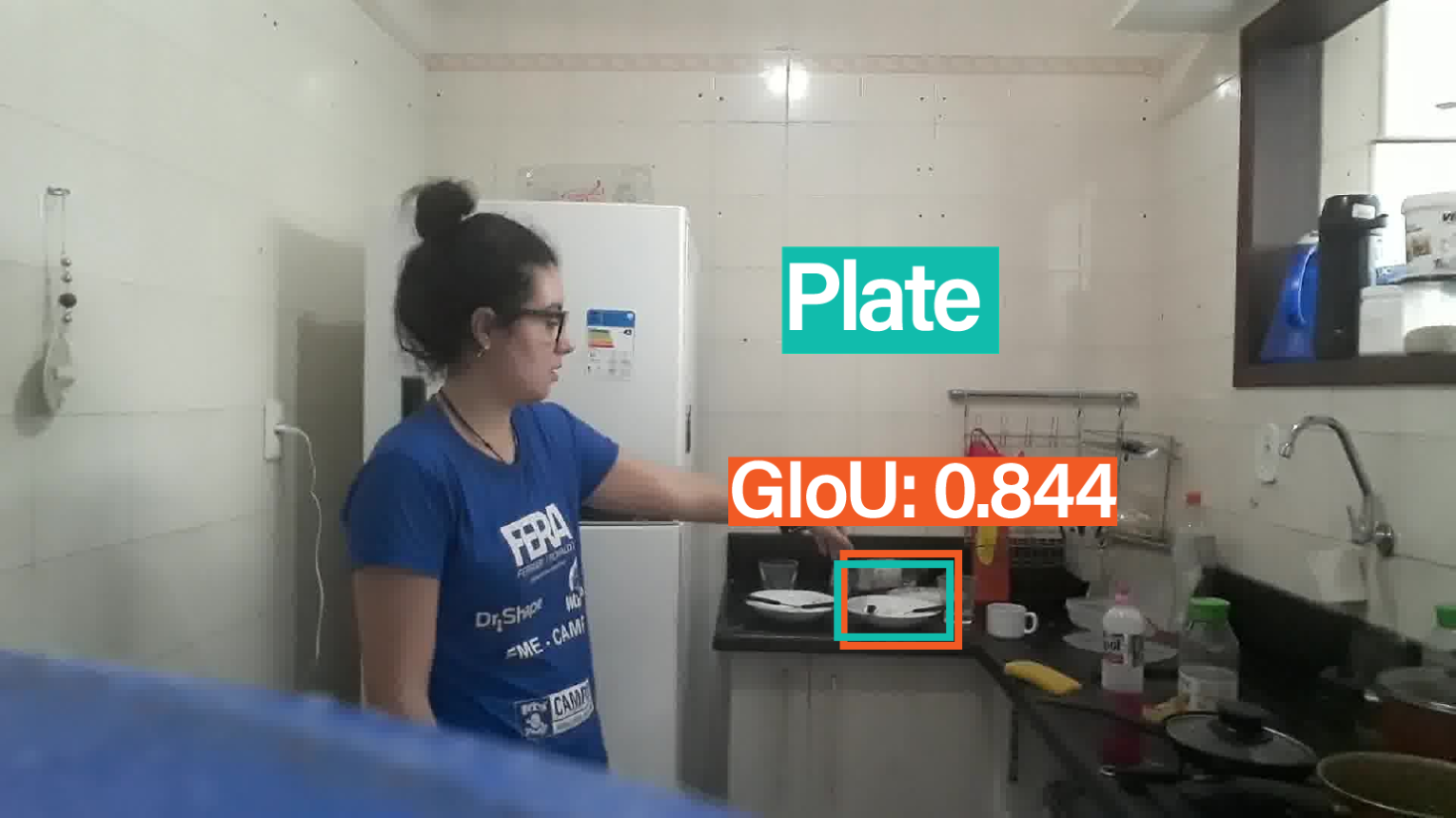}%
            \includegraphics[width=.5\linewidth]{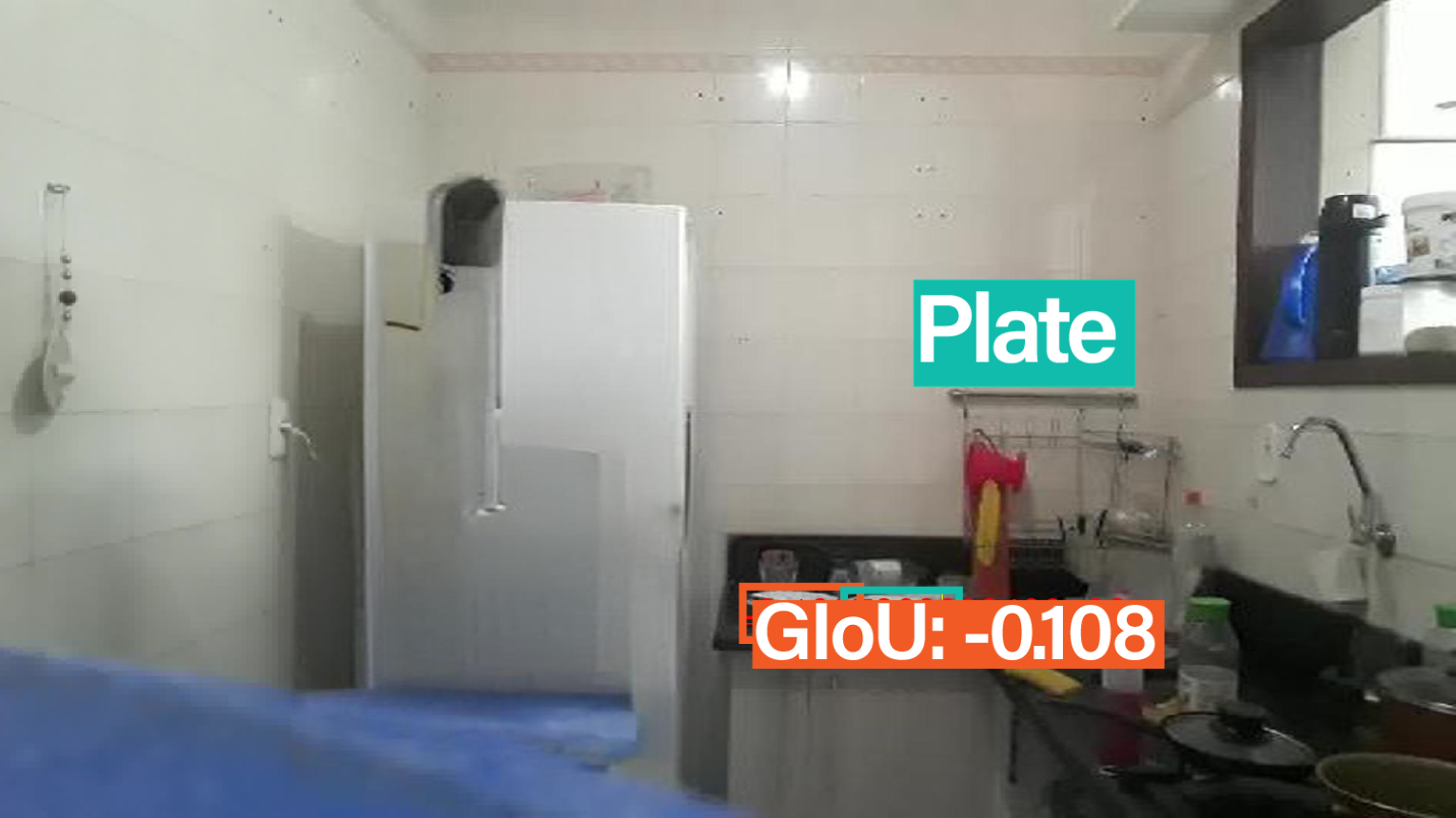}%
            \caption{}
        \end{subfigure}%
    \end{minipage}%
    \begin{minipage}[b]{0.271\linewidth}
        \begin{subfigure}[b]{\linewidth}
            \centering
            \begin{overpic}
                [width=.5\linewidth]{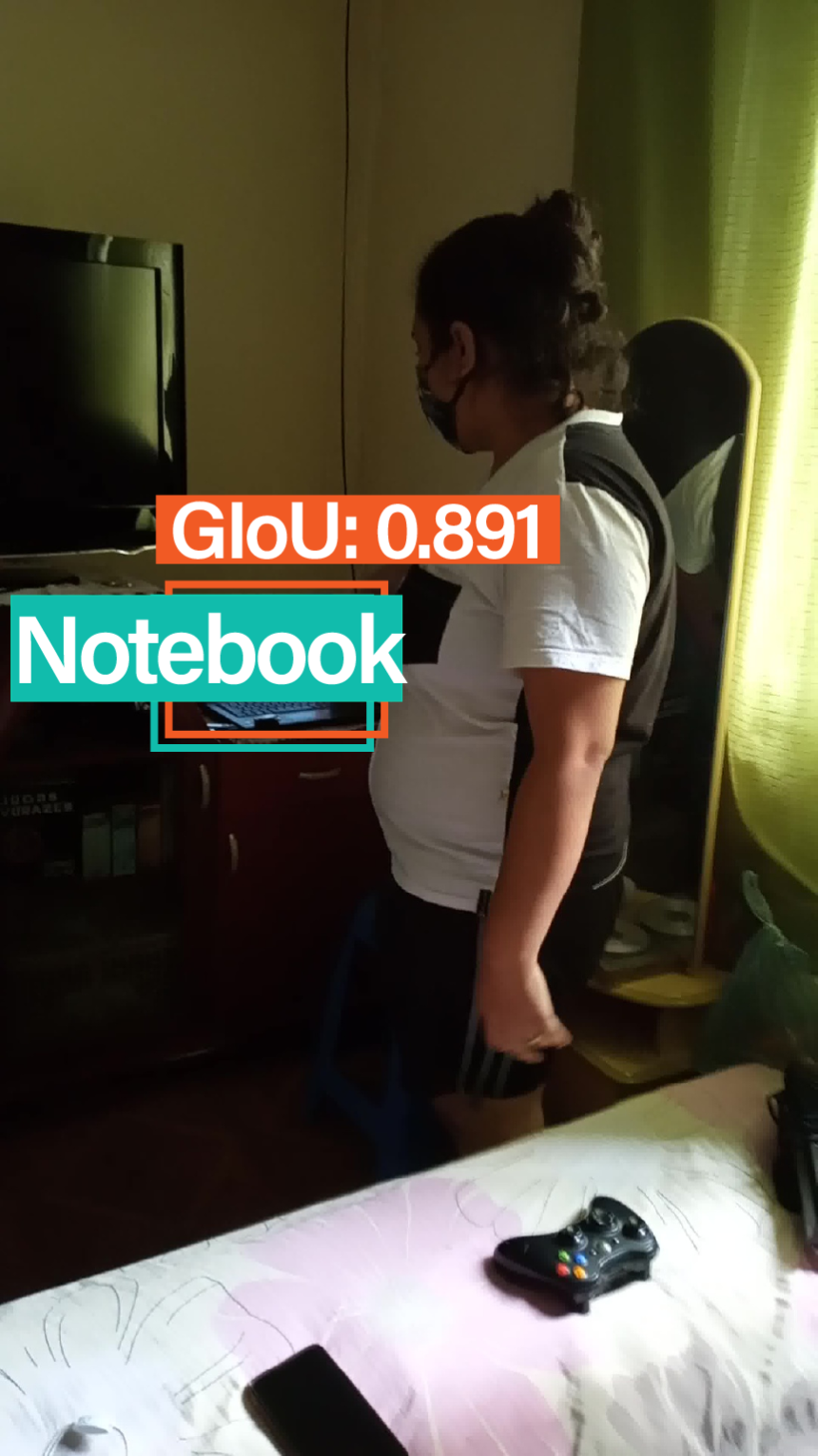}
                \put(15,102){\color{black}original}
            \end{overpic}%
            \begin{overpic}
                [width=.5\linewidth]{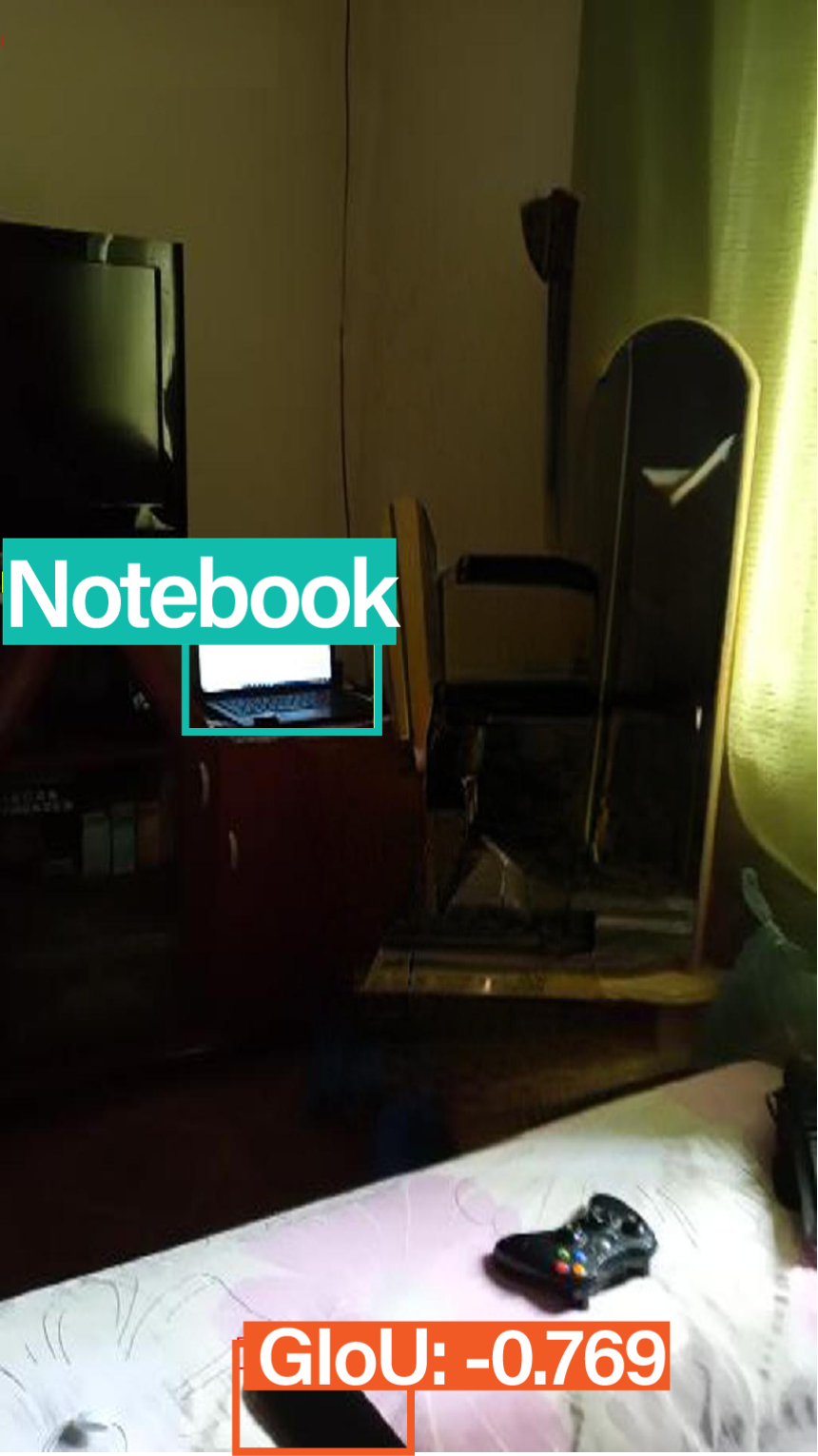}
                \put(12,102){\color{black}inpainted}
            \end{overpic}%
            \caption{}
        \end{subfigure}%
        \\%
        \begin{subfigure}[b]{\linewidth}
            \centering
            \includegraphics[width=.5\linewidth]{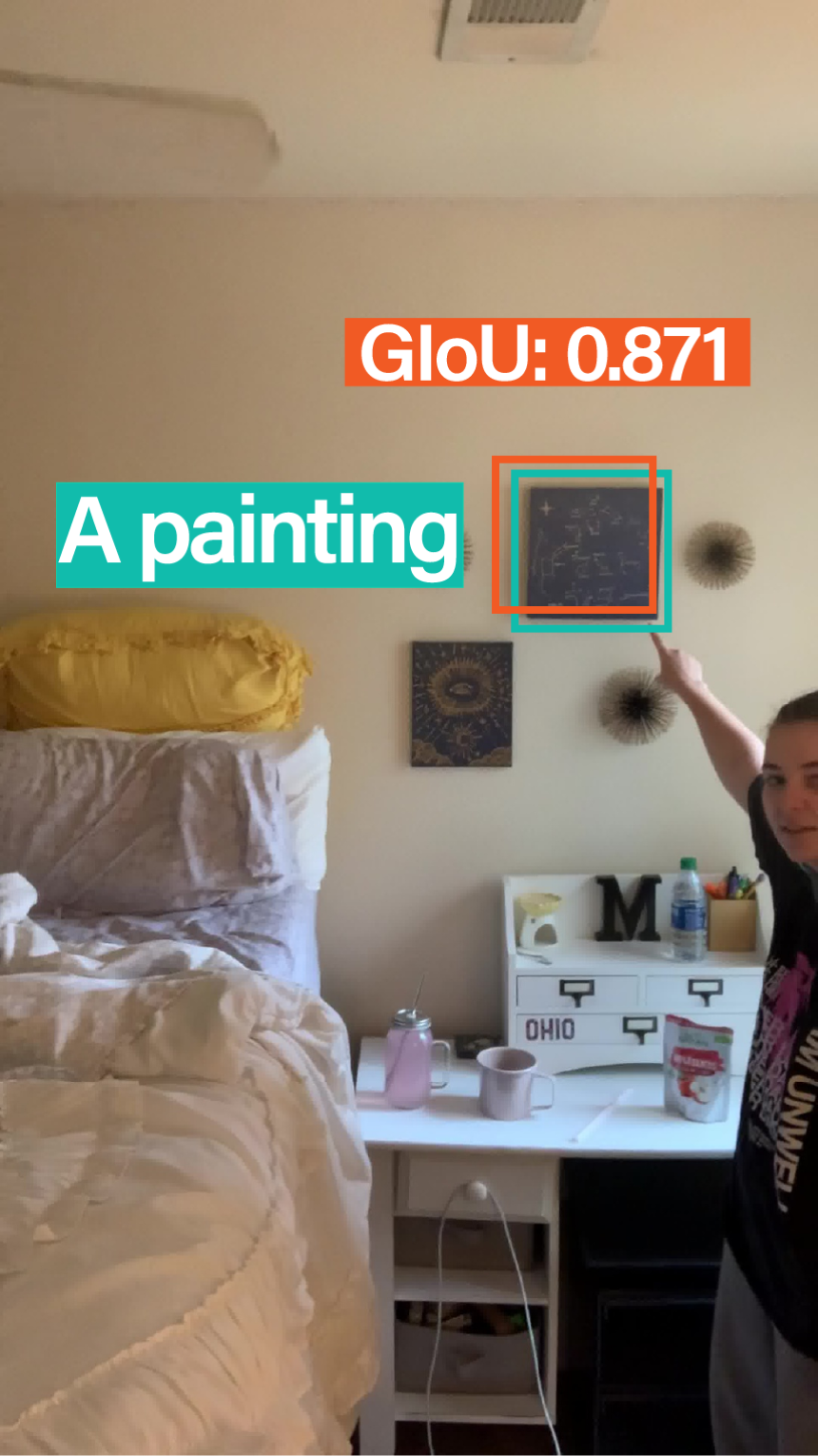}%
            \includegraphics[width=.5\linewidth]{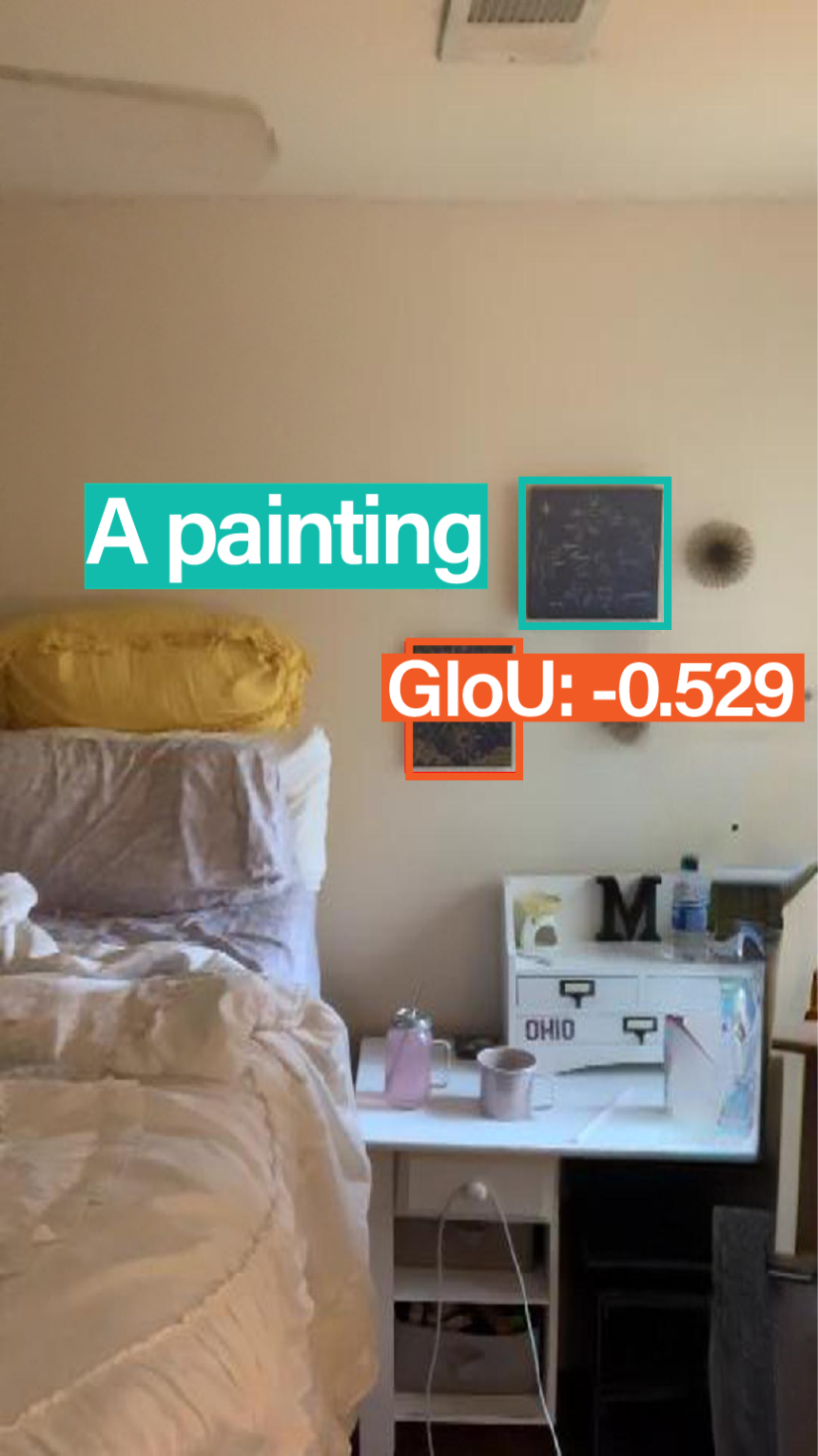}%
            \caption{}
        \end{subfigure}%
    \end{minipage}
    \caption{\textbf{Training with inpainted images leads to a performance drop.} Removing human contexts leads to scenarios wherein determining which object is referred to by the natural language expression becomes difficult (a, b, c, i) or even impossible (d, e, f, g, h, j). We call it impossible to determine the referent using natural language input when multiple objects in the image satisfy the descriptions in the natural language input sentence. (Green sentence: natural language input; Green box: ground-truth referent location; Red box: predicted referent location; Red numbers near boxes: GIoU between the predicted box and ground truth box.)}
    \label{fig:perf_drop} 
\end{figure}

\clearpage
\subsection{Implicitly Learned Nonverbal Signals}

\begin{wraptable}{r}{0.28\linewidth}
    \vspace{-6pt}
    \caption{\textbf{Effects of implicitly learned nonverbal signals.}}
    \label{tab:inpaint}
    \centering
    \small
    \resizebox{\linewidth}{!}{
        \begin{tabular}{ccc}
            \toprule
            IoU     & Original     & Inpainting \\
            \midrule
            0.25 & 64.9 & 59.1 \textcolor[RGB]{0,200,0}{(-5.8)}\\
            0.50 & 57.4 & 51.3 \textcolor[RGB]{0,200,0}{(-6.1)}\\
            0.75 & 37.2 & 32.4 \textcolor[RGB]{0,200,0}{(-4.8)}\\
            \bottomrule
        \end{tabular}%
    }%
\end{wraptable}

We further investigate whether our model can implicitly learn nonverbal gestural signals without being explicitly asked to predict the \acp{vtl} or the \acp{ewl}. Our model performs much worse when nonverbal gestural signals are absent. Specifically, removing humans from input images results in 5.8\%, 6.1\%, and 4.8\% performance drop under the IoU threshold of 0.25, 0.50, and 0.75, respectively; see details in \cref{tab:inpaint}.

These performance drops indicate that the implicitly learned model lacks useful information given only the inpainted images. Specifically, removing humans in the scenes also eliminates useful communicative signals, including pointing, poses, and the gaze direction \citep{sebanz2006joint}. The lack of human context introduces additional ambiguities in identifying the referents, especially when the natural language input alone cannot uniquely refer to an object in the scene.

For example in \cref{fig:perf_drop:f}, given the natural language input ``the books in front of me,'' determining which piles of books this person refers to is difficult: one on the table and one in the scene's lower right corner, thus in need of nonverbal gestural signals, such as pointing. Similarly in \cref{fig:perf_drop:d}, we cannot distinguish which picture frame the person is referring to without considering the pointing gesture.

\subsection{Additional Results}

\paragraph{Attention weights visualizations}

\begin{figure*}[h!]
    \centering
    \vspace{+16 pt}
    \begin{subfigure}[b]{.5\linewidth}
        \centering
        \begin{overpic}
            [width=.5\linewidth]{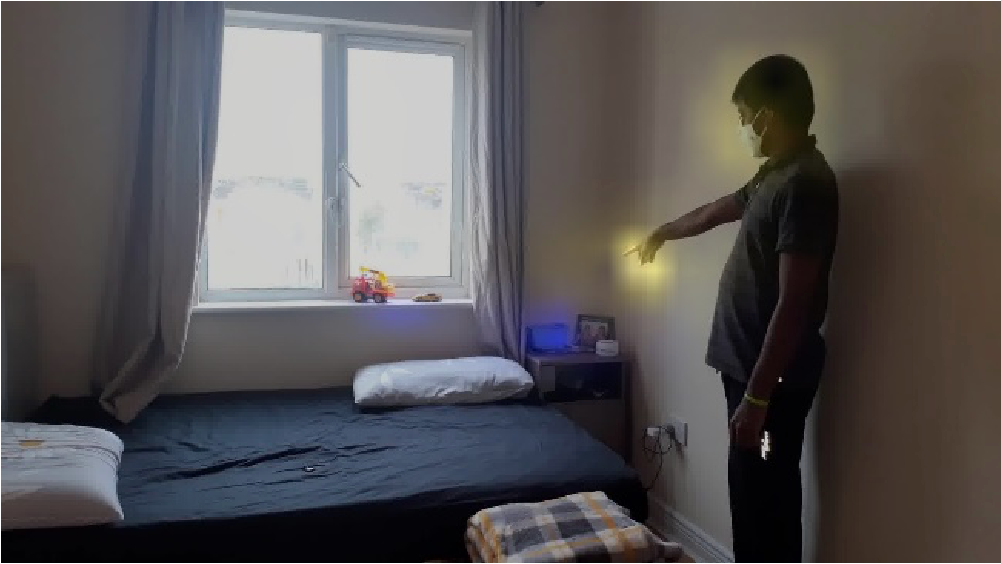}
            \put(30,60){\color{black}Attention}
        \end{overpic}%
        \begin{overpic}
            [width=.5\linewidth]{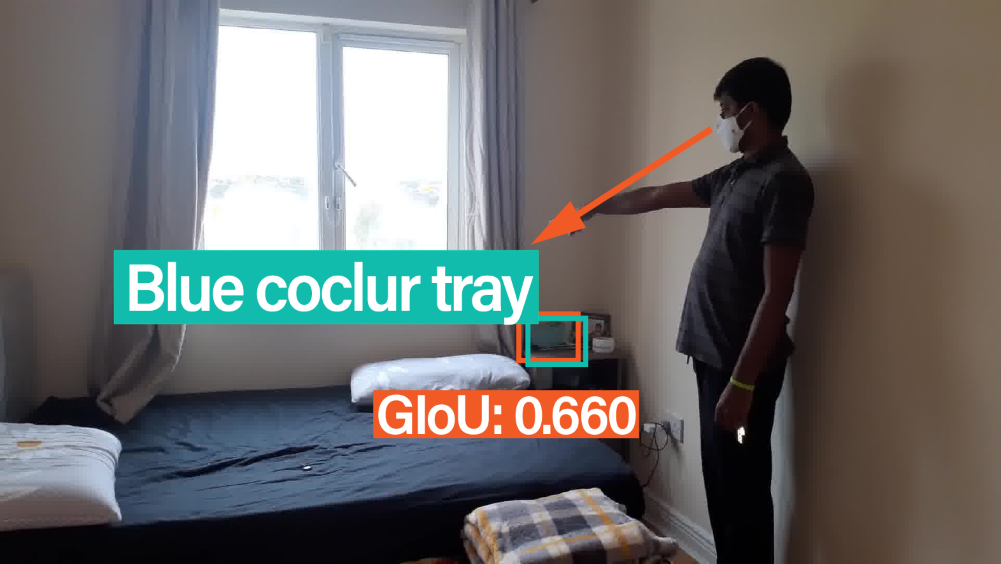}
            \put(30,60){\color{black}Prediction}
        \end{overpic}%
    \end{subfigure}%
    \begin{subfigure}[b]{.5\linewidth}
        \centering
        \begin{overpic}
            [width=.5\linewidth]{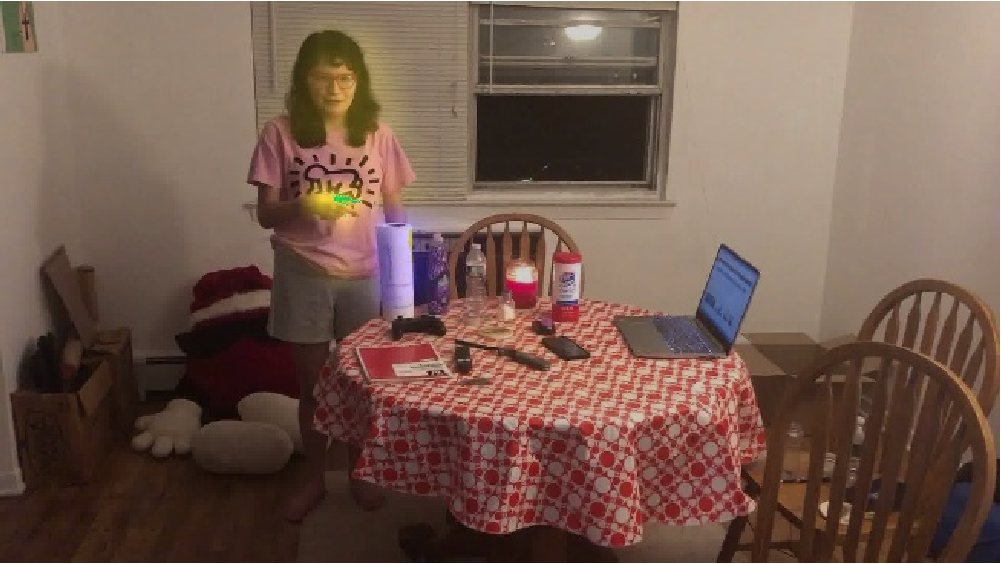}
            \put(30,60){\color{black}Attention}
        \end{overpic}%
        \begin{overpic}
            [width=.5\linewidth]{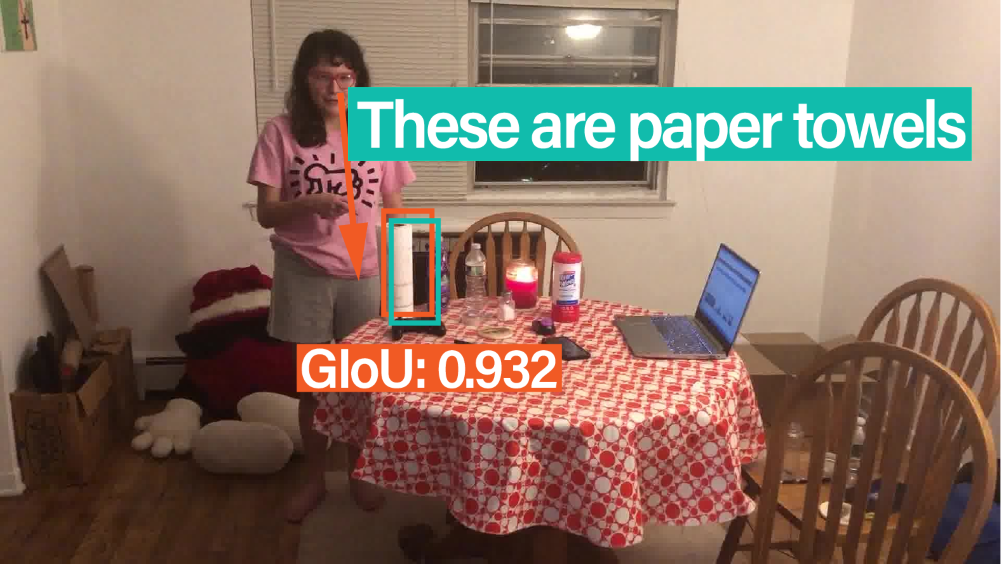}
            \put(30,60){\color{black}Prediction}
        \end{overpic}%
    \end{subfigure}%
    \\%
    \begin{subfigure}[b]{.5\linewidth}
        \centering
        \includegraphics[width=.5\linewidth]{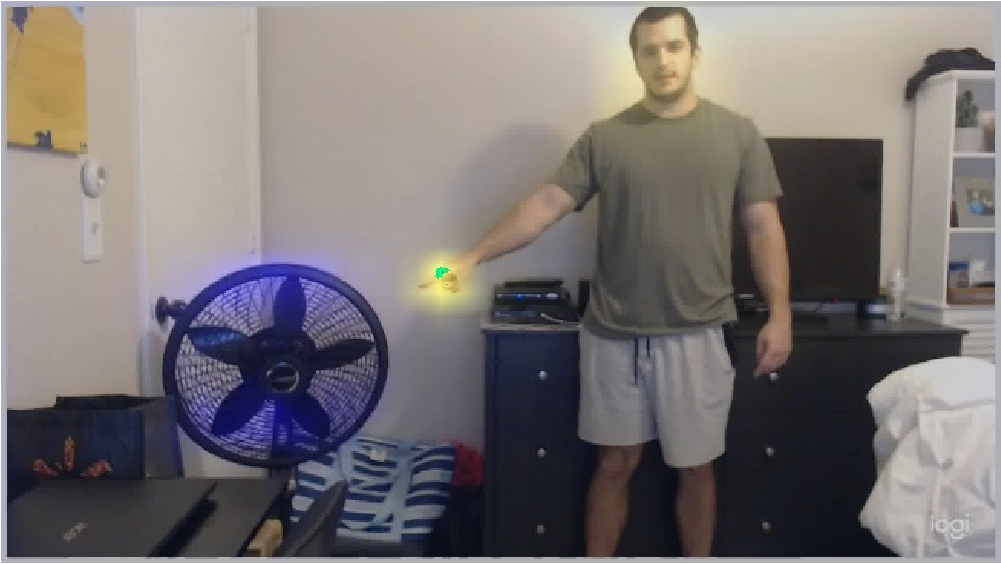}%
        \includegraphics[width=.5\linewidth]{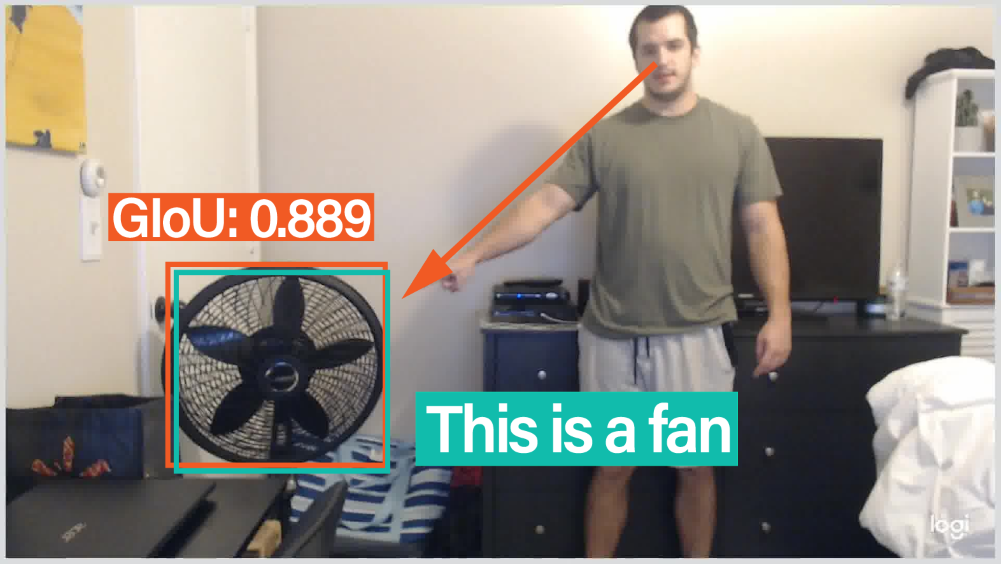}%
    \end{subfigure}%
    \begin{subfigure}[b]{.5\linewidth}
        \centering
        \includegraphics[width=.5\linewidth]{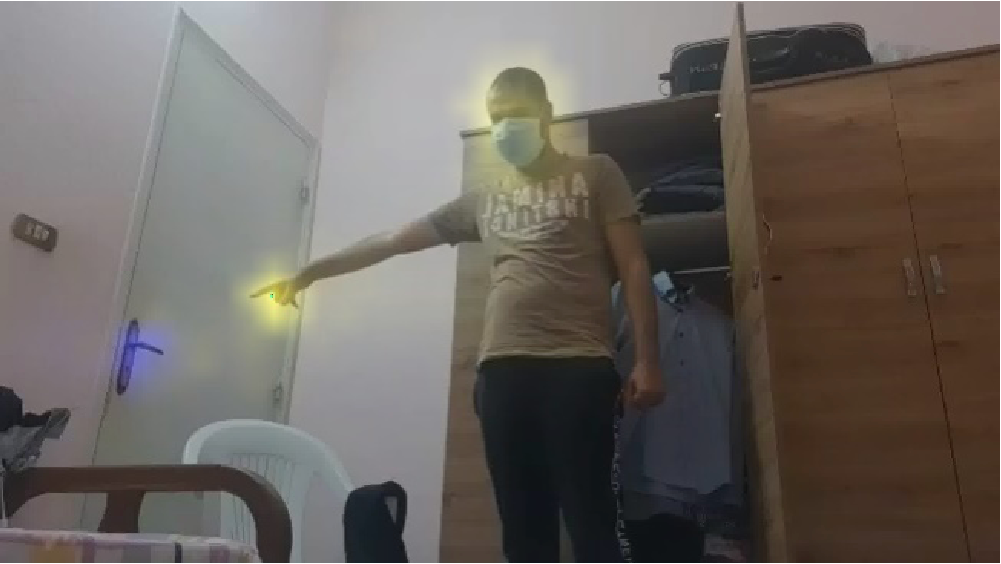}%
        \includegraphics[width=.5\linewidth]{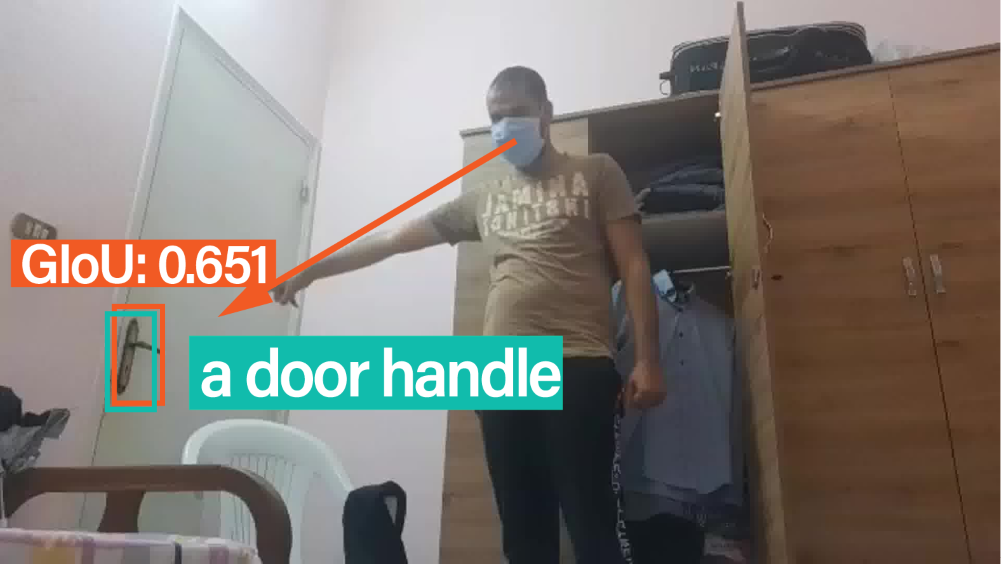}%
    \end{subfigure}%
    \\%
    \begin{subfigure}[b]{.5\linewidth}
        \centering
        \includegraphics[width=.5\linewidth]{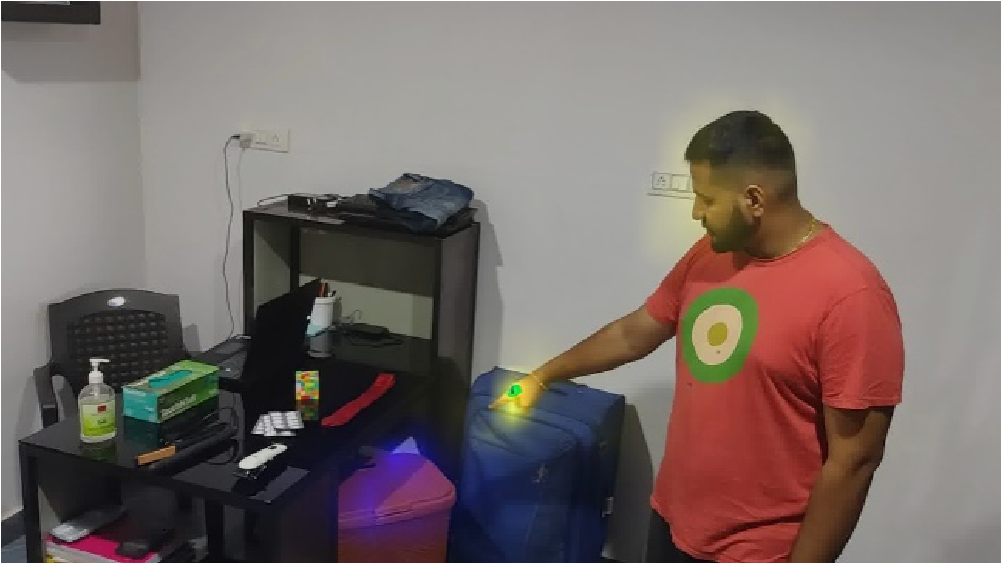}%
        \includegraphics[width=.5\linewidth]{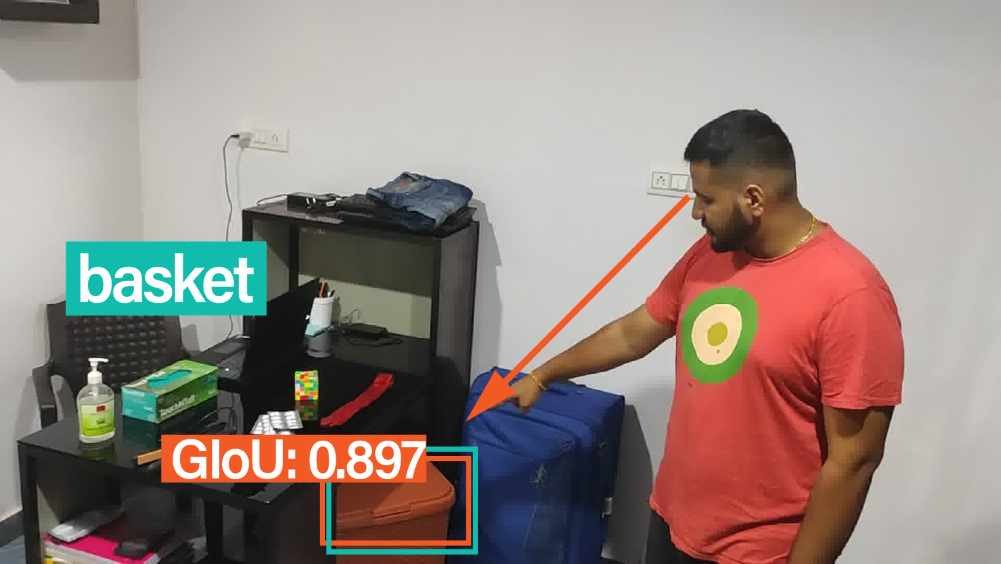}%
    \end{subfigure}%
    \begin{subfigure}[b]{.5\linewidth}
        \centering
        \includegraphics[width=.5\linewidth]{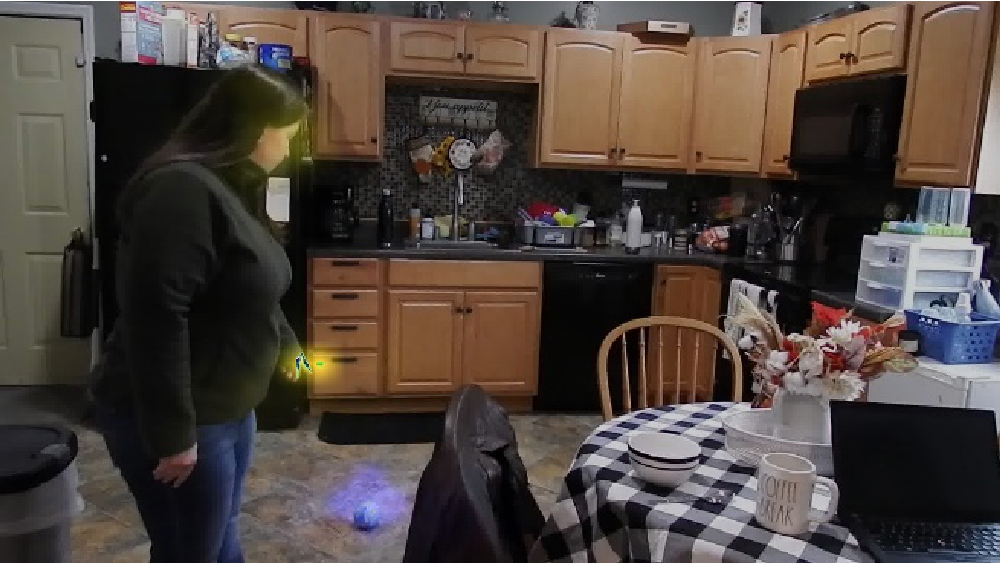}%
        \includegraphics[width=.5\linewidth]{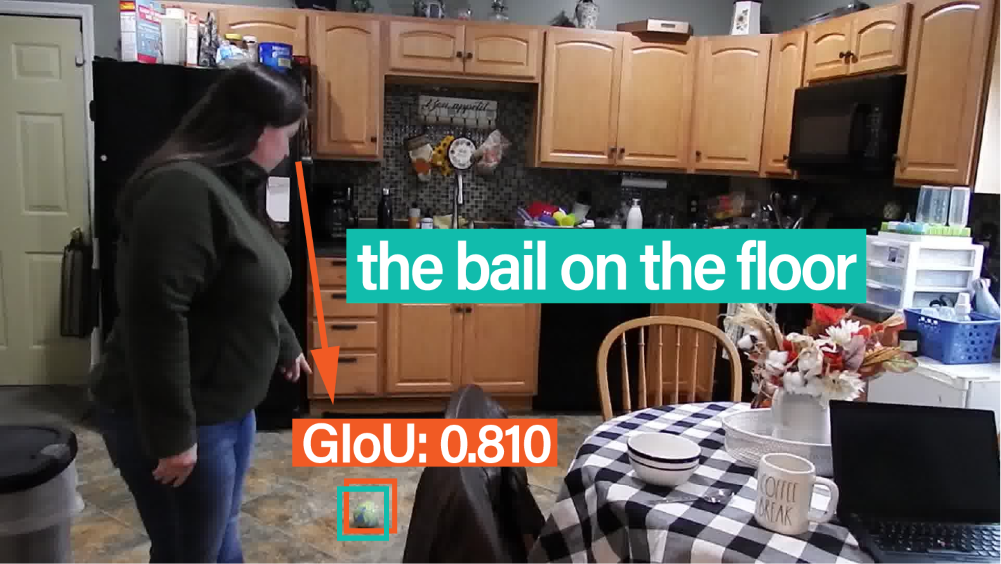}%
    \end{subfigure}%
    \caption{\textbf{Attention weights visualizations.} Attention visualization of models trained to explicitly learn \acp{vtl}. (Blue: attention from object tokens; Yellow: attention from gestural key point tokens; Green sentence: natural language input; Green box: ground-truth referent location; Red box: predicted referent location; Red numbers near boxes: GIoU between the predicted box and ground truth box.)}
    \vspace{+12 pt}
    \label{fig:attention}
\end{figure*}

We visualize the attention weights of our model trained with \ac{vtl} (\cref{fig:attention}). We use yellow to visualize the attention weights of matched gesture keypoint queries and blue for matched object queries. Visualizations show that the attention of object queries can attend to the regions of the target objects while the attention of keypoint queries primarily focus on humans' heads and hands. Taken together, these results indicate that our model successfully learns gestural features that boost the model performance.

\paragraph{Failure cases}

While the \ac{vtl} effectively helps the model leverage the nonverbal gestural signals, the performance of our model still could not surpass the 74\% human performance. The gap between model and human performance is partly attributed to the model's inability to distinguish the subtle differences between similar objects. In other words, while explicitly predicting the \acp{vtl} helps our model utilize gestural signals to determine the orientation of the object, it does not help the model recognize the minor differences between multiple objects in the same orientation.

\begin{figure}[h!]
    \centering
    \vspace{+12pt}
    \begin{subfigure}[b]{0.25\linewidth}
        \includegraphics[width=\linewidth]{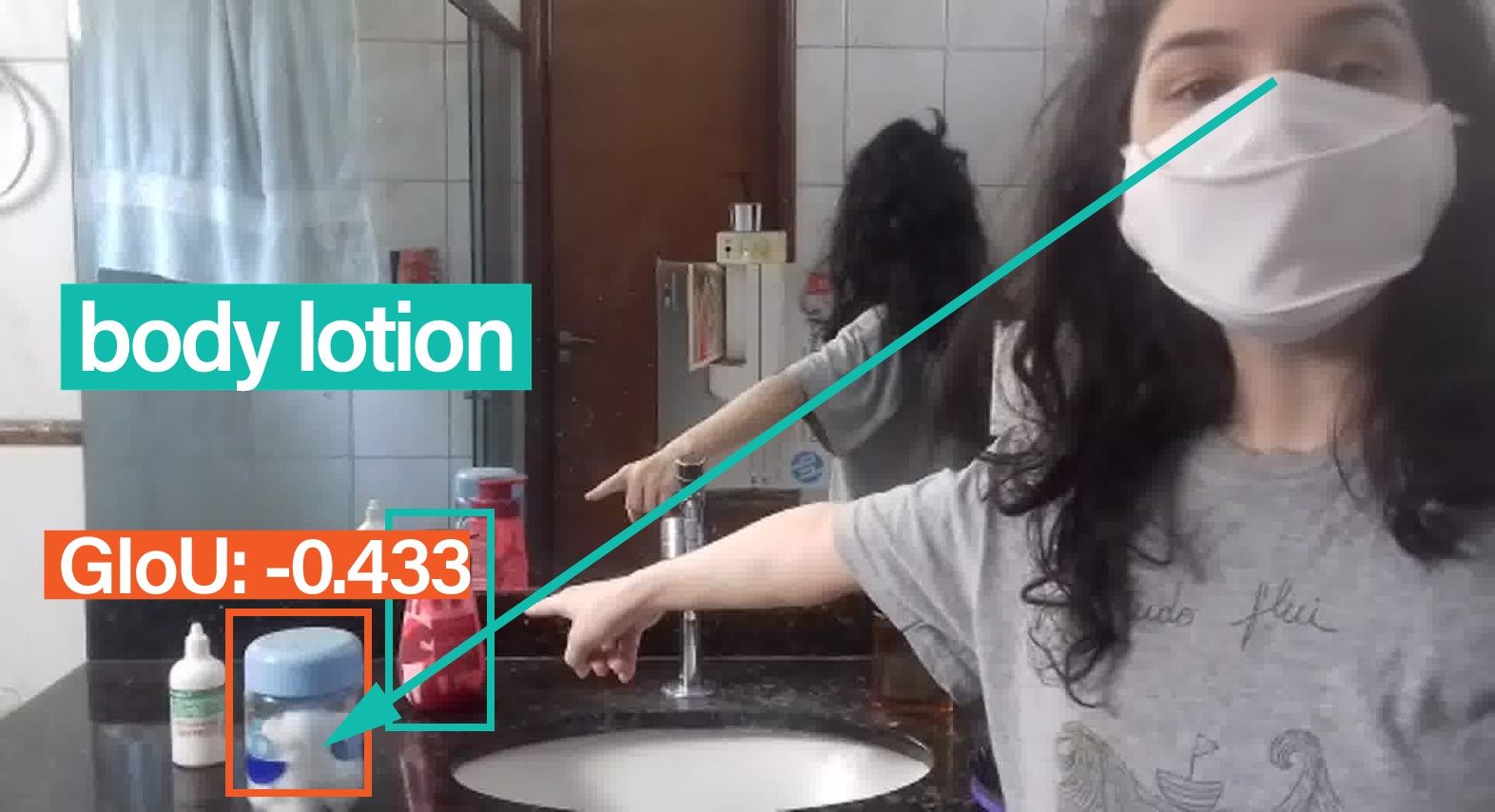}%
        \caption{}
        \label{fig:failure:a}
    \end{subfigure}%
    \begin{subfigure}[b]{0.25\linewidth}
        \includegraphics[width=\linewidth]{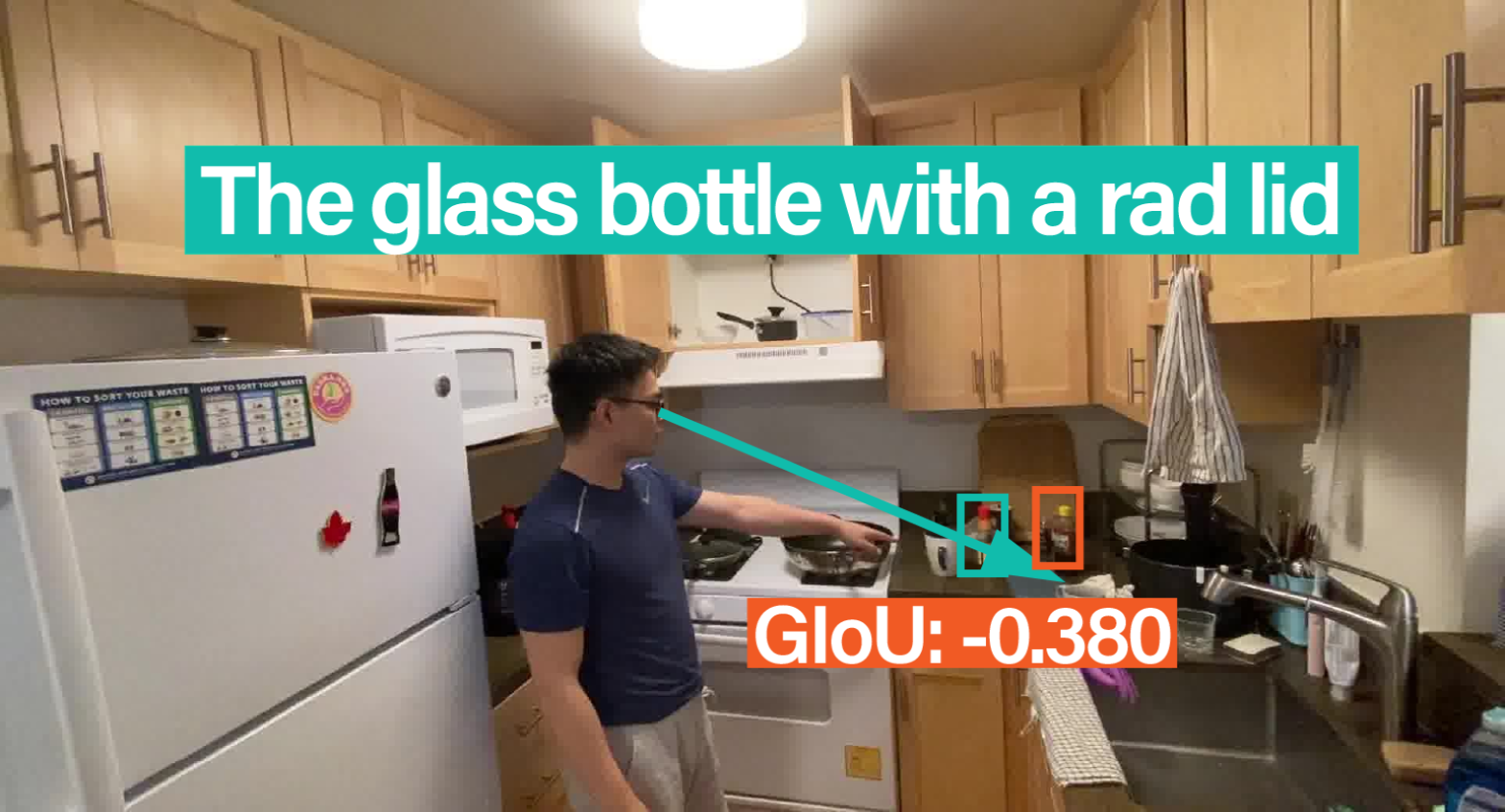}%
        \caption{}
        \label{fig:failure:b}
    \end{subfigure}%
    \begin{subfigure}[b]{0.25\linewidth}
        \includegraphics[width=\linewidth]{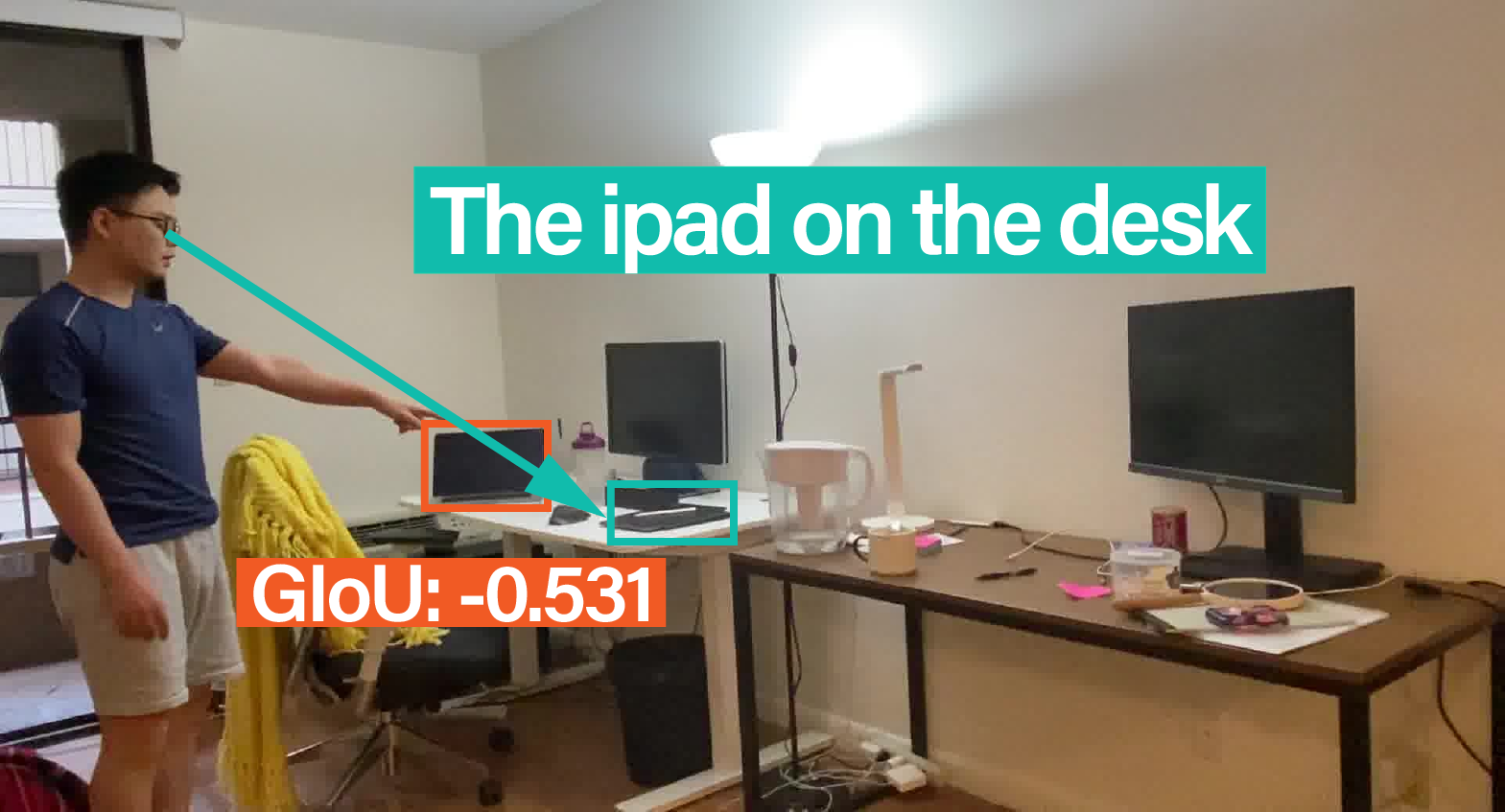}%
        \caption{}
    \end{subfigure}%
    \begin{subfigure}[b]{0.25\linewidth}
        \includegraphics[width=\linewidth]{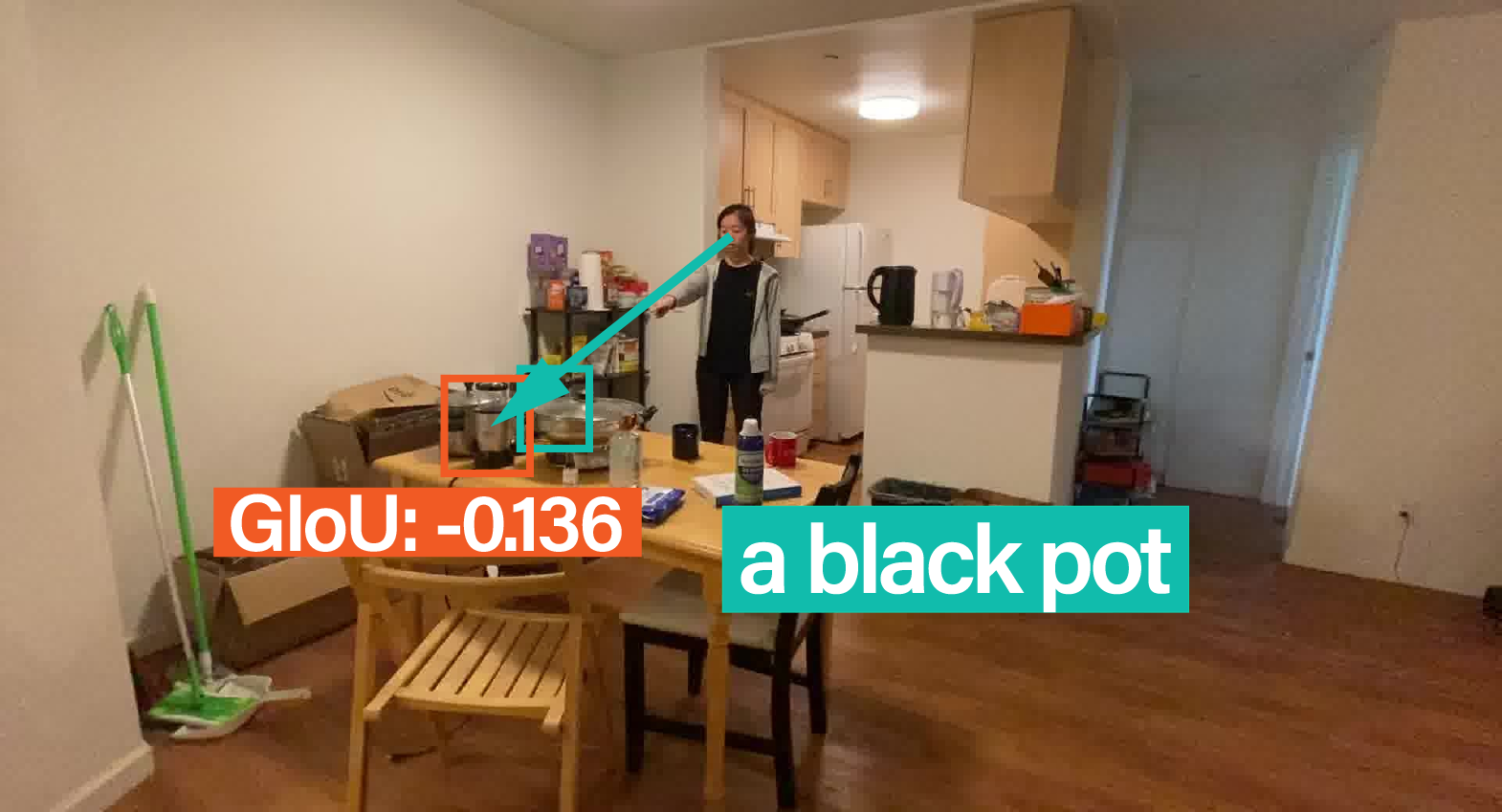}%
        \caption{}
    \end{subfigure}%
    \caption{\textbf{Examples of failure cases.} Green sentence: natural language input; Green box: ground-truth referent location; Red box: predicted referent location; Red numbers near boxes: GIoU between the predicted box and ground truth box.}
    \label{fig:failure} 
\end{figure}

For example, the \ac{vtl} passes two bottles in \cref{fig:failure:a}. While the \ac{vtl} indeed helps to narrow down the options to these two bottles, it does not help our model recognize which one of them is body lotion. Similarly in \cref{fig:failure:b}, the \ac{vtl} indicates two bottles---one with a red lid and the other with a yellow lid; yet it does not help the model distinguish the subtle differences between the lid colors.

\begin{figure}[h!]
    \centering
    \vspace{+12pt}
    \begin{subfigure}[b]{0.333\linewidth}
        \centering
        \begin{overpic}
            [width=1\linewidth]{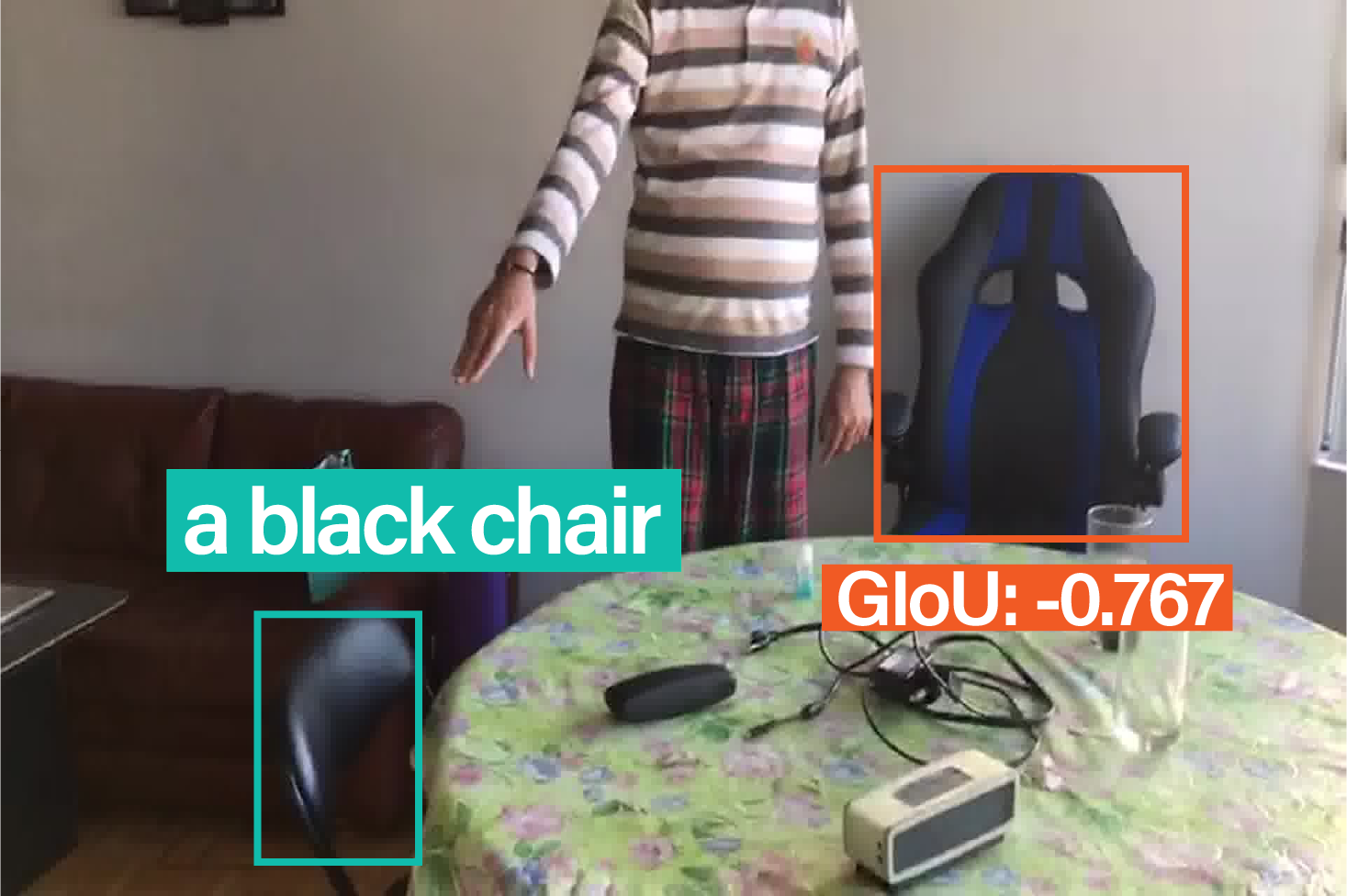}
        \end{overpic}%
        \caption{}
        \label{fig:invisible_head:a}
    \end{subfigure}%
    \begin{subfigure}[b]{0.333\linewidth}
        \centering
        \begin{overpic}
            [width=1\linewidth]{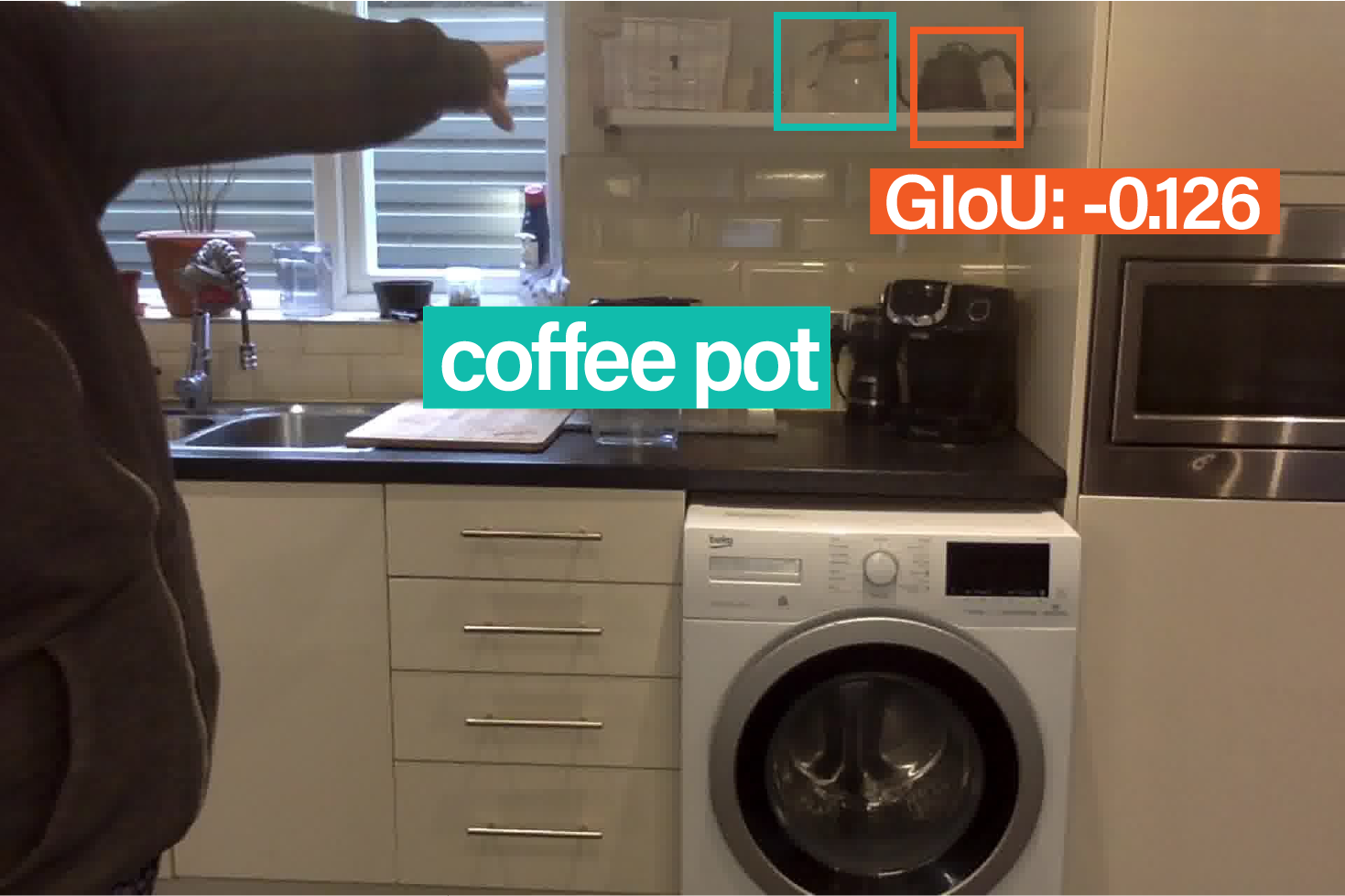}
        \end{overpic}%
        \caption{}
        \label{fig:invisible_head:b}
    \end{subfigure}%
    \begin{subfigure}[b]{0.333\linewidth}
        \centering
        \begin{overpic}
            [width=1\linewidth]{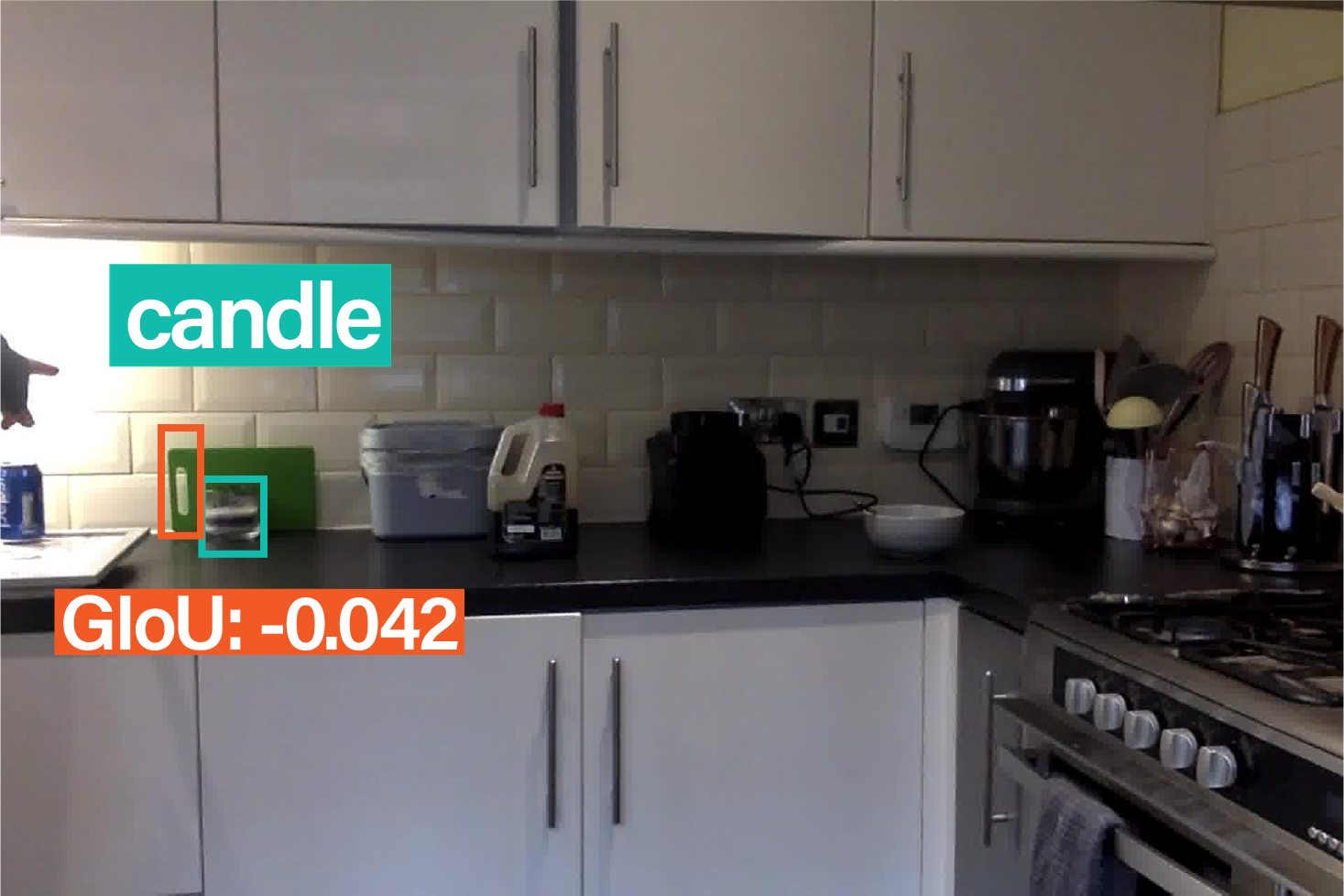}
        \end{overpic}%
        \caption{}
        \label{fig:invisible_head:c}
    \end{subfigure}%
    \caption{\textbf{Rare cases where the human head is not visible.}}
    \vspace{+12pt}
    \label{fig:invisible_head} 
\end{figure}

Additionally, in very rare cases  (\cref{fig:invisible_head}), the application of our VTL is limited because the human head is not visible from the input image. Without human heads, a touch line cannot be drawn in the image. 

\paragraph{Computed cosine similarities}

The computed cosine similarity is much higher when using the \acp{vtl} (\cref{tab:cossim}), indicating the \acp{vtl} are more co-linear to the referents. This result verifies our hypothesis that the \acp{vtl} are more reliable in referring to the objects' locations.

\begin{table}[htb!]
    \begin{minipage}[t]{0.5\linewidth}
        \vspace{+12pt}
        \caption{\textbf{Computed cosine similarities.} (tgt: computed using ground-truth box centers; pred: computed using predicted box centers.)}
        \label{tab:cossim}
        \centering
        \small
        \begin{tabular}{ccc}
            \toprule
            gesture & cos sim tgt & cos sim pred \\
            \midrule
            \ac{ewl} & 0.9580 & 0.9579 \\
            \ac{vtl} & 0.9901 & 0.9878 \\
            \bottomrule
            \vspace{+6pt}
        \end{tabular}
    \end{minipage}
    \begin{minipage}[t]{0.5\linewidth}
        \vspace{+12pt}
        \caption{\textbf{Effects of the reference alignment loss.} Removing the reference alignment loss results in performance drops in locating referents.  }
        \setlength{\tabcolsep}{3pt}
        \label{tab:arm_align_coef}
        \centering
        \small
        \resizebox{\linewidth}{!}{
            \begin{tabular}{cccccc}
                \toprule
                gesture  & ref. align. loss   & IoU=.25     & IoU=.50 & IoU=.75 & cos sim pred\\
                \midrule
                \ac{vtl} & True & 71.1  & 63.5 & 39.0 & 0.9878  \\
                \ac{vtl} & False  & 67.1 & 59.0 & 36.4 & 0.9815 \\
                & & \textcolor[RGB]{0,200,0}{(-4.0)} & \textcolor[RGB]{0,200,0}{(-4.5)} & \textcolor[RGB]{0,200,0}{(-2.6)} & \textcolor[RGB]{0,200,0}{(-0.0063)} \\
                \bottomrule
                \vspace{+6pt}
            \end{tabular}
        }%
        \setlength{\tabcolsep}{6pt}
    \end{minipage}
\end{table}

\begin{table}[h!]
    \caption{\textbf{Model performances \wrt the object detection objects of different object sizes.} \textcolor[RGB]{200,0,0}{Red} and \textcolor[RGB]{0,0,200}{blue} represents \textcolor[RGB]{200,0,0}{the first} and \textcolor[RGB]{0,0,200}{the second} highest performance, respectively.}
    \label{fig:perf_size}
    \centering
    \resizebox{\linewidth}{!}{
        \begin{tabular}{lcccccccccccc}
            \toprule
            IoUs & \multicolumn{4}{c}{0.25} & \multicolumn{4}{c}{0.50} & \multicolumn{4}{c}{0.75}\\
            \midrule
            Object Sizes & All & S & M & L & All & S & M & L & All  & S & M & L\\
            \midrule
            FAOA \citep{yang2019fast} & 44.5 & 30.6 & 48.6 & 54.1 & 30.4 & 15.8 & 36.2 & 39.3 & 8.5 & 1.4 & 9.6 & 14.4\\
            ReSC \citep{yang2020improving} & 49.2 & 32.3 & 54.7 & 60.1 & 34.9 & 14.1 & 42.5 & 47.7 & 10.5 & 0.2 & 10.6 & 20.1\\
            YourefIt PAF \citep{chen2021yourefit} & 52.6 & 35.9 & 60.5 & 61.4 & 37.6 & 14.6 & 49.1 & 49.1  & 12.7 & 1.0 & 16.5 & 20.5  \\
            YouRefIt Full \citep{chen2021yourefit} & 54.7 & 38.5 & 64.1 & 61.6 & 40.5 & 16.3 & 54.4 & 51.1 & 14.0 & 1.2 & 17.2 & 23.3 \\\midrule
            Ours (Inpainting) & 59.0  & 41.3 & 59.3 & 75.8 & 51.3 & 32.1 & 54.6 & 66.7  & 32.4 & 9.4 & 33.3 & 53.4     \\
            
            \textbf{Ours (No Pose)} & \textcolor[RGB]{0,0,0}{64.9}  & \textcolor[RGB]{0,0,0}{49.6} & \textcolor[RGB]{0,0,0}{67.9} & \textcolor[RGB]{0,0,0}{76.7} & \textcolor[RGB]{0,0,0}{57.4} & \textcolor[RGB]{0,0,0}{40.8} & \textcolor[RGB]{0,0,0}{62.1} & \textcolor[RGB]{0,0,0}{69.0}  & \textcolor[RGB]{0,0,200}{37.2} & \textcolor[RGB]{200,0,0}{14.4} & \textcolor[RGB]{0,0,200}{39.7} & \textcolor[RGB]{0,0,200}{56.7}     \\
            
            \textbf{Ours (\ac{ewl})} & \textcolor[RGB]{0,0,200}{69.5}  & \textcolor[RGB]{200,0,0}{56.6} & \textcolor[RGB]{0,0,200}{71.7} & \textcolor[RGB]{0,0,200}{80.0} & \textcolor[RGB]{0,0,200}{60.7} & \textcolor[RGB]{0,0,200}{44.4} & \textcolor[RGB]{0,0,200}{66.2} & \textcolor[RGB]{0,0,200}{71.2}  & \textcolor[RGB]{0,0,0}{35.5} & \textcolor[RGB]{0,0,0}{11.8} & \textcolor[RGB]{0,0,0}{38.9} & \textcolor[RGB]{0,0,0}{55.0}     \\
            
            \textbf{Ours (\ac{vtl})} & \textcolor[RGB]{200,0,0}{71.1}  & \textcolor[RGB]{0,0,200}{55.9} & \textcolor[RGB]{200,0,0}{75.5} & \textcolor[RGB]{200,0,0}{81.7} & \textcolor[RGB]{200,0,0}{63.5} & \textcolor[RGB]{200,0,0}{47.0} & \textcolor[RGB]{200,0,0}{70.2} & \textcolor[RGB]{200,0,0}{73.1}  & \textcolor[RGB]{200,0,0}{39.0} & \textcolor[RGB]{0,0,200}{13.4} & \textcolor[RGB]{200,0,0}{45.2} & \textcolor[RGB]{200,0,0}{57.8}     \\
            \midrule
            Human \citep{chen2021yourefit} & 94.2  & 93.7 & 92.3 & 96.3 & 85.8 & 81.0 & 86.7 & 89.4  & 53.3 & 33.9 & 55.9 & 68.1 \\
            \bottomrule
            \vspace{+12pt}
        \end{tabular}%
    }%
\end{table}

\paragraph{Reference alignment loss}

An ablation study (\cref{tab:arm_align_coef}) shows that our reference alignment loss plays a significant role in leveraging nonverbal gestural signals.

\paragraph{Effects of object sizes}

We evaluate the model performance \wrt the object detection of different object sizes (\cref{fig:perf_size}). We define small (S), medium (M), and large (L) objects following the size thresholds in \citet{chen2021yourefit}. Compared to the relatively consistent human performance, deep learning models' performances are significantly lower when detecting diminutive objects than when detecting larger objects, especially under the less stringent IoU thresholds of 0.25 and 0.50. The degraded performance indicates that future artificial intelligence models need to improve the performance of small object detection.

\paragraph{Precision computed using GIoU thresholds}\label{sec:precision_giou}

We provide the performance of our models when evaluated using GIoU instead of IoU thresholds; see \cref{tab:giou_performance}.

\begin{wraptable}{r}{0.4\linewidth}
    \vspace{-6pt}
    \caption{\textbf{Model performances when evaluated using GIoU thresholds.}}
    \label{tab:giou_performance}
    \centering
    \resizebox{\linewidth}{!}{
        \begin{tabular}{llll}
            \toprule
            & IoU=.25     & IoU=.50 & IoU=.75 \\
            \midrule
            Ours (Inpainting) & 57.9  & 50.9 & 31.4   \\
            Ours (No Pose) & 63.7 & 56.5 & 36.2   \\
            Ours (\ac{ewl}) & 67.9 & 59.7 & 34.8   \\
            Ours (\ac{vtl}) & 70.0 & 62.5 & 38.2\\
            \bottomrule
        \end{tabular}%
    }%
    \vspace{-6pt}
\end{wraptable}

\section{Conclusion and Limitations}

We presented an effective approach, named Touch-Line Transformer, to utilize the simple but effective \acp{vtl} to improve an artificial agent's ability to locate referents referred to by humans in the wild. Our approach is inspired by recent findings in psychology studies on the touch-line hypothesis, which further revealed that people frequently misinterpret other people's referring expressions.
Our proposed architecture, combined with the proposed \ac{vtl}, significantly reduced the gap between model performance and human performance. 


Some limitations exist in our work. First, resizing operations before and after inpainting might influence model performance. Next, we primarily study the eyes' location and the upper limbs' orientation regarding nonverbal signals, leaving the study of other types of nonverbal signals to future works, such as gazing direction, the direction of the finger, and the orientation of the lower limbs.



\bibliographystyle{iclr2023_conference}
\clearpage
\bibliography{reference}

\clearpage
\appendix
\section{Appendix}

\subsection{Inpainting}\label{appx:inpaint}

We hypothesize that our model may learn to use postural signals without being explicitly asked to. To test our hypothesis, we remove postural signals (by removing humans) from the images mainly using the Mask R-CNN \citep{massa2018maskrcnn} and MAT \citep{li2022mat}. We investigate whether our model performs worse when trained using images without gestural signals. 

We hypothesize that transformer models may learn to use postural signals without being explicitly asked to learn this type of signals. To test our hypothesis, we conduct two groups of experiments: the inpainting group and the control group. In the inpainting group, we remove postural signals in the input image. In the control group, we do not modify the input image. 

\begin{figure}[h]
    \centering
    \includegraphics[width=\linewidth]{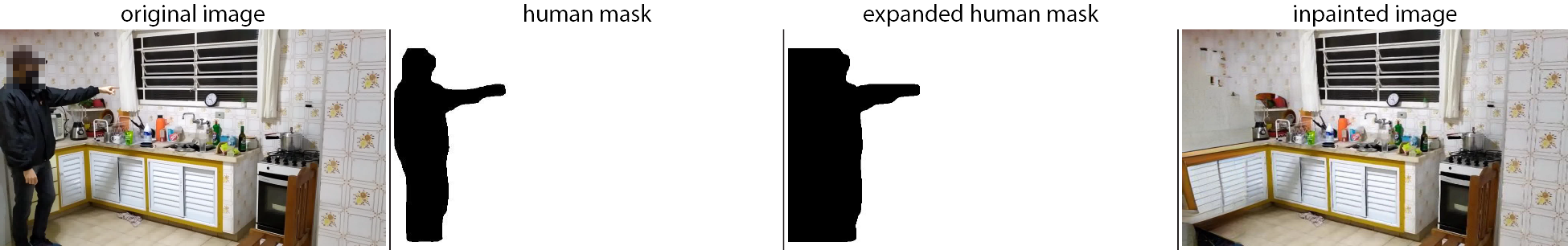}
    \caption{Illustration of the inpainting process. We remove gestural signals from input images before feeding images into our model to study the effects of implicitly learned postural signals.}
    \label{inpaint_illustration} 
\end{figure}

In the inpainting group, we modify input images. Specifically, we remove postural signals from input images by removing humans and filling missing portions of the image (which were originally occupied by humans) using MAT \citep{li2022mat}. Specifically, we first use Mask R-CNN X-101 \citep{massa2018maskrcnn} to produce human masks. After that, we expand the human masks produced by the mask rcnn to both the left and the right sides to completely cover the edge of humans. We make sure that the expanded mask never encroaches on regions occupied by the ground truth bounding box for the referent. After that, we feed the expanded masks into MAT. With input masks, MAT removes the regions covered by the masks and fills these regions. Examples of masks, expanded masks, and inpaintings are in \cref{inpaint_illustration}.

We remove gestural signals (through inpainting) from images when studying our model's ability to implicitly learn these signals. Before generating inpaintings, we expand the human mask to both sides by 50 pixels. We reshape masks and images to 512 $\times$ 512 before feeding them into the MAT model because the checkpoint produced by MAT only works for inputs of size 512 $\times$ 512. Outputs of the MAT model are reshaped to their original sizes before feeding into our model. We observe that, for a very small number of images, human masks cannot be generated by F-RCNNs. In these very rare cases, we use the original image instead.


\end{document}